\documentclass{article} 
\usepackage{iclr2020_conference,times}

\usepackage{amsmath,amsfonts,bm}

\def\eqref#1{equation~\ref{#1}}

\def\1{\bm{1}}

\DeclareMathAlphabet{\mathsfit}{\encodingdefault}{\sfdefault}{m}{sl}
\SetMathAlphabet{\mathsfit}{bold}{\encodingdefault}{\sfdefault}{bx}{n}

\newcommand{\pdata}{p_{\rm{data}}}

\DeclareMathOperator*{\argmin}{arg\,min}

\usepackage[hidelinks]{hyperref}
\usepackage{url}
\usepackage{notation}
\usepackage{graphicx}
\usepackage{booktabs}
\usepackage{multirow}
\usepackage{nicefrac}
\usepackage{balance}
\usepackage{enumitem}
\usepackage{xcolor}
\usepackage{tikz}
\usepackage{multirow} 
\usepackage{caption}
\usepackage{subcaption}

\graphicspath{{./figs/}}

\title{From Variational\\ to Deterministic Autoencoders}

\author{Partha Ghosh\dag\thanks{Equal contribution.}\addtocounter{footnote}{-1} \And Mehdi S. M. Sajjadi\dag\footnotemark \And Antonio Vergari\ddag\And Michael Black\dag \AND Bernhard Sch{\"o}lkopf\dag\\[5pt]
\dag\ Max Planck Institute for Intelligent Systems, T\"ubingen, Germany\\
{\tt\small \{pghosh,msajjadi,black,bs\}@tue.mpg.de}\\
\textrm{\ddag\ University of California, Los Angeles, USA}\\
{\tt\small aver@cs.ucla.edu}}

\iclrfinalcopy 
\begin{document}

\maketitle


\begin{abstract}

Variational Autoencoders (VAEs) provide a theoretically-backed and
popular framework for deep generative models. However, learning a VAE
from data poses still unanswered theoretical questions and
considerable practical challenges. In this work, we propose an
alternative framework for generative modeling that is simpler, easier
to train, and deterministic, yet has many of the advantages of
VAEs. We observe that sampling a stochastic encoder in a Gaussian VAE
can be interpreted as simply injecting noise into the input of a
deterministic decoder. We investigate how substituting this kind of
stochasticity, with other explicit and implicit regularization
schemes, can lead to an equally smooth and meaningful latent space
without forcing it to conform to an arbitrarily chosen prior. To
retrieve a generative mechanism to sample new data, we introduce an
ex-post density estimation step that can be readily applied also to existing VAEs, improving their sample quality.  We show, in a rigorous empirical study, that the proposed regularized deterministic autoencoders are able to generate samples that are comparable to, or better than, those of VAEs and more powerful alternatives when applied to images as well as to structured data such as molecules. \footnote{An implementation is available at: \url{https://github.com/ParthaEth/Regularized_autoencoders-RAE-}}

\end{abstract}






\section{Introduction}
\label{sec:introduction}

Generative models lie at the core of machine learning.
By capturing the mechanisms behind the data generation process, one can reason
about data probabilistically, access and traverse the low-dimensional manifold
the data is assumed to live on, and ultimately \emph{generate new data}. It is
therefore not surprising that generative models have gained momentum in
applications such as computer vision~\citep{NIPS2015_5775, biggan},
NLP~\citep{bowman2015generating, semeniuta2017hybrid}, and
chemistry~\citep{kusner2017grammar, jin2018junction, gomez2018automatic}.

Variational Autoencoders (VAEs)~\citep{Kingma2014, Rezende2014}
cast learning representations for  high-dimensional distributions as a variational inference problem.
Learning a VAE amounts to the optimization of an objective balancing the quality of
samples that are autoencoded through a stochastic encoder--decoder pair while encouraging
the latent space to follow a fixed prior distribution.
Since their introduction, VAEs have become one of the frameworks of choice among the different 
generative models. VAEs promise theoretically well-founded and more stable
training than Generative Adversarial Networks (GANs)~\citep{Goodfellow2014} and
more efficient sampling mechanisms than autoregressive
models~\citep{Larochelle2011,germain2015made}.
%

However, the VAE framework is still far from
delivering the promised generative mechanism,
as there are several practical and theoretical challenges yet to
be solved.
A major weakness of VAEs is the tendency to strike an unsatisfying compromise between sample quality and reconstruction quality.
In practice, this has been attributed to overly simplistic prior
distributions~\citep{tomczak2017vae,twostagevae} or alternatively, to the
inherent over-regularization induced by the KL
divergence term in the VAE objective~\citep{Tolstikhin2017}.
Most importantly, the VAE objective itself poses several
challenges as it admits trivial solutions that decouple the latent space from
the input~\citep{chen2016variational, zhao2017towards}, leading to the posterior
collapse phenomenon in conjunction with powerful
decoders~\citep{van2017neural}.
Furthermore, due to its variational formulation, training a VAE
requires approximating expectations through sampling at the cost of
increased variance in gradients~\citep{burda2015importance, rebar17}, making
initialization, validation, and annealing of hyperparameters essential in
practice~\citep{bowman2015generating, Higgins2017, BauMni18}.
Lastly, even after a satisfactory convergence of the objective, the learned aggregated posterior
distribution rarely matches the assumed latent prior in
practice~\citep{kingma1606improving,BauMni18,twostagevae}, ultimately hurting the
quality of generated samples.
%
All in all, much of the attention around
VAEs is still directed towards ``fixing'' the aforementioned drawbacks associated with them.

In this work, we take a different route: we question whether the
variational framework adopted by VAEs is necessary for generative
modeling and, in particular, to obtain a smooth latent space.
We propose to adopt a simpler, deterministic version of VAEs that
scales better, is simpler to optimize, and, most
importantly, still produces a meaningful latent space and equivalently
good or better samples than VAEs or stronger alternatives, \eg
Wasserstein Autoencoders (WAEs)~\citep{Tolstikhin2017}.
We do so by observing 
that, under commonly used distributional assumptions, training a
stochastic encoder--decoder  pair in VAEs does not differ from training a
deterministic architecture where noise is added to the decoder's
input. 
%
We investigate how to substitute this noise
injection mechanism with other regularization schemes in the proposed deterministic
\emph{Regularized Autoencoders} (RAEs),
and we thoroughly analyze how this affects performance.
%
Finally, we equip RAEs with a generative mechanism via a simple \de{} density estimation step on the learned latent space.

In summary, our contributions are as follows:
i) we introduce the \ourmodel{} framework for generative modeling
as a drop-in replacement for many common VAE architectures;
ii) we propose an \de{} density estimation scheme which greatly improves
  sample quality for VAEs, WAEs and \ourmodel{}s without the need to retrain the models;
  iii) we conduct a rigorous empirical evaluation
  to compare RAEs with VAEs and several baselines on standard image
  datasets and on more challenging
  structured domains such as molecule generation~\citep{kusner2017grammar, gomez2018automatic}.
%
%



\section{Variational Autoencoders}
\label{sec:vae}

For a general discussion, we consider a collection of high-dimensional
i.i.d.\
samples $\mathcal{X}=\{\x_{i}\}_{i=1}^{N}$ drawn from the true data distribution
$p_{\mathsf{data}}(\x)$ over a random variable $\X$ taking values in the input
space. The aim of generative modeling is to learn from $\mathcal{X}$ a mechanism
to draw new samples $\x_{\mathsf{new}} \sim p_{\mathsf{data}}$.
Variational Autoencoders provide a powerful latent variable framework to infer
such a mechanism. The generative process of the VAE is defined as
\begin{equation}
  \label{eq:vae-generative}
  \z_{\mathsf{new}}\sim p(\Z),
  \quad\quad
  \x_{\mathsf{new}}\sim p_{\theta} (\X\cbar\Z=\z_{\mathsf{new}})
\end{equation}
where $p(\Z)$ is a fixed prior distribution over a low-dimensional latent
space $\Z$. A stochastic decoder
\begin{equation}
  \D(\z) = \x \sim \lik = p(\X\cbar g_{\theta}(\z))
\end{equation}
links the latent space to the input space through the
\emph{likelihood} distribution $p_{\theta}$, where
$g_{\theta}$ is an expressive non-linear function parameterized by
$\theta$.\footnote{With slight abuse of notation, we use lowercase letters for
both random variables and their realizations, \eg $\lik$ instead of
$p(\X\cbar\Z=\z)$, when it is clear to discriminate between the two.} As a
result, a VAE estimates $p_{\mathsf{data}}(\x)$ as the infinite mixture model
$p_{\theta}(\x)=\int \lik\prior d\mathbf{z}$. At the same time, the input space
is mapped to the latent space via a stochastic encoder
\begin{equation}
  \En(\x) = \z \sim \post = q(\Z\cbar f_{\phi}(\x))
\end{equation}
where $\post$ is the \emph{posterior} distribution given by a second function
$f_{\phi}$ parameterized by $\phi$.
Computing the marginal log-likelihood $\log p_{\theta}(\x)$ is generally
intractable. One therefore follows a variational approach, maximizing the evidence
lower bound (ELBO) for a sample $\x$:
\begin{align}
  \label{eq:elbo}
  \log p_{\theta}(\x)\geq
   \ELBO(\phi, \theta, \x) =\ \EX_{\z\sim\post}\log\lik -\KLD(\post || \prior)
\end{align}
Maximizing Eq.~\ref{eq:elbo} over data $\mathcal{X}$ \wrt model parameters
$\phi$, $\theta$ corresponds to minimizing the loss
\begin{equation}
  \label{eq:elbo-loss}
  \argmin_{\phi,\theta}\; \EX_{\x\sim p_{\mathsf{data}}}\; \mathcal{L}_{\ELBO} =
  \EX_{\x\sim p_{\mathsf{data}}}\; \lossr + \losskl
\end{equation}
where $\mathcal{L}_{\mathsf{REC}}$ and $\mathcal{L}_{\mathsf{KL}}$ are defined
for a sample $\x$ as follows:
\begin{align}
  \label{eq:rec-kl-loss}
  \mathcal{L}_{\mathsf{REC}}=-\EX_{\z\sim\post}\log\lik\quad\quad\quad
  \mathcal{L}_{\mathsf{KL}} = \KLD(\post || \prior)
\end{align}
Intuitively, the reconstruction loss $\lossr$ takes into account the quality of
autoencoded samples $\x$ through $\D(\En(\x))$, while the KL-divergence term
$\losskl$ encourages $\post$ to match the prior $\prior$ for each $\z$ which
acts as a regularizer during training~\citep{hoffman2016elbo}.
%

\subsection{Practice and shortcomings of VAEs}
\label{sec:practice}

To fit a VAE to data through Eq.~\ref{eq:elbo-loss} one has to specify the
parametric forms for $\prior$, $\post$, $\lik$, and hence the deterministic
mappings $f_{\phi}$ and $g_{\theta}$. In practice, the choice for the above
distributions is guided by trading off computational complexity with model
expressiveness.
In the most commonly adopted formulation of the VAE, $\post$ and $\lik$ are assumed to be
Gaussian:
\begin{align}
  \label{eq:vae-sampling-encoder}
  \En(\x)\sim \mathcal{N}(\Z|\boldsymbol\mu_{\phi}(\x), \diag(\boldsymbol\sigma_{\phi}(\x)))\quad\quad\quad
  \D(\En(\x))\sim \mathcal{N}(\X|\boldsymbol\mu_{\theta}(\z),
  \diag(\boldsymbol\sigma_{\theta}(\z)))
\end{align}
with means $\mu_{\phi}, \mu_{\theta}$ and covariance parameters $\sigma_{\phi},
\sigma_{\theta}$ given by $f_{\phi}$ and $g_{\theta}$.
In practice, the covariance of the decoder is set to the identity matrix for
all $\z$, \ie $\sigma_{\theta}(\z) = 1$~\citep{twostagevae}. The expectation of
$\lossr$ in Eq.~\ref{eq:rec-kl-loss} must be approximated via $k$ Monte Carlo
point estimates.
%
It is expected that the quality of the Monte Carlo estimate,
and hence convergence during learning and sample quality increases
for larger $k$~\citep{burda2015importance}.
%
However, only a
1-sample approximation is generally carried out~\citep{Kingma2014} since
memory and time requirements are prohibitive for large $k$.
With the 1-sample
approximation, $\lossr$ can be computed as the mean squared error
between input samples and their mean reconstructions $\mu_{\theta}$
by a decoder that is deterministic in practice:
\begin{equation}
  \label{eq:vae-gauss-dec}
  \lossr = || \x - \boldsymbol\mu_{\theta}(E_\phi(\x)) ||_2^2
\end{equation}
Gradients \wrt the encoder parameters $\phi$ are computed through the expectation of $\lossr$ in Eq.~\ref{eq:rec-kl-loss} via the reparametrization
trick~\citep{Kingma2014} where the stochasticity of $\En$ is relegated to an
auxiliary random variable $\epsilon$ which does not depend on $\phi$:
\begin{equation}
  \label{eq:rep-trick}
  \En(\x) = \boldsymbol\mu_{\phi}(\x) + \boldsymbol\sigma_{\phi}(\x) \odot \boldsymbol\epsilon,
  \quad\quad
  \boldsymbol\epsilon\sim\mathcal{N}(\mathbf{0}, \mathbf{I})
\end{equation}
%
    where $\odot$ denotes the Hadamard product. An additional simplifying assumption
involves fixing the prior $\prior$ to be a $d$-dimensional isotropic Gaussian
$\mathcal{N}(\Z\cbar\mathbf{0}, \mathbf{I})$. For this choice, the KL-divergence
for a sample $\x$ is given in closed form: $2\mathcal{L}_{\mathsf{KL}}=
    ||\boldsymbol\mu_{\phi}(\x)||_{2}^{2}
    + d
    + \sum_{i}^{d}\boldsymbol\sigma_{\phi}(\x)_{i}
    - \log\boldsymbol\sigma_{\phi}(\x)_{i}$.

%
%
While the above assumptions make VAEs easy to implement,
the stochasticity in the encoder and decoder are still
problematic in practice~\citep{Makhzani2015, Tolstikhin2017, twostagevae}.
In particular, one has to carefully balance the trade-off between the $\losskl$ term
and $\lossr$ during optimization~\citep{twostagevae, BauMni18}.
%
A too-large weight on the $\losskl$ term
can dominate $\losselbo$, having the effect of
\emph{over-regularization}.
As this would smooth the latent space, it can directly affect sample
quality in a negative way.
Heuristics to avoid this
include manually fine-tuning or  gradually annealing the importance of $\losskl$  during
training~\citep{bowman2015generating, BauMni18}.
%
We also observe this trade-off in a practical experiment in Appendix~\ref{sec:rec-reg}.


Even after employing the full array of approximations and ``tricks'' to reach
convergence of Eq.~\ref{eq:elbo-loss} for a satisfactory set of parameters,
there is no guarantee that the learned latent space is distributed according to
the assumed prior distribution. In other words, the aggregated posterior
distribution $q_{\phi}(\z)=\mathbb{E}_{\x\sim p_{\mathsf{data}}}q(\z|\x)$ has
been shown not to conform well to $p(\z)$ after training~\citep{Tolstikhin2017,
BauMni18, twostagevae}. This critical issue severely hinders the generative
mechanism of VAEs (cf. Eq.~\ref{eq:vae-generative}) since latent codes sampled from
$p(\z)$ (instead of $q(\z)$) might lead to regions of the latent space that are
previously unseen to $\D$ during training. This results in
generating out-of-distribution samples. 
We refer the reader to Appendix \ref{sec:viz_ex_post_density} for
a visual demonstration of this phenomenon on the latent space of VAEs.
We analyze solutions to this problem in
Section~\ref{sec:xpde}.

\subsection{Constant-Variance Encoders}
\label{sec:cvvae}

Before introducing our fully-deterministic take on VAEs, it is worth
investigating intermediate flavors of VAEs with reduced stochasticity.
Analogous to what is commonly done for decoders as discussed in
the previous section,
one can fix the variance of $\post$ to be constant for all $\x$.
This simplifies the computation of $\En$ from Eq.~\ref{eq:rep-trick} to
%
\begin{equation}
  \label{eq:const-var-enc}
  \En^{\mathsf{CV}}(\x)
  = \boldsymbol\mu_{\phi}(\x)
  + \boldsymbol\epsilon,
  \quad\quad
  \epsilon\sim\mathcal{N}(\mathbf{0}, \sigma\mathbf{I})
\end{equation}
where $\sigma$ is a fixed scalar.
Then, the KL loss term in a Gaussian VAE
simplifies (up to a constant) to
$\mathcal{L}_{\mathsf{KL}}^{\mathsf{CV}}=
   ||\boldsymbol\mu_{\phi}(\x)||_{2}^{2}$.
%
%
We name this variant Constant-Variance VAEs (CV-VAEs).
While CV-VAEs have been adopted in some applications such as variational image
compression~\citep{balle2016end} and adversarial robustness~\citep{ghosh2019resisting}, to the best of our knowledge, there is no
systematic study of them in the literature.
We will fill this gap in our experiments in Section~\ref{sec:exps}.
Lastly,
%
note that now $\sigma$ in Eq.\ref{eq:const-var-enc} is not learned along the encoder as in
Eq.~\ref{eq:rep-trick}.
Nevertheless, it can still be fitted as an hyperparameter, \eg by
cross-validation, to maximise the model likelihood.
%
%
%
This highlights the possibility to estimate a better parametric form for the latent space
distribution after training, or in a outer-loop including training.
We address this provide a more complex and flexible solution to deal
with the prior structure over $\Z$ via \de\ density estimation in
Section~\ref{sec:xpde}.



\section{Deterministic Regularized Autoencoders}
\label{sec:method-practical}

Autoencoding in VAEs is defined in a
probabilistic fashion: $\En$ and $\D$ map data points not to a single point, but
rather to parameterized distributions
(\cf Eq.~\ref{eq:vae-sampling-encoder}).
However, common implementations of VAEs as discussed in
Section~\ref{sec:vae} admit a simpler, deterministic view for
this probabilistic mechanism.
A glance at the autoencoding mechanism of the VAE is revealing.

The encoder deterministically maps a data point $\x$  to mean $\mu_{\phi}(\x)$ and variance $\sigma_{\phi}(\x)$ in
the latent space. The input to $\D$ is then simply the mean
$\mu_{\phi}(\x)$ \emph{augmented with Gaussian noise} scaled by
$\sigma_{\phi}(\x)$ via the reparametrization trick (\cf Eq.~\ref{eq:rep-trick}). In the CV-VAE, this relationship is even more obvious, as
the magnitude of the noise is fixed for all data points
(\cf Eq.~\ref{eq:const-var-enc}). In this light, a VAE can be seen as a
\emph{deterministic} autoencoder where (Gaussian) noise is added to the decoder's
input.

We argue that this noise injection mechanism is a key factor in having
a regularized decoder.
Using random noise injection to regularize neural networks is a
well-known technique that dates back several decades~\citep{sietsma1991creating,
  an1996effects}.
It implicitly helps to smooth the function learned by the network at the price of increased variance in the gradients during training.
In turn, decoder regularization is a key component in generalization
for VAEs, as it improves random sample quality and achieves a smoother latent space.
Indeed, from a generative perspective, regularization is motivated by
the goal to learn a smooth latent space where similar data points $\x$ are
mapped to similar latent codes $\z$, and small variations in $\Z$ lead to
reconstructions by $\D$ that vary only slightly.

%
We propose to \emph{substitute noise injection with an explicit regularization scheme for the decoder}. This entails the substitution of the variational framework in VAEs, which enforces
regularization on the encoder posterior through $\losskl$, with a
deterministic framework that applies other flavors of decoder regularization.
%
%
By removing noise injection from a CV-VAE, we are effectively
left with a deterministic autoencoder (AE).
%
Coupled with explicit regularization for the decoder, 
we obtain a \emph{Regularized Autoencoder} (RAE).
Training a RAE thus involves minimizing the simplified loss
\begin{equation}
  \label{eq:rae-loss}
  \lossrae =  \lossr + \beta \lossraekl + \lambda \lossreg
\end{equation}
where $\lossreg$ represents the explicit regularizer for $\D$ (discussed in Section~\ref{sec:raereg}) and
$\lossraekl=\nicefrac{1}{2} ||\z||_{2}^{2}$ (resulting from simplifying $\mathcal{L}_{\mathsf{KL}}^{\mathsf{CV}}$)
is equivalent to constraining the size of the learned latent space, which is still needed to prevent unbounded optimization. Finally, $\beta$ and $\lambda$ are two hyper parameters that balance the different loss terms.

Note that for RAEs, no Monte Carlo approximation is required to compute $\lossr$.
This relieves the need for more samples from $\post$ to achieve better image
quality (\cf Appendix~\ref{sec:rec-reg}).
Moreover, by abandoning the
variational framework and the $\losskl$ term,  there is no need in
RAEs for a fixed prior distribution over $\Z$.
%
Doing so however loses a clear generative mechanism for RAEs to sample from $\Z$.
We propose a method to regain random sampling ability in Section~\ref{sec:xpde} by performing
density estimation on $\Z$ ex-post, a step that is otherwise still
needed for VAEs to alleviate the posterior mismatch issue.

\subsection{Regularization Schemes for RAEs}
\label{sec:raereg}

Among possible choices for $\lossreg$, a first obvious
candidate is Tikhonov regularization~\citep{tikhonov1977solutions} since is known
to be related to the addition of low-magnitude input
noise~\citep{bishop2006pattern}. Training a RAE within this framework thus
amounts to adopting
%
$
   \lossreg =  \mathcal{L}_{\mathsf{L}_{2}} = ||\theta||_{2}^{2}
$
%
which effectively applies weight decay on the decoder parameters $\theta$.

Another option comes from the recent GAN literature where regularization is a
hot topic~\citep{kurach2018gan} and where injecting noise to the input of the
adversarial discriminator has led to improved performance in a technique called
\emph{instance noise}~\citep{sonderby2014apparent}. To enforce Lipschitz
continuity on adversarial discriminators, weight clipping has been
proposed~\citep{arjovsky2017wasserstein}, which is however known to significantly
slow down training. More successfully, a \emph{gradient penalty} on the
discriminator can be used similar to~\citet{gulrajani2017improved,
mescheder2018training}, yielding the objective
%
$
  \lossreg =  \lossgp = ||\nabla \D(\En(\x))||_2^2
$
%
which bounds the gradient norm of the decoder \wrt its input.

Additionally, spectral normalization (SN) has been successfully proposed as an alternative
way to bound the Lipschitz norm of an adversarial discriminator~\citep{miyato2018spectral}.
SN normalizes each weight matrix $\theta_{\ell}$ in the decoder by an
estimate of its largest singular value:
%
$
     \theta_{\ell}^{\SN} = \theta_{\ell}/\mathsf{s}(\theta_{\ell})
$
%
where $\mathsf{s}(\theta_{\ell})$ is the current estimate obtained through the
power method.

In light of the recent successes of deep networks
\emph{without} explicit regularization~\citep{Zagoruyko2016WRN, zhang2016understanding}, it is intriguing to question
the need for explicit regularization of the decoder in order to obtain a meaningful
latent space.
The assumption here is that techniques such as
dropout~\citep{dropout}, batch normalization~\citep{batchnorm}, adding noise
during training~\citep{an1996effects} implicitly regularize the networks enough.
Therefore, as a natural baseline to the $\lossrae$
objectives introduced above, we also consider the RAE framework without
$\lossreg$ and $\lossraekl$, \ie a standard deterministic autoencoder optimizing
$\lossr$ only.

%
To complete our ``autopsy'' of the VAE loss,  we additionally investigate deterministic
autoencoders with decoder regularization, but without the $\lossraekl$
term, as well as possible combinations of different regularizers in
our RAE framework
(cf. Table \ref{tab:fids-aux} in Appendix~\ref{sec:more-regs}).

Lastly, it is worth questioning if it is possible to formally derive
our RAE framework from first principles.
We answer this affirmatively, and show how to augment the ELBO
optimization problem of a VAE with an
explicit constraint, while not fixing a parametric form for
$\post$. This indeed leads to a special case of the RAE loss in Eq.~\ref{eq:rae-loss}.
Specifically, we derive a regularizer like $\lossgp$ for a
deterministic version of the CV-VAE.
Note that this derivation legitimates bounding the
decoder's gradients and as such it justifies the spectral norm
regularizer as well since the latter enforces the decoder's Lipschitzness.
We accommodate the full derivation in Appendix~\ref{sec:method}.

\section{Ex-Post Density Estimation}
\label{sec:xpde}

By removing stochasticity and ultimately, the KL divergence term $\losskl$ from
RAEs, we have simplified the original VAE objective at the cost of detaching the
encoder from the prior $\prior$ over the latent space. This implies
that i) we cannot
ensure that the latent space $\Z$ is distributed according to a simple
distribution (\eg isotropic Gaussian) anymore and consequently, ii) we lose the
simple mechanism provided by $\prior$ to sample from $\Z$ as in
Eq.~\ref{eq:vae-generative}.

As discussed in Section~\ref{sec:practice}, issue i) is compromising the VAE
framework in any case, as reported in several works
\citep{hoffman2016elbo, dmvi, twostagevae}. To fix this, some works extend the
VAE objective by encouraging the aggregated posterior to match
$\prior$~\citep{Tolstikhin2017} or by utilizing more complex
priors~\citep{kingma1606improving, tomczak2017vae, BauMni18}.

To overcome both i) and ii), we instead propose to employ \emph{\de{} density
estimation} over $\Z$. We fit a density estimator denoted as $\priorde$ to $\{\z
= \En(\x)|\x\in\mathcal{X}\}$. This simple approach not only fits our RAE
framework well, but it can also be readily adopted for any VAE or variants
thereof such as the WAE as a practical remedy to the aggregated posterior
mismatch without adding any computational overhead to the costly training phase.

The choice of $\priorde$ needs to trade-off \emph{expressiveness} -- to provide
a good fit of an arbitrary space for $\Z$ -- with \emph{simplicity}, to
improve generalization. For example, placing a Dirac distribution on each latent
point $\z$ would allow the decoder to output only training sample
reconstructions which have a high quality, but do not generalize.
Striving for simplicity, 
we employ and compare a full covariance
multivariate Gaussian with a 10-component Gaussian mixture model (GMM) in our
experiments.



\section{Related works}
\vspace{-1mm}
\label{sec:relatedworks}

Many works have focused on diagnosing the VAE framework,  the terms in its
objective~\citep{hoffman2016elbo, zhao2017towards, Alemi2017}, and ultimately
augmenting it to solve optimization issues~\citep{Rezende2018, twostagevae}. With
RAE, we argue that a simpler deterministic framework can be competitive for
generative modeling.

Deterministic denoising~\citep{vincent2008extracting} and contractive
autoencoders (CAEs)~\citep{rifai2011contractive} have received attention in the
past for their ability to capture a smooth data manifold. Heuristic attempts to
equip them with a generative mechanism include MCMC schemes~\citep{Rifai2012,
Bengio2013}. However, they are hard to diagnose for convergence, require a
considerable effort in tuning~\citep{cowles1996markov}, and have not scaled
beyond MNIST, leading to them being superseded by VAEs.
While computing the Jacobian for CAEs~\citep{rifai2011contractive} is
close in spirit to $\lossgp$ for RAEs, the latter is much more
computationally efficient.

Approaches to cope with the aggregated posterior mismatch involve fixing a more
expressive form for $\prior$~\citep{kingma1606improving, BauMni18} therefore
altering the VAE objective and requiring considerable additional computational
efforts. Estimating the latent space of a VAE with a second VAE~\citep{twostagevae}
reintroduces many of the optimization shortcomings discussed for VAEs and is
much more expensive in practice compared to fitting a simple $\priorde$ after
training.


Adversarial Autoencoders (AAE)~\citep{Makhzani2015} add a discriminator to a
deterministic encoder--decoder pair, leading to sharper samples at the expense
of higher computational overhead and the introduction of instabilities caused by
the adversarial nature of the training process.

\begin{figure*}[!th]
  \centering
  \scalebox{.9}
{\setlength\tabcolsep{3pt}
\begin{sc}
  \small
    \begin{tabular}{ r c c c}
    \toprule
        & Reconstructions & Random Samples & Interpolations \\
        \midrule
      \raisebox{10pt}{GT}
      & \includegraphics[trim={ 0 576 0
        0},clip,width=0.32\linewidth]{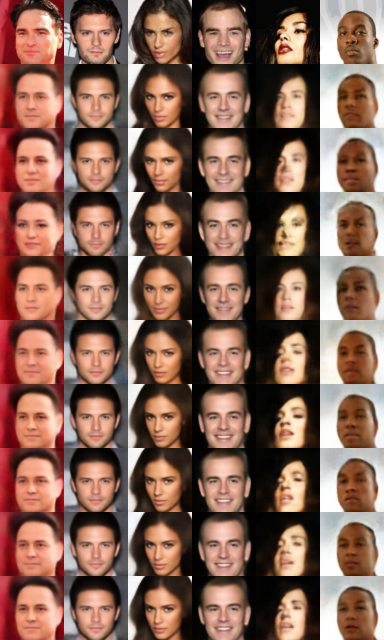}
      \\[-3.5pt]

      \raisebox{10pt}{VAE}
      & \includegraphics[trim={ 0 512 0 64},clip,width=0.32\linewidth]{figs/pics_new/CELEBA_recons.png}&\includegraphics[trim={ 0 512 0 64},clip,width=0.32\linewidth]{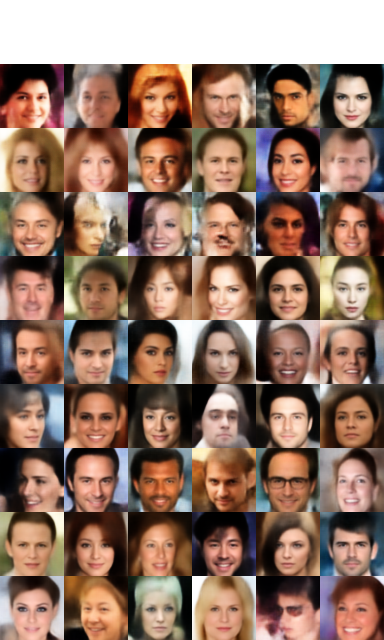} & \includegraphics[trim={ 0 512 0 64},clip,width=0.32\linewidth]{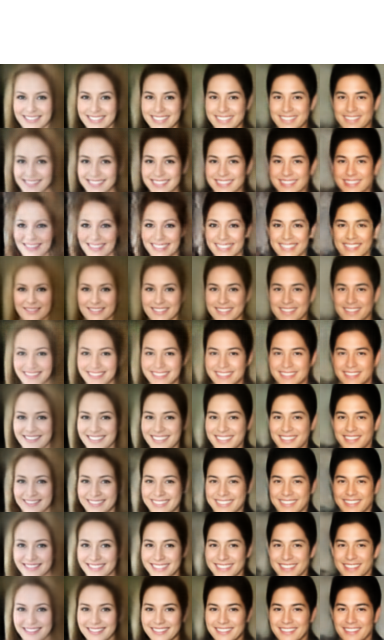} \\[-3.5pt]

      \raisebox{10pt}{CV-VAE}
      & \includegraphics[trim={ 0 448 0 128},clip,width=0.32\linewidth]{figs/pics_new/CELEBA_recons.png}&\includegraphics[trim={ 0 448 0 128},clip,width=0.32\linewidth]{figs/pics_new/CELEBA_random.png} & \includegraphics[trim={ 0 448 0 128},clip,width=0.32\linewidth]{figs/pics_new/CELEBA_interp.png} \\[-3.5pt]

      \raisebox{10pt}{WAE}
      & \includegraphics[trim={ 0 384 0 192},clip,width=0.32\linewidth]{figs/pics_new/CELEBA_recons.png}&\includegraphics[trim={ 0 384 0 192},clip,width=0.32\linewidth]{figs/pics_new/CELEBA_random.png} & \includegraphics[trim={ 0 384 0 192},clip,width=0.32\linewidth]{figs/pics_new/CELEBA_interp.png} \\[-3.5pt]

      \raisebox{10pt}{2sVAE}
      & \includegraphics[trim={ 0 320 0 256},clip,width=0.32\linewidth]{figs/pics_new/CELEBA_recons.png}&\includegraphics[trim={ 0 320 0 256},clip,width=0.32\linewidth]{figs/pics_new/CELEBA_random.png} & \includegraphics[trim={ 0 320 0 256},clip,width=0.32\linewidth]{figs/pics_new/CELEBA_interp.png} \\[-3.5pt]

      \raisebox{10pt}{RAE-GP}
      & \includegraphics[trim={ 0 256 0 320},clip,width=0.32\linewidth]{figs/pics_new/CELEBA_recons.png}&\includegraphics[trim={ 0 256 0 320},clip,width=0.32\linewidth]{figs/pics_new/CELEBA_random.png} & \includegraphics[trim={ 0 256 0 320},clip,width=0.32\linewidth]{figs/pics_new/CELEBA_interp.png} \\[-3.5pt]

      \raisebox{10pt}{RAE-L2}
      & \includegraphics[trim={ 0 192 0 384},clip,width=0.32\linewidth]{figs/pics_new/CELEBA_recons.png}&\includegraphics[trim={ 0 192 0 384},clip,width=0.32\linewidth]{figs/pics_new/CELEBA_random.png} & \includegraphics[trim={ 0 192 0 384},clip,width=0.32\linewidth]{figs/pics_new/CELEBA_interp.png} \\[-3.5pt]

      \raisebox{10pt}{RAE-SN}
      & \includegraphics[trim={ 0 128 0 448},clip,width=0.32\linewidth]{figs/pics_new/CELEBA_recons.png}&\includegraphics[trim={ 0 128 0 448},clip,width=0.32\linewidth]{figs/pics_new/CELEBA_random.png} & \includegraphics[trim={ 0 128 0 448},clip,width=0.32\linewidth]{figs/pics_new/CELEBA_interp.png} \\[-3.5pt]

      \raisebox{10pt}{RAE}
      & \includegraphics[trim={ 0 64 0 512},clip,width=0.32\linewidth]{figs/pics_new/CELEBA_recons.png}&\includegraphics[trim={ 0 64 0 512},clip,width=0.32\linewidth]{figs/pics_new/CELEBA_random.png} & \includegraphics[trim={ 0 64 0 512},clip,width=0.32\linewidth]{figs/pics_new/CELEBA_interp.png} \\[-3.5pt]

      \raisebox{10pt}{AE}
      & \includegraphics[trim={ 0 0 0 576},clip,width=0.32\linewidth]{figs/pics_new/CELEBA_recons.png}&\includegraphics[trim={ 0 0 0 576},clip,width=0.32\linewidth]{figs/pics_new/CELEBA_random.png} & \includegraphics[trim={ 0 0 0 576},clip,width=0.32\linewidth]{figs/pics_new/CELEBA_interp.png} \\

      \bottomrule

    \end{tabular}
\end{sc}}
  \vspace{2.5mm}
  \caption{Qualitative evaluation of sample quality for VAEs, WAEs, 2sVAEs, and
  RAEs on CelebA. RAE provides slightly sharper samples and reconstructions
  while interpolating smoothly in the latent space. Corresponding qualitative
  overviews for MNIST and CIFAR-10 are provided in
  Appendix~\ref{sec:more-pics}.}
  \label{fig:celeba-pics}
\end{figure*}

Wasserstein Autoencoders
(WAE) ~\citep{Tolstikhin2017} have been introduced as a generalization of AAEs by
casting autoencoding as an optimal transport (OT) problem. Both stochastic and
deterministic models can be trained by minimizing a relaxed OT cost function
employing either an adversarial loss term or the maximum mean discrepancy score
between $\prior$ and $q_{\phi}(\z)$ as a regularizer in place of $\losskl$.
Within the RAE framework, we look at this problem from a different perspective:
instead of explicitly imposing a simple structure on $\Z$ that might impair the
ability to fit high-dimensional data during training, we propose to model the
latent space by an ex-post density estimation step.

The most successful VAE architectures for images and audio so far are variations
of the VQ-VAE~\citep{van2017neural, razavi2019generating}. Despite the name,
VQ-VAEs are neither stochastic, nor variational, but they are deterministic
autoencoders. VQ-VAEs are similar to RAEs in that they adopt ex-post density
estimation. However, VQ-VAEs necessitates complex discrete autoregressive
density estimators and a training loss that is non-differentiable due to
quantizing $\Z$.

Lastly, RAEs share some similarities with
GLO~\citep{bojanowski2017optimizing}.
However, differently from RAEs, GLO can be interpreted as a deterministic AE without and
encoder, and when the latent space is built ``on-demand'' by
optimization.
On the other hand, RAEs augment deterministic decoders as in GANs with
deterministic encoders.



\vspace{-1mm}
\section{Experiments}
\vspace{-1mm}
\label{sec:exps}


%
Our experiments are designed to answer the following questions: \textbf{Q1:} Are
sample quality and latent space structure in RAEs comparable to VAEs?
\textbf{Q2:} How do different regularizations impact RAE performance?
\textbf{Q3:} What is the effect of ex-post density estimation on VAEs and its
variants?
%

\begin{table*}[!th]
  \setlength{\tabcolsep}{5pt}
  \centering
  \begin{sc}
    \scriptsize
    \begin{tabular}{l l r r r r r r r r r r r r r r}
      \toprule
       & \multicolumn{4}{c}{MNIST} & \multicolumn{4}{c}{CIFAR} & \multicolumn{4}{c}{CelebA} \\
    \cmidrule(r){2-5}\cmidrule(r){6-9}\cmidrule(r){10-13}
      & \multirow{2}[3]{*}{Rec.} & \multicolumn{3}{c}{Samples} & \multirow{2}[3]{*}{Rec.} & \multicolumn{3}{c}{Samples} & \multirow{2}[3]{*}{Rec.} & \multicolumn{3}{c}{Samples}\\
      \cmidrule(lr){3-5} \cmidrule(lr){7-9}\cmidrule(lr){11-13}
       & &  $\mathcal{N}$ & $\mathsf{GMM}$ & $\mathsf{Interp.}$& &  $\mathcal{N}$ & $\mathsf{GMM}$&$\mathsf{Interp.}$& &  $\mathcal{N}$ & $\mathsf{GMM}$&$\mathsf{Interp.}$\\
      \cmidrule(r){2-5}\cmidrule(r){6-9}\cmidrule(r){10-13}
      VAE                   & 18.26 & {19.21} & 17.66  & 18.21 & 57.94 & 106.37 & 103.78 & 88.62 & 39.12 & {48.12} & 45.52 & 44.49\\
      CV-VAE                & 15.15 & 33.79 & 17.87 & 25.12 & 37.74 & 94.75 & 86.64 & 69.71 & 40.41 & 48.87 & 49.30 & 44.96\\
      WAE                   & \textbf{10.03} & 20.42 & {9.39} & \textbf{14.34} & 35.97 & 117.44 & 93.53 & 76.89 & \textbf{34.81} & 53.67 & {42.73} & {40.93}\\
      2sVAE                 & 20.31 & \textbf{18.81} & -- & 18.35 & 62.54 & 109.77 & -- & 89.06 & 42.04 & 49.70 & -- & 47.54 \\
      \midrule
      \ourmodel{}-GP        & 14.04 & 22.21 & 11.54 &  15.32 & 32.17 & {83.05} & 76.33 & 64.08 & 39.71 & 116.30 & 45.63 & 47.00 \\
      \ourmodel{}-L2        & {10.53} & 22.22 & \textbf{8.69} & {14.54} & 32.24 & \textbf{80.80} & \textbf{74.16} & {62.54} & 43.52 & 51.13 & 47.97 & 45.98 \\
      \ourmodel{}-SN        & 15.65 & {19.67} & 11.74 & 15.15 & \textbf{27.61} & 84.25 & {75.30} & 63.62 & {36.01} & \textbf{44.74} & \textbf{40.95} & \textbf{39.53}\\
      \ourmodel{}           & 11.67 & 23.92 & 9.81 & 14.67 & {29.05} & 83.87 & 76.28 & 63.27 & 40.18 & 48.20 & 44.68 & 43.67\\
      AE                    & 12.95 & 58.73 & 10.66 & 17.12 & {30.52} & 84.74 & 76.47 & \textbf{61.57} & 40.79 & 127.85 & 45.10 & 50.94\\
      AE-L2 &  11.19 & 315.15 & 9.36 &  17.15 & 34.35 & 247.48 & 75.40 & 61.09 & 44.72 & 346.29 & 48.42 & 56.16 & \\
      \bottomrule
  \end{tabular}
  \end{sc}
  \vspace{2.5mm}
  \caption{
  Evaluation of all models by FID (lower is better, best models in bold). We
  evaluate each model by
  \textsc{Rec.}: test sample reconstruction;
  $\mathcal{N}$: random samples generated according to the prior distribution
  $\prior$ (isotropic Gaussian for VAE / WAE, another VAE for 2SVAE) or by fitting a Gaussian to $\priorde$ (for the
  remaining models);
  $\mathsf{GMM}$: random samples generated by fitting a mixture of 10 Gaussians
  in the latent space;
  $\mathsf{Interp.}$: mid-point interpolation between random pairs of test
  reconstructions.
  The RAE models are competitive with or outperform previous models throughout
  the evaluation. Interestingly, interpolations do not suffer from the lack of
  explicit priors on the latent space in our models.
  %
  }
  \label{tab:fids-main}
\end{table*}

\subsection{RAEs for image modeling}

We evaluate all regularization schemes from Section~\ref{sec:raereg}: RAE-GP,
RAE-L2, and RAE-SN.
For a thorough ablation study, we also consider only adding the latent code
regularizer $\lossraekl$ to $\lossr$ (RAE), and an autoencoder without any
explicit regularization (AE).
We check the effect of applying one regularization scheme while not
including the $\lossraekl$ term in the AE-L2 model.
As baselines, we employ the regular VAE, constant-variance VAE (CV-VAE),
Wasserstein Autoencoder (WAE) with the MMD loss as a state-of-the-art method,
and the recent 2-stage VAE (2sVAE)~\citep{twostagevae} which performs a form of
ex-post density estimation via another VAE.
For a fair comparison, we use the same network architecture for all models.
Further details about the architecture and training are given in
Appendix~\ref{sec:architecture-training}.

We measure the following quantities: held-out sample reconstruction quality,
random sample quality, and interpolation quality.
While reconstructions give us a lower bound on the best quality achievable by
the generative model, random sample quality indicates how well the model
generalizes. Finally, interpolation quality sheds light on the structure of the
learned latent space.
The evaluation of generative models is a nontrivial research
question~\citep{Theis2015, enhancenet, comparegan}. We report here the
ubiquitous Fr\'echet Inception Distance (FID)~\citep{fid} and we provide
precision and recall scores (PRD)~\citep{prd} in Appendix~\ref{sec:prd-eval}.
%


Table~\ref{tab:fids-main} summarizes our main results. All of the proposed RAE
variants are competitive with the VAE, WAE and 2sVAE \wrt generated image quality in
all settings. Sampling RAEs achieve the best FIDs across all datasets when a
modest 10-component GMM is employed for ex-post density estimation. Furthermore,
even when $\mathcal{N}$ is considered as $\priorde$, RAEs rank first with the
exception of MNIST, where it competes for the second position with a
VAE.
%
%
Our best RAE FIDs are lower than the best results reported for VAEs in
the large scale comparison of~\citep{comparegan}, challenging even the best
scores reported for GANs.
While we are employing a slightly different
architecture than theirs, our models underwent only modest finetuning instead of
an extensive hyperparameter search.
A comparison of the different regularization schemes for RAEs (Q2) yields no clear winner across all settings as all perform equally well.
Striving for a simpler implementation, one may prefer RAE-L2 over the GP and SN
variants.

For completeness, we investigate applying multiple regularization
schemes to our RAE models.
We report the results of all possible combinations in Table \ref{tab:fids-aux}, Appendix
\ref{sec:more-regs}.
There, no significant boost of performance can be spotted when
comparing to singly regularized RAEs.
%


Surprisingly, the implicitly regularized RAE and AE models are shown to be able
to score impressive FIDs when $\priorde$ is fit through GMMs. FIDs for AEs
decrease from 58.73 to 10.66 on MNIST and from 127.85 to 45.10 on CelebA -- a
value close to the state of the art. This is a remarkable result that follows a
long series of recent confirmations that neural networks are surprisingly smooth
by design~\citep{neyshabur2017geometry}. It is also surprising that the
lack of an explicitly fixed structure on the latent space of the RAE does not
impede interpolation quality. This is further confirmed by the qualitative
evaluation on CelebA as reported in Fig.~\ref{fig:celeba-pics} and for the other
datasets in Appendix~\ref{sec:more-pics}, where RAE interpolated samples seem
sharper than competitors and transitions smoother.
%

Our results further confirm and quantify the effect of the aggregated posterior
mismatch. In Table~\ref{tab:fids-main}, ex-post density estimation consistently
improves sample quality across all settings and models. A 10-component GMM
halves FID scores from ${\sim}20$ to ${\sim}10$ for WAE and RAE models on MNIST
and from 116 to 46 on CelebA. This is especially striking since this additional
step is much cheaper and simpler than training a second-stage VAE as in 2sVAE
(Q3). In summary, the results strongly support the conjecture that the
simple deterministic RAE framework can challenge VAEs and stronger alternatives
(Q1).

\begin{figure}[!t]
  \centering
  \begin{subfigure}[t]{0.3\textwidth}
    \vspace{-95pt}
  \setlength{\tabcolsep}{3pt}
  \centering
  \begin{sc}
    \scriptsize
    \begin{tabular}{l l l r}
      \toprule
      {Problem} & {Model} & {\% Valid} & {Avg. score} \\
      \midrule
      \multirow{3}{*}{Expressions}
       & GRAE & \textbf{1.00 }$\pm$ \textbf{0.00} &  {3.22} $\pm$ {0.03} \\
      & GCVVAE & 0.99 $\pm$ 0.01 &  \textbf{2.85} $\pm$ \textbf{0.08} \\
       & GVAE & 0.99 $\pm$ 0.01 & 3.26 $\pm$ 0.20 \\
       & CVAE & 0.82 $\pm$ 0.07 & 4.74 $\pm$ 0.25 \\
       \cmidrule(r){1-4}
      \multirow{3}{*}{Molecules}
      & GRAE & 0.72 $\pm$ 0.09 & \textbf{-5.62} $\pm$ \textbf{0.71}\\
      & GCVVAE & \textbf{0.76} $\pm$ \textbf{0.06} &  -6.40 $\pm$ 0.80\\
       & GVAE & 0.28 $\pm$ 0.04 & -7.89 $\pm$ 1.90 \\
       & CVAE & 0.16 $\pm$ 0.04 & -25.64 $\pm$ 6.35 \\
      \bottomrule
    \end{tabular}\vspace{10pt}
    \begin{tabular}{l l l l}
      \toprule
      {Model} & {\#} & {Expression} & {Score} \\
      \cmidrule(r){1-4}

      \multirow{3}{*}{GRAE} & 1 & $\sin(3)+x$ & 0.39 \\
      & 2 & $x+1/\exp(1)$ & \textbf{0.39} \\
      & 3 & $x+1+2*\sin(3+1+2)$ & \textbf{0.43} \\
      \cmidrule(r){1-4}
      \multirow{3}{*}{GCVVAE} & 1 & $x+\sin(3)*1$ & {0.39} \\
      & 2 & $x/x/3+x$ & 0.40 \\
      & 3 & $\sin(\exp(\exp(1)))+x/2*2$ & \textbf{0.43} \\
      \cmidrule(r){1-4}
      \multirow{3}{*}{GVAE} & 1 & $x/1+\sin(x)+\sin(x*x)$ & \textbf{0.10}\\
      & 2 & $1/2+(x)+\sin(x*x)$ & 0.46 \\
      & 3 & $x/2+\sin(1)+(x/2)$ & 0.52 \\
      \cmidrule(r){1-4}

      \multirow{3}{*}{CVAE} & 1 & $x*1+\sin(x)+\sin(3+x)$ & 0.45\\
      & 2 & $x/1+\sin(1)+\sin(2*2)$ & 0.48\\
      & 3 & $1/1+(x)+\sin(1/2)$ & 0.61\\
      \bottomrule
  \end{tabular}
  \end{sc}
  \label{tab:eq-mol_comp}
  \end{subfigure}\hspace{60pt}
\begin{subfigure}[t]{0.5\textwidth}
  \setlength{\tabcolsep}{2pt}
  \centering
  \begin{sc}
    \scriptsize
    \begin{tabular}{l c c c}
      \toprule
      {Model} & 1st & 2nd & 3rd \\
      \cmidrule(r){1-4}
      \raisebox{10pt}{GRAE} & \includegraphics[trim={ 0 50 450 50}, clip, width=0.25\linewidth]{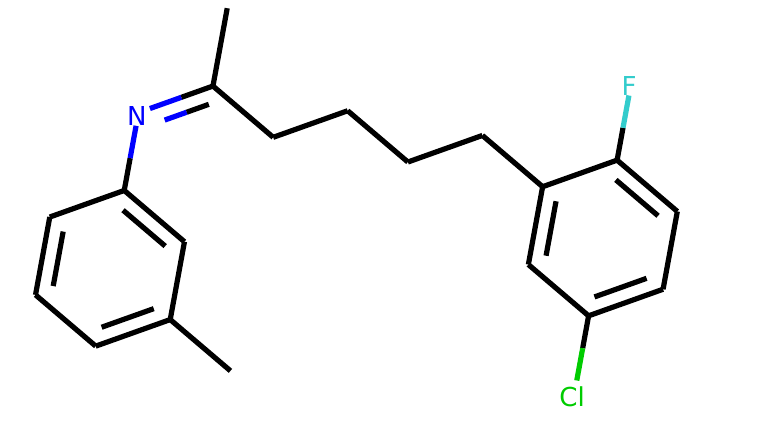}&\includegraphics[trim={ 200 50 200 50}, clip, width=0.25\linewidth]{figs/pics/eq_mol_bo/best_grammar_rae5_molecule}&\includegraphics[trim={ 450 50 0 50}, clip, width=0.25\linewidth]{figs/pics/eq_mol_bo/best_grammar_rae5_molecule}\\[5pt]
      Score & \textbf{3.74} & \textbf{3.52} & \textbf{3.14}\\[5pt]
      
      \raisebox{10pt}{GCVVAE} & \includegraphics[trim={ 0 50 450 50}, clip, width=0.25\linewidth]{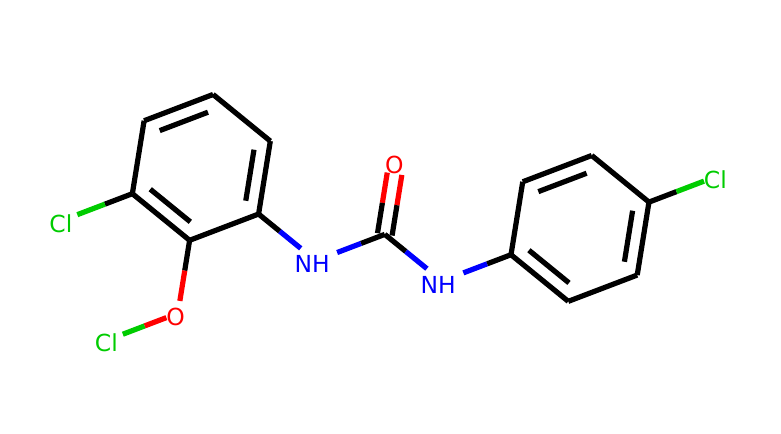}&\includegraphics[trim={ 230 50 230 50}, clip, width=0.25\linewidth]{figs/pics/eq_mol_bo/best_grammar_cv_vae2_molecule.pdf}&\includegraphics[trim={ 450 50 0 50}, clip, width=0.25\linewidth]{figs/pics/eq_mol_bo/best_grammar_cv_vae2_molecule.pdf}\\[5pt]
      Score & 3.22 & 2.83 & 2.63 \\[5pt]
      
      \raisebox{10pt}{GVAE} & \includegraphics[trim={ 0 50 450 50}, clip, width=0.25\linewidth]{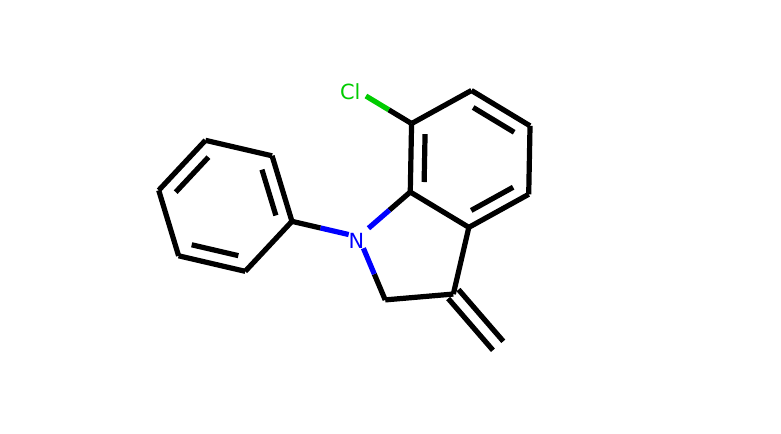}&\includegraphics[trim={ 200 50 250 50}, clip, width=0.25\linewidth]{figs/pics/eq_mol_bo/best_grammar_molecule}&\includegraphics[trim={ 450 50 0 50}, clip, width=0.25\linewidth]{figs/pics/eq_mol_bo/best_grammar_molecule}\\[0pt]
      Score & 3.13 & 3.10 & 2.37\\[5pt]
      \raisebox{10pt}{CVAE} & \includegraphics[trim={ 0 80 450 60}, clip, width=0.25\linewidth]{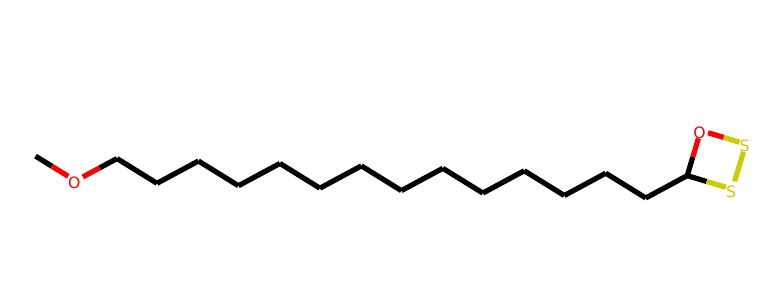}&\includegraphics[trim={ 250 90 250 70}, clip, width=0.25\linewidth]{figs/pics/eq_mol_bo/best_character_molecule}&\includegraphics[trim={ 450 60 0 50}, clip, width=0.25\linewidth]{figs/pics/eq_mol_bo/best_character_molecule}\\[2pt]
      Score & 2.75 & 0.82 & 0.63\\
      \bottomrule
  \end{tabular}
  \end{sc}
  \label{tab:eq-mol_comp2}
\end{subfigure}
\caption{Generating structured objects by GVAE,
    CVAE and GRAE. (Upper left) Percentage of
    valid samples and their average mean score (see text, Section \ref{sec:GrammarRAE}).
    The three best expressions (lower left) and molecules (upper
    right) and their scores are reported for all models.
  }
  \label{fig:molecules-eq}
\end{figure}

\subsection{GrammarRAE: modeling structured inputs}
\label{sec:GrammarRAE}

We now evaluate RAEs for generating complex structured objects such as molecules
and arithmetic expressions.
We do this with a twofold aim: i) to investigate the latent space learned by
RAE for more challenging input spaces that abide to some structural
constraints, and
ii) to quantify the gain of replacing the VAE in a state-of-the-art generative model with a RAE.

To this end, we adopt the exact architectures and experimental settings of the
GrammarVAE (GVAE)~\citep{kusner2017grammar}, which has been shown to outperform
other generative alternatives such as the CharacterVAE
(CVAE)~\citep{gomez2018automatic}.
As in \citet{kusner2017grammar}, we are interested in traversing the latent
space learned by our models to generate samples (molecules or expressions) that
best fit some downstream metric.
This is done by Bayesian optimization (BO) by considering the $\log(1 +
\mathsf{MSE})$ (lower is better) for the generated expressions \wrt some ground
truth points, and the water-octanol partition coefficient ($\log
P$)~\citep{pyzer2015high} (higher is better) in the case of molecules.
A well-behaved latent space will not only generate molecules or expressions with
better scores during the BO step, but it will also contain syntactically valid
ones, \ie, samples abide to a grammar of rules describing the problem.
%

Figure~\ref{fig:molecules-eq} summarizes our results over 5 trials of BO.
Our GRAEs (Grammar RAE) achieve better average scores than CVAEs and GVAEs in
generating expressions and molecules.
This is visible also for the three best samples and their scores for
all models, with the exception of the first best expression of GVAE.
We include in the comparison also the GCVVAE, the equivalent of a
CV-VAE for structured objects, as an additional baseline.
We can observe that while the GCVVAE delivers better average scores
for the simpler task of generating equations (even though the single
three best equations are on par with GRAE), when generating molecules
GRAEs deliver samples associated to much higher scores.

More interestingly, while GRAEs are almost equivalent to GVAEs for the easier
task of generating expressions, the proportion of syntactically valid molecules
for GRAEs greatly improves over GVAEs (from 28\% to 72\%).

\section{Conclusion}
\label{sec:conclusion}

While the theoretical derivation of the VAE has helped popularize the framework
for generative modeling, recent works have started to expose some discrepancies
between theory and practice. We have shown that viewing sampling in VAEs as
noise injection to enforce smoothness can enable one to distill a deterministic
autoencoding framework that is compatible with several regularization techniques
to learn a meaningful latent space. We have demonstrated that such an
autoencoding framework can generate comparable or better samples than VAEs while
getting around the practical drawbacks tied to a stochastic framework.
Furthermore, we have shown that our solution of fitting a simple density
estimator on the learned latent space consistently improves sample quality both
for the proposed RAE framework as well as for VAEs, WAEs, and 2sVAEs which
solves the mismatch between the prior and the aggregated posterior in VAEs.



\section*{Acknowledgements} We would like to thank Anant Raj, Matthias Bauer,
Paul Rubenstein and Soubhik Sanyal for fruitful discussions.

\scalebox{0.01}{variatio delectat!}

\bibliography{rae}
\bibliographystyle{iclr2020_conference}


\clearpage
\appendix

\section*{Appendix}
\label{sec:appendix}

\section{Reconstruction and regularization trade-off}
\label{sec:rec-reg}

\begin{figure}[!htbp]
  \centering
  \includegraphics[trim={0 0 0
    0},clip,width=.45\linewidth]{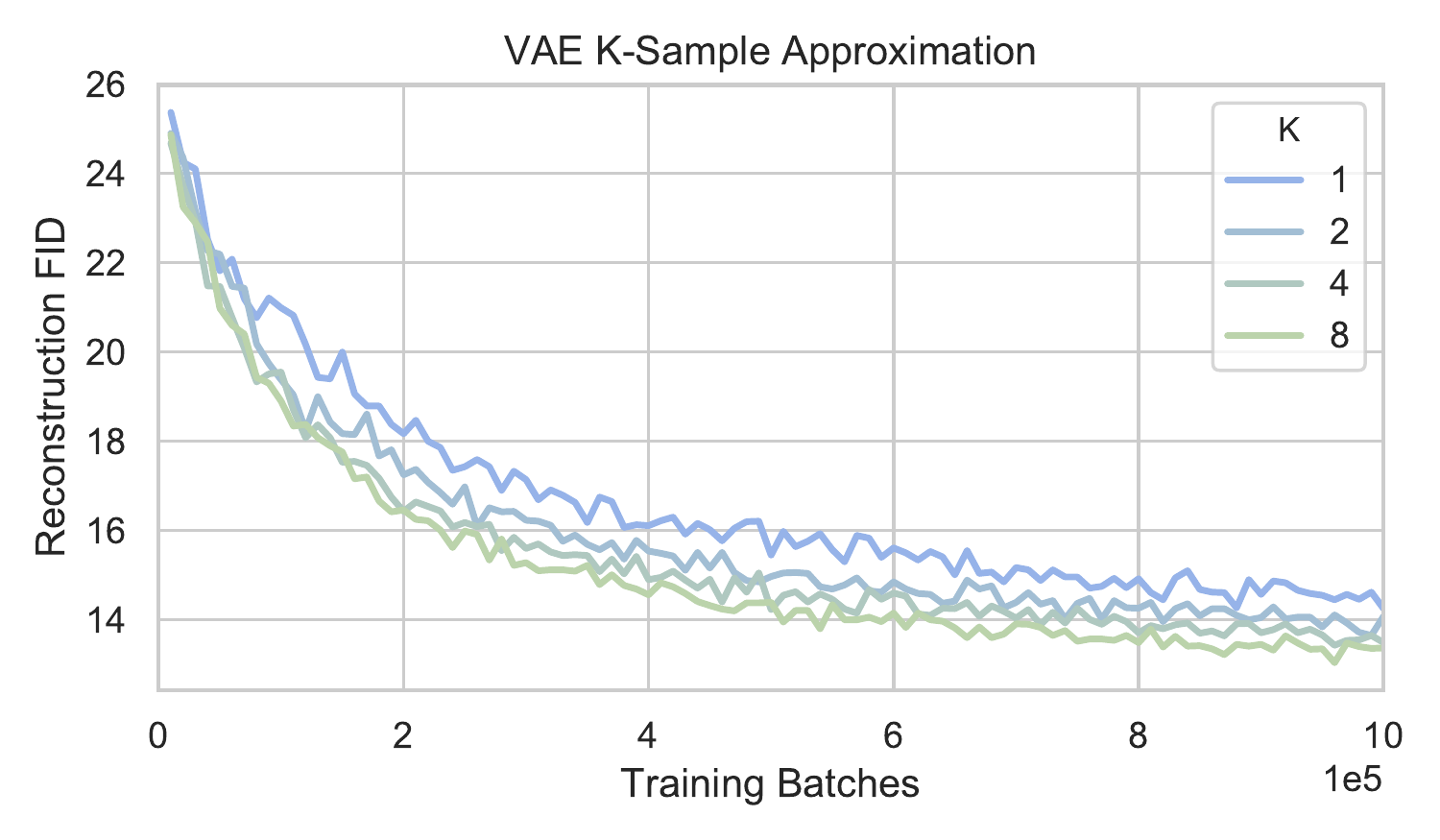}
  \includegraphics[trim={0 0 0 0},clip,width=.45\linewidth]{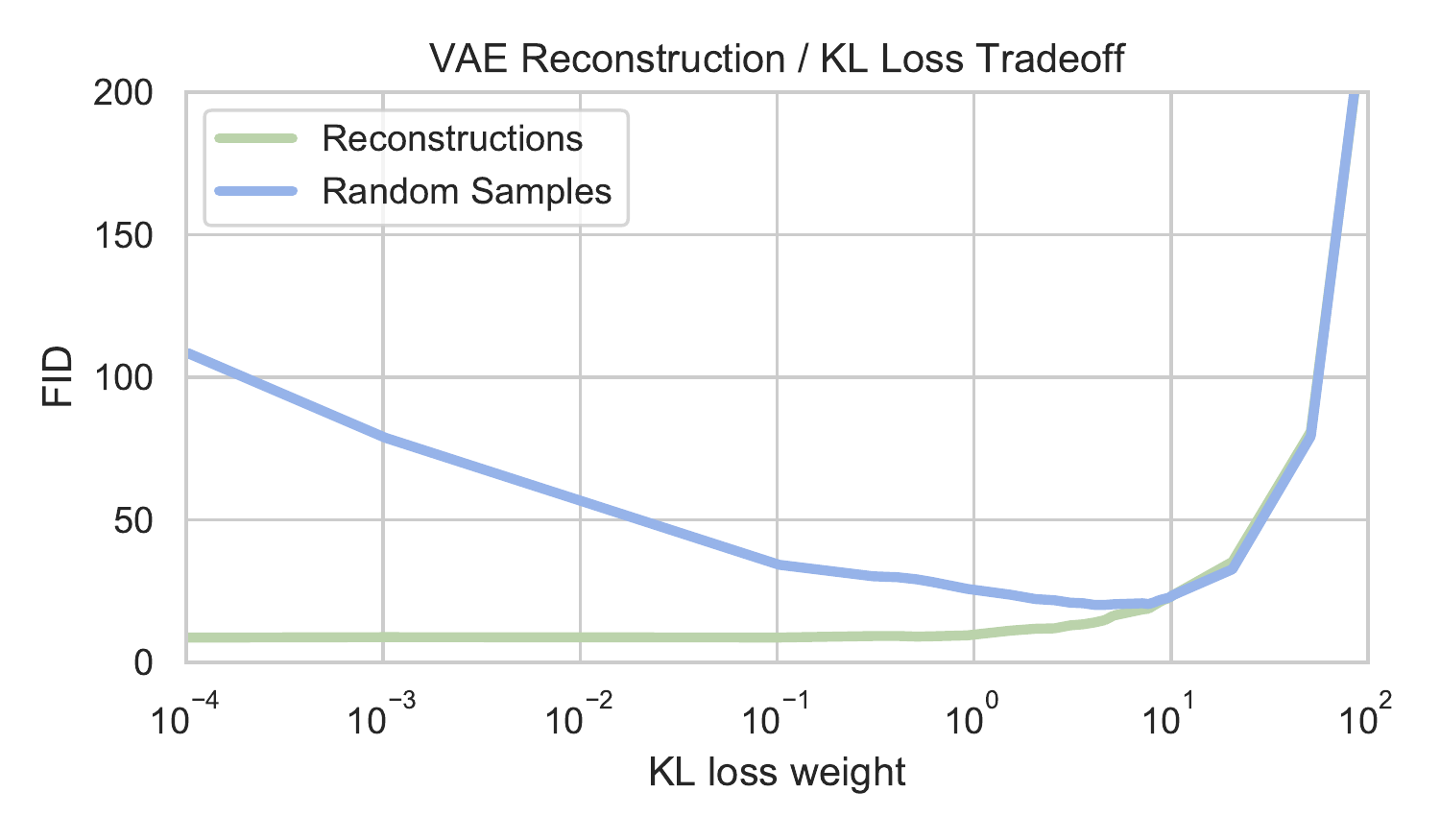}
  \caption{(Left) Test reconstruction quality for a VAE trained on MNIST with different
  numbers of samples in the latent space as in Eq.~\ref{eq:vae-sampling-encoder}
  measured by FID (lower is better). Larger numbers of Monte-Carlo samples clearly improve
  training, however, the increased accuracy comes with larger requirements for
  memory and computation. In practice, the most common choice is therefore
  $k=1$.
  (Right) Reconstruction and random sample quality (FID, y-axis, lower is
  better) of a VAE on MNIST for different trade-offs between $\lossr$ and
  $\losskl$ (x-axis, see Eq.~\ref{eq:elbo-loss}). Higher weights for $\losskl$
  improve random samples but hurt reconstruction. This is especially noticeable towards the optimality point ($\beta \approx 10^1$). This indicates that enforcing structure in the VAE latent space leads to a penalty in quality.
}
  \vspace{-2.5mm}
  \label{fig:k-step-approx}
\end{figure}

We train a VAE on MNIST while monitoring the test set reconstruction quality by
FID. Figure~\ref{fig:k-step-approx} (left) clearly shows the impact of more
expensive $k > 1$ Monte Carlo approximations of
Eq.~\ref{eq:vae-sampling-encoder} on sample quality during training. The
commonly used 1-sample approximation is a clear limitation for VAE training.

Figure~\ref{fig:k-step-approx} (right) depicts the inherent trade-off between
reconstruction and random sample quality in VAEs. Enforcing structure and
smoothness in the latent space of a VAE affects random sample quality in a
negative way. In practice, a compromise needs to be made, ultimately leading to
subpar performance.


\section{A Probabilistic Derivation of Regularization}
\label{sec:method}

In this section, we propose an alternative view on enforcing smoothness on the
output of $\D$ by augmenting the ELBO optimization problem for VAEs with an
explicit constraint. While we keep the Gaussianity assumptions over a stochastic
$\D$ and $\prior$ for convenience, we however are not fixing a parametric form for
$\post$ yet. We discuss next how some parametric restrictions over $\post$
lead to a variation of the RAE framework in Eq.~\ref{eq:rae-loss},
specifically the introduction of $\lossgp$ as a regularizer of a
deterministic version of the CV-VAE.
To start, we  augment Eq.~\ref{eq:elbo-loss} as:
\begin{align}
  \label{eq:const_opt}
    \argmin_{\phi,\theta}\;
    &\EX_{\x\sim p_{\mathsf{data}}(\X)}\;
    \mathcal{L}_{\mathsf{REC}} +
    \mathcal{L}_{\mathsf{KL}}\\\nonumber
    \text{s.t. } ||\D(\z_1) - \D(\z_2)||_p &< \epsilon \quad
    \forall \; \z_1, \z_2 \sim \post \quad\forall \x \sim \pdata
\end{align}
where $\D(\z)=\boldsymbol\mu_{\theta}(E_\phi(\x))$ and the constraint on the
decoder encodes that the output has to vary, in the sense of an $L_{p}$ norm,
only by a small amount $\epsilon$ for any two possible draws from the encoding of $\x$ .
Let $\D(\z):\mathbb{R^{\dim(\z)}}\rightarrow\mathbb{R^{\dim(\x)}}$ be given by a set of $\dim(\x)$ given by $\{d_i(\z):\mathbb{R^{\dim(\z)}}\rightarrow\mathbb{R}^1\}$. Now we can upper bound the quantity $||\D(\z_1) - \D(\z_2)||_p$ by $\dim(\x)*sup_i\{||d_i(\z_1) - d_i(\z_2)||_p\}$.  Using mean value theorem $||d_i(\z_1) - d_i(\z_2)||_p \leq ||\nabla_t d_i((1-t)\z_1 + t\z_2)||_p*||\z_1-\z_2||_p$. Hence $sup_i\{||d_i(\z_1) - d_i(\z_2)||_p\} \leq sup_i\{||\nabla_t d_i((1-t)\z_1 + t\z_2)||_p*||\z_1-\z_2||_p\}$. Now if we choose the domain of $\post$ to be isotopic the contribution of $||\z_2-\z_1||_p$ to the afore mentioned quantity becomes a constant factor. Loosely speaking it is the radios of the bounding ball of domain of $\post$. Hence the above term simplifies to $sup_i\{||\nabla_t d_i((1-t)\z_1 + t\z_2)||_p\}$. Recognizing that here $\z_1$ and $\z_2$ is arbitrary lets us simplify this further to $sup_i\{||\nabla_z d_i(\z)||_p\}$

From this form of the smoothness
constraint, it is apparent why the choice of a parametric form for $\post$ can
be impactful during training. For a compactly supported isotropic PDF
$q_\phi(\z|\x)$, the extension of the support $sup\{||\z_1 - \z_2||_p\}$ would
depend on its entropy $\HE(\post)$.
through some functional
$r$. For instance, a uniform posterior over a hypersphere in $\z$ would
ascertain $r(\HE(\post)) \cong e^{\HE(\post)/n}$ where $n$ is the dimensionality
of the latent space.

Intuitively, one would look for parametric distributions that do not favor
overfitting, \eg degenerating in Dirac-deltas (minimal entropy and support)
along any dimensions. To this end, an isotropic nature of $q_\phi(\z|\x)$ would
favor such a robustness against decoder over-fitting. We can now rewrite the constraint as
\begin{equation}
  r(\HE(\post)) \cdot sup\{||\nabla \D(\z)\||_p\} < \epsilon
  \label{final_opt_practical}
\end{equation}
The $\losskl$ term can be expressed in terms of $\HE(\post)$, by decomposing it
as $\losskl = \lossce - \lossh$, where $\lossh = \HE(\post)$ and
$\lossce=\HE(\post, \prior)$ represents a cross-entropy term.
Therefore, the constrained problem in Eq.~\ref{eq:const_opt} can be written in a
Lagrangian formulation by including Eq.~\ref{final_opt_practical}:
\begin{equation}
  \begin{split}
    \argmin_{\phi,\theta}\; \EX_{\x\sim p_{\mathsf{data}}}\; \lossr + \lossce - \lossh +\lambda\losslang \\
  \end{split}
    \label{eq:const_opt_lagrange}
\end{equation}
where $\losslang=r(\HE(\post))*||\nabla \D(\z)||_p$. We argue that a reasonable
simplifying assumption for $\post$ is to fix $\HE(\post)$ to a single constant
for all samples $\x$. Intuitively, this can be understood as fixing the variance
in $\post$ as we did for the CV-VAE in Section~\ref{sec:cvvae}. With this simplification, Eq.~\ref{eq:const_opt_lagrange} further reduces to
\begin{equation}
  \begin{split}
    \argmin_{\phi,\theta}\; \EX_{\x\sim p_{\mathsf{data}}(\X)}\; \lossr + \lossce +\lambda||\nabla \D(\z)||_{p} \\
  \end{split}
  \label{eq:const-ent}
\end{equation}
We can see that $||\nabla \D(\z)||_{p}$ results to be the gradient penalty
$\lossgp$ and $\lossce=||\z||_{2}^{2}$ corresponds to
$\mathcal{L}_{\mathsf{KL}}^{\RAE}$, thus recovering our RAE framework as
presented in Eq.~\ref{eq:rae-loss}.


\section{Network architecture, Training Details and Evaluation}
\label{sec:architecture-training}

We follow the models adopted by \citet{Tolstikhin2017} with the difference that
we consistently apply batch normalization~\citep{batchnorm}. The latent space
dimension is 16 for MNIST~\citep{mnist}, 128 for CIFAR-10~\citep{cifar10} and 64 for CelebA~\citep{celeba}.

For all experiments, we use the Adam optimizer with a starting learning rate of
$10^{-3}$ which is cut in half every time the validation loss plateaus. All
models are trained for a maximum of $100$ epochs on MNIST and CIFAR and $70$
epochs on CelebA. We use a mini-batch size of $100$ and pad MNIST digits with
zeros to make the size $32{\times}32$.

We use the official train, validation and test splits of CelebA. For MNIST and
CIFAR, we set aside 10k train samples for validation. For random sample
evaluation, we draw samples from $\mathcal{N}(0, I)$ for VAE and WAE-MMD and for
all remaining models, samples are drawn from a multivariate Gaussian whose parameters (mean
and covariance) are estimated using training set embeddings. For the GMM density
estimation, we also utilize the training set embeddings for fitting and
validation set embeddings to verify that GMM models are not over fitting to
training embeddings. However, due to the very low number of mixture components (10),
we did not encounter overfitting at this step. The GMM parameters are estimated
by running EM for at most $100$ iterations.

\begin{table}[h]
  \scriptsize
  \setlength{\tabcolsep}{5pt}
\begin{sc}
   \begin{tabular}{c p{0.27\linewidth} p{0.27\linewidth} p{0.27\linewidth}}
       & MNIST & CIFAR\_10 & CELEBA \\
       \toprule
       Encoder: &
       $x \in \mathcal{R}^{32{\times}32}$ \newline
       $\rightarrow \text{Conv}_{128} \rightarrow \text{BN} \rightarrow \text{ReLU}$\newline
       $\quad \rightarrow \text{Conv}_{256}\rightarrow \text{BN} \rightarrow \text{ReLU}$\newline
       $\quad \rightarrow \text{Conv}_{512}\rightarrow \text{BN} \rightarrow \text{ReLU}$\newline
       $\quad \rightarrow \text{Conv}_{1024}\rightarrow \text{BN} \rightarrow \text{ReLU}$\newline
       $\quad \rightarrow \text{Flatten} \rightarrow \text{FC}_{16{\times}M}$
     & $x \in \mathcal{R}^{32{\times}32}$ \newline
       $\rightarrow \text{Conv}_{128} \rightarrow \text{BN} \rightarrow \text{ReLU}$\newline
       $\quad \rightarrow \text{Conv}_{256}\rightarrow \text{BN} \rightarrow \text{ReLU}$\newline
       $\quad \rightarrow \text{Conv}_{512}\rightarrow \text{BN} \rightarrow \text{ReLU}$\newline
       $\quad \rightarrow \text{Conv}_{1024}\rightarrow \text{BN} \rightarrow \text{ReLU}$\newline
       $\quad \rightarrow \text{Flatten} \rightarrow \text{FC}_{128{\times}M}$
     & $x \in \mathcal{R}^{64{\times}64}$ \newline
       $\rightarrow \text{Conv}_{128} \rightarrow \text{BN} \rightarrow \text{ReLU}$\newline
       $\quad \rightarrow \text{Conv}_{256}\rightarrow \text{BN} \rightarrow \text{ReLU}$\newline
       $\quad \rightarrow \text{Conv}_{512}\rightarrow \text{BN} \rightarrow \text{ReLU}$\newline
       $\quad \rightarrow \text{Conv}_{1024}\rightarrow \text{BN} \rightarrow \text{ReLU}$\newline
       $\quad \rightarrow \text{Flatten} \rightarrow \text{FC}_{64{\times}M}$\\

\midrule
Decoder: &
       $z \in \mathcal{R}^{16} \rightarrow \text{FC}_{8{\times}8{\times}1024}$\newline
       $\rightarrow \text{BN} \rightarrow \text{ReLU}$\newline
       $\rightarrow \text{ConvT}_{512}\rightarrow \text{BN} \rightarrow \text{ReLU}$\newline
       $\rightarrow \text{ConvT}_{256}\rightarrow \text{BN} \rightarrow \text{ReLU}$\newline
       $\rightarrow \text{ConvT}_{1}$
     & $z \in \mathcal{R}^{128} \rightarrow \text{FC}_{8{\times}8{\times}1024}$\newline
       $\rightarrow \text{BN} \rightarrow \text{ReLU}$\newline
       $\rightarrow \text{ConvT}_{512}\rightarrow \text{BN} \rightarrow \text{ReLU}$\newline
       $\rightarrow \text{ConvT}_{256}\rightarrow \text{BN} \rightarrow \text{ReLU}$\newline
       $\rightarrow \text{ConvT}_{1}$
     & $z \in \mathcal{R}^{64} \rightarrow \text{FC}_{8{\times}8{\times}1024}$\newline
       $\rightarrow \text{BN} \rightarrow \text{ReLU}$\newline
       $\rightarrow \text{ConvT}_{512}\rightarrow \text{BN} \rightarrow \text{ReLU}$\newline
       $\rightarrow \text{ConvT}_{256}\rightarrow \text{BN} \rightarrow \text{ReLU}$\newline
       $\rightarrow \text{ConvT}_{128}\rightarrow \text{BN} \rightarrow \text{ReLU}$\newline
       $\rightarrow \text{ConvT}_{1}$ \\
 \bottomrule
   \end{tabular}
\end{sc}
\end{table}

$\text{Conv}_n$ represents a convolutional layer with $n$ filters. All
convolutions $\text{Conv}_n$ and transposed convolutions $\text{ConvT}_n$ have a
filter size of $4{\times}4$ for MNIST and CIFAR-10 and $5{\times}5$ for CELEBA.
They all have a stride of size 2 except for the last convolutional layer in the
decoder. Finally, $M=1$ for all models except for the VAE which has $M=2$ as the
encoder has to produce both mean and variance for each input.

\section{Evaluation Setup}
\label{sec:eval-setup}

We compute the FID of the reconstructions of random validation samples against
the test set to evaluate reconstruction quality. For evaluating generative
modeling capabilities, we compute the FID between the test data and randomly
drawn samples from a single Gaussian that is either the isotropic $\prior$
fixed for VAEs and WAEs, a learned second stage VAE for 2sVAEs,
or a single Gaussian fit to $\priorde$ for CV-VAEs
and RAEs. For all models, we also evaluate random samples from a 10-component
Gaussian Mixture model (GMM) fit to $\priorde$. Using only 10 components
prevents us from overfitting (which would indeed give good FIDs when compared
with the test set)\footnote{We note that fitting GMMs with up to 100 components only
improved results marginally. Additionally, we provide nearest-neighbours from
the training set in Appendix~\ref{sec:knn} to show that our models are not
overfitting.}.

For interpolations, we report the FID for the furthest
interpolation points resulted by applying spherical interpolation to randomly
selected validation reconstruction pairs.

We use 10k samples for all FID and PRD evaluations. Scores for random samples
are evaluated against the test set. Reconstruction scores are computed from
validation set reconstructions against the respective test set. Interpolation
scores are computed by interpolating latent codes of a pair of randomly chosen
validation embeddings vs test set samples. The visualized interpolation samples
are interpolations between two randomly chosen test set images.

\section{Evaluation by Precision and Recall}
\label{sec:prd-eval}

\begin{table}[h]
 \centering
 \begin{sc}
   \small
   \begin{tabular}{l c c c c c c}
     \toprule
      & \multicolumn{2}{c}{MNIST} & \multicolumn{2}{c}{CIFAR-10} & \multicolumn{2}{c}{CelebA} \\
   \cmidrule(r){2-3}\cmidrule(r){4-5}\cmidrule(r){6-7}
      &  $\mathcal{N}$ & $\mathsf{GMM}$&  $\mathcal{N}$ &$\mathsf{GMM}$& $\mathcal{N}$ &$\mathsf{GMM}$\\
     \cmidrule(r){2-3}\cmidrule(r){4-5}\cmidrule(r){6-7}
     VAE  & 0.96 / 0.92 & 0.95 / 0.96 & 0.25 / 0.55 & 0.37 / 0.56 & 0.54 / 0.66 & 0.50 / 0.66\\

     CV-VAE  & 0.84 / 0.73 & 0.96 / 0.89 & 0.31 / 0.64 & 0.42 / 0.68 & 0.25 / 0.43 & 0.32 / 0.55\\

     WAE  & 0.93 / 0.88 & \textbf{0.98} / 0.95 & 0.38 / 0.68 & 0.51 / \textbf{0.81} & 0.59 / 0.68 & \textbf{0.69} / \textbf{0.77}\\
     \midrule
     \ourmodel{}-GP & 0.93 / 0.87 & 0.97 / \textbf{0.98} & 0.36 / 0.70 & 0.46 / 0.77 & 0.38 / 0.55 & 0.44 / 0.67\\

     \ourmodel{}-L2 & 0.92 / 0.87 & \textbf{0.98} / \textbf{0.98} & 0.41 / 0.77 & \textbf{0.57} / \textbf{0.81} & 0.36 / 0.64 & 0.44 / 0.65\\

     \ourmodel{}-SN  & 0.89 / 0.95 & \textbf{0.98} / 0.97 & 0.36 / 0.73 & 0.52 / \textbf{0.81} & 0.54 / 0.68 & 0.55 / 0.74\\

     \ourmodel{}  & 0.92 / 0.85 & \textbf{0.98} / \textbf{0.98} & 0.45 / 0.73 & 0.53 / 0.80 & 0.46 / 0.59 & 0.52 / 0.69\\

     AE   & 0.90 / 0.90 & \textbf{0.98} / 0.97 & 0.37 / 0.73 & 0.50 / 0.80 & 0.45 / 0.66 & 0.47 / 0.71\\
     \bottomrule
 \end{tabular}
 \end{sc}
 \vspace{2.5mm}
 \caption{Evaluation of random sample quality by precision / recall~\citep{prd}
 (higher numbers are better, best value for each dataset in bold). It is
 notable that the proposed ex-post density estimation improves not only
 precision, but also recall throughout the experiment. For example, WAE seems
 to have a comparably low recall of only $0.88$ on MNIST which is raised
 considerably to $0.95$ by fitting a GMM. In all cases, GMM gives the best
 results. Another interesting point is the low precision but high recall of all
 models on CIFAR-10 -- this is also visible upon inspection of the samples in
 Fig.~\ref{fig:cifar-pics}.}
 \label{tab:prd-main}
\end{table}

\begin{figure*}[h]
 \centering
 \scalebox{.9}
{\setlength\tabcolsep{3pt}
\begin{sc}
 \small
   \begin{tabular}{c c c}
   \toprule
       PRD all RAEs  & PRD all traditional VAEs & WAE VS RAE-SN VS WAE-GMM \\
       \midrule

       \includegraphics[width=0.3\linewidth]{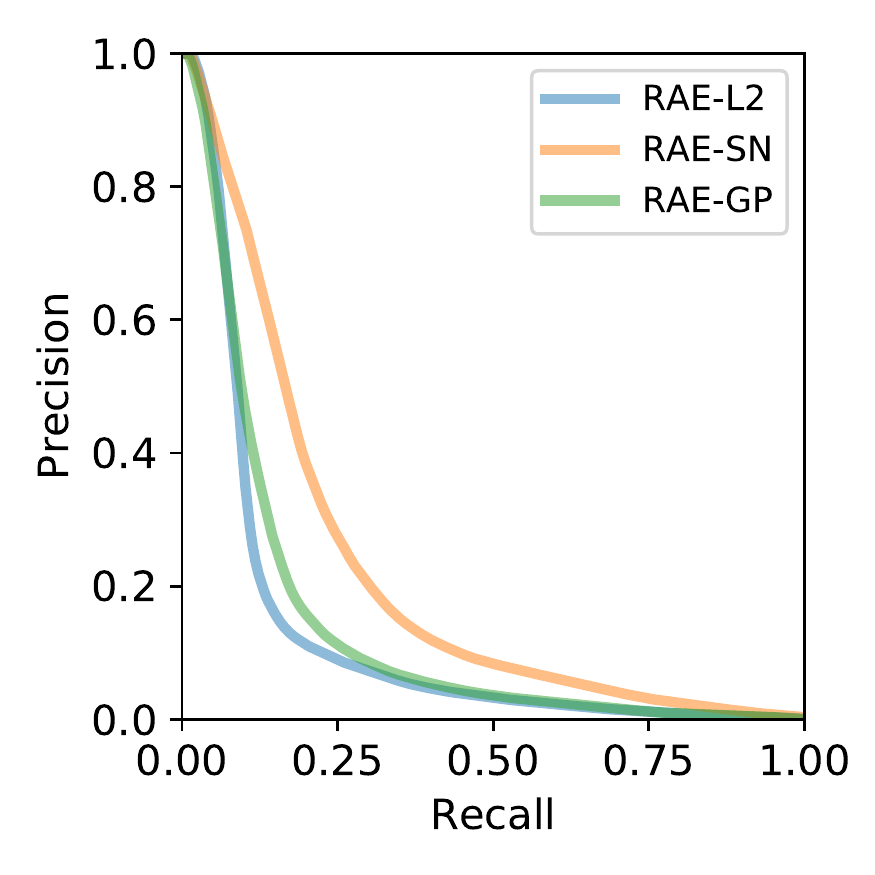}& 
       \includegraphics[width=0.3\linewidth]{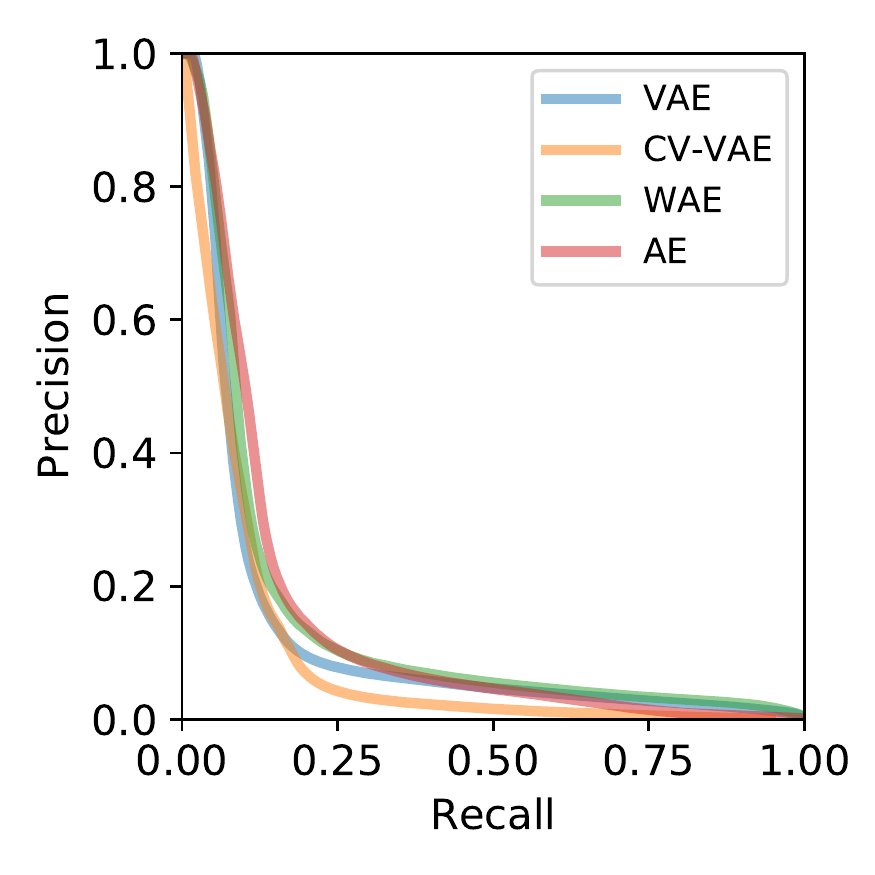}&     \includegraphics[width=0.3\linewidth]{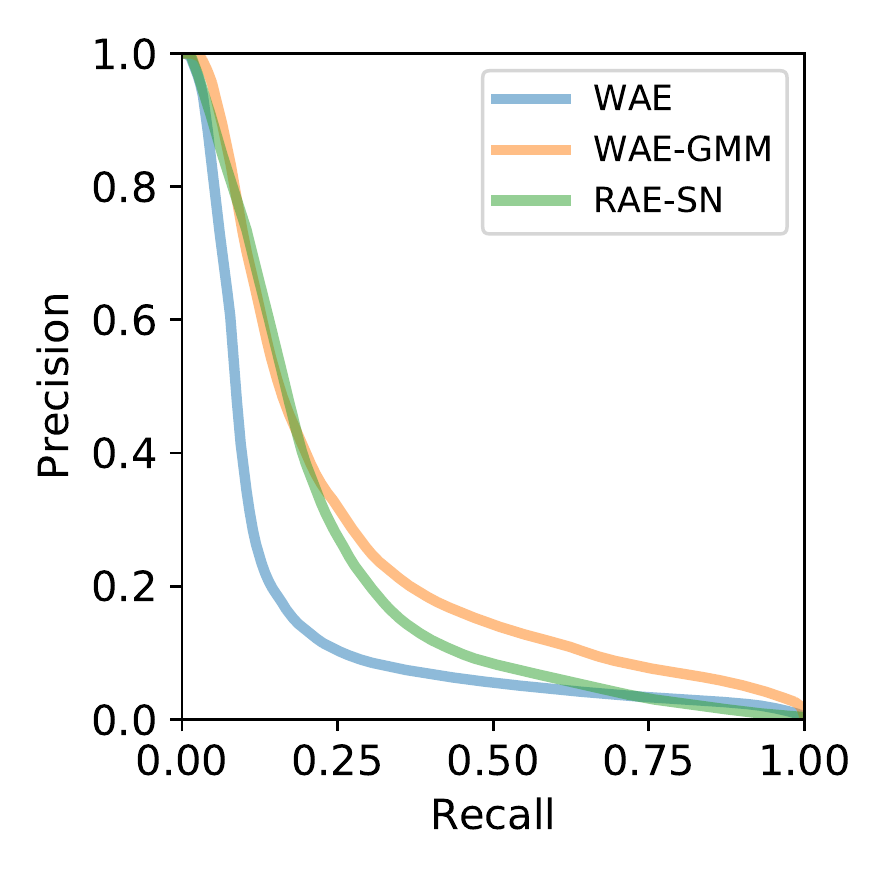} \\[-3.5pt]

     \bottomrule

   \end{tabular}
\end{sc}}
 \vspace{2.5mm}
 \caption{PRD curves of all RAE methods (left), reflects a similar story as FID scores do. RAE-SN seems to perform the best in both precision and recall metric.  PRD curves of all traditional VAE variants (middle). Similar to the conclusion predicted by FID scores there are no clear winner. PRD curves for the WAE (with isotropic Gaussian prior) ,
  WAE-GMM model with ex-post density estimation by a 10-component GMM
  and RAE+SN-GMM (right). This finer grained view shows how the  WAE-GMM scores higher recall but lower precision
  than a RAE+SN-GMM while scoring comparable FID scores. Note that
  ex-post density estimation greatly boosts the WAE model in both PRD
  and FID scores.}
 \label{fig:prd_wae_sn_cmp}
\end{figure*}

\begin{figure*}[h]
 \centering
 \scalebox{.9}
{\setlength\tabcolsep{3pt}
\begin{sc}
 \small
   \begin{tabular}{c c c}
     \toprule
     & MNIST &\\
     \midrule
     VAE  & CV-VAE & WAE \\
     \midrule
     \includegraphics[width=0.3\linewidth]{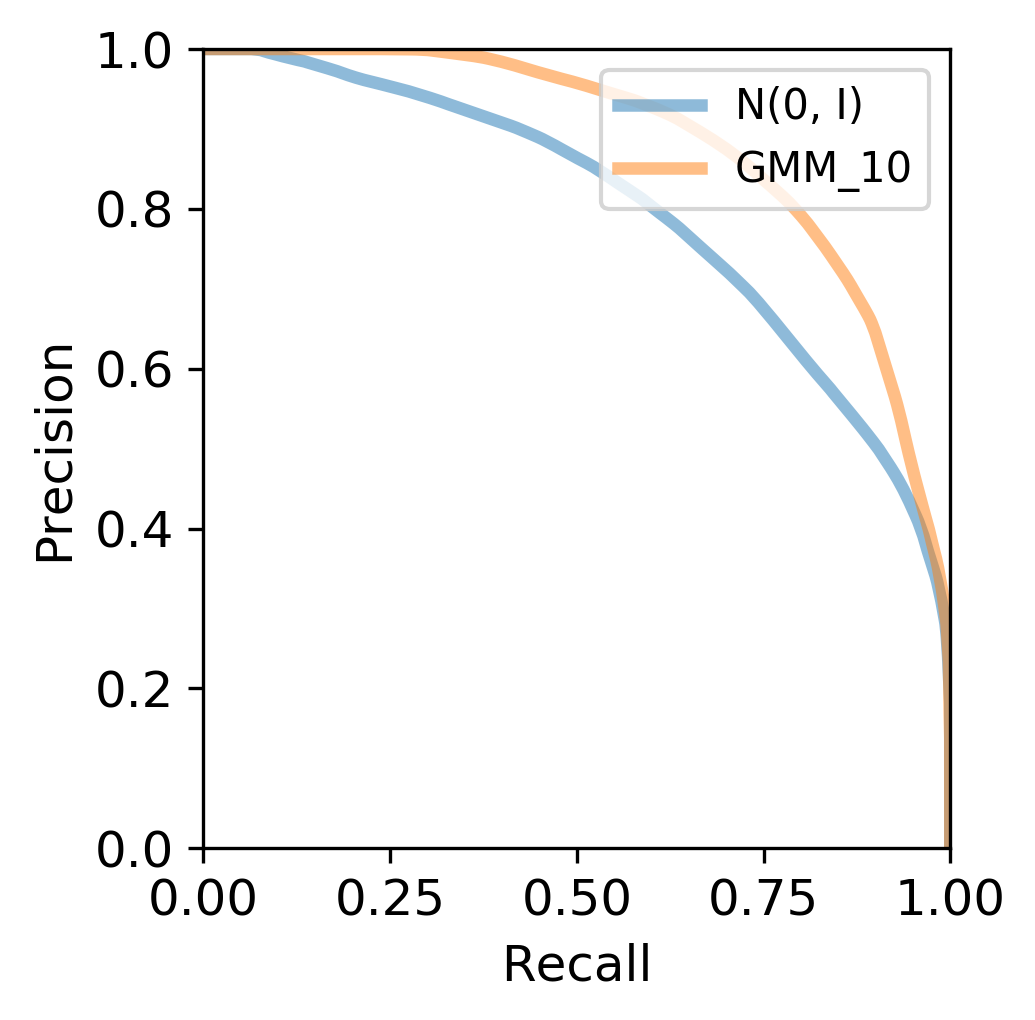}&
     \includegraphics[width=0.3\linewidth]{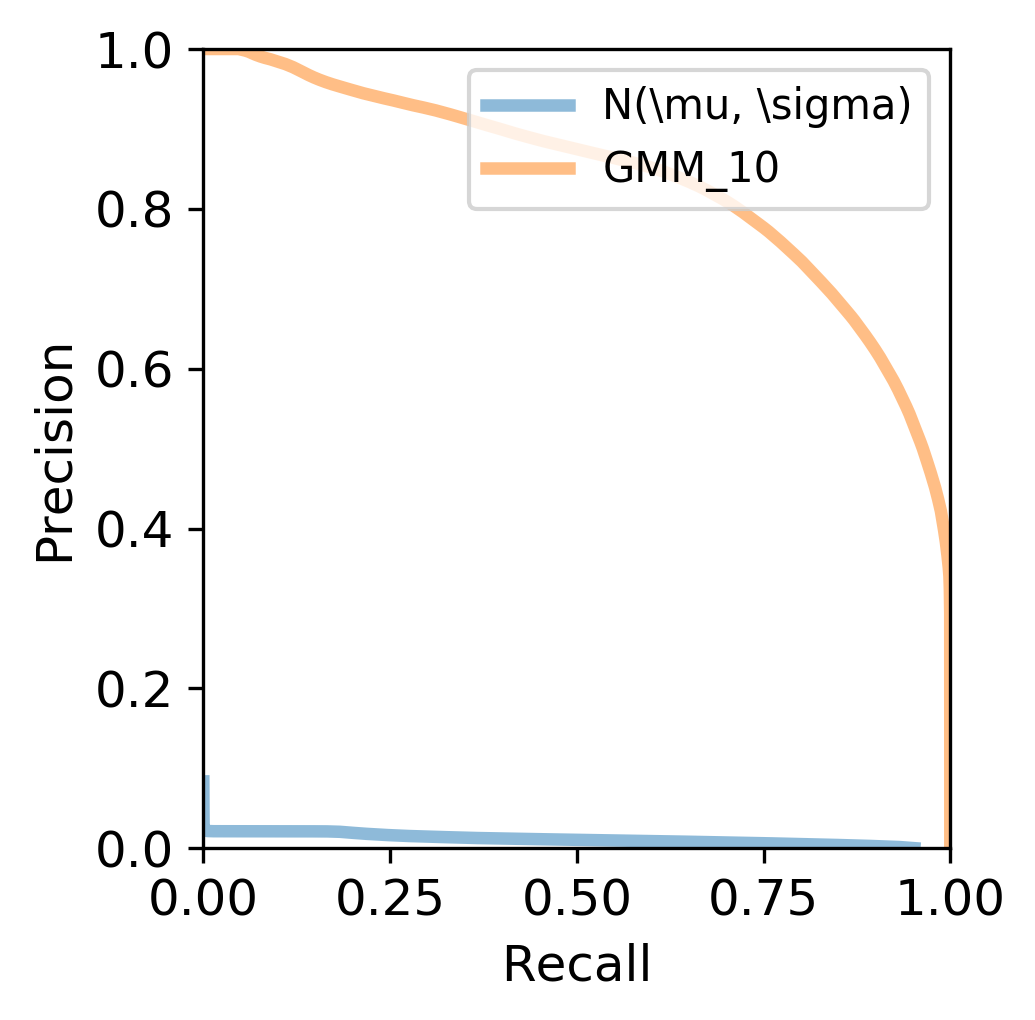}&
     \includegraphics[width=0.3\linewidth]{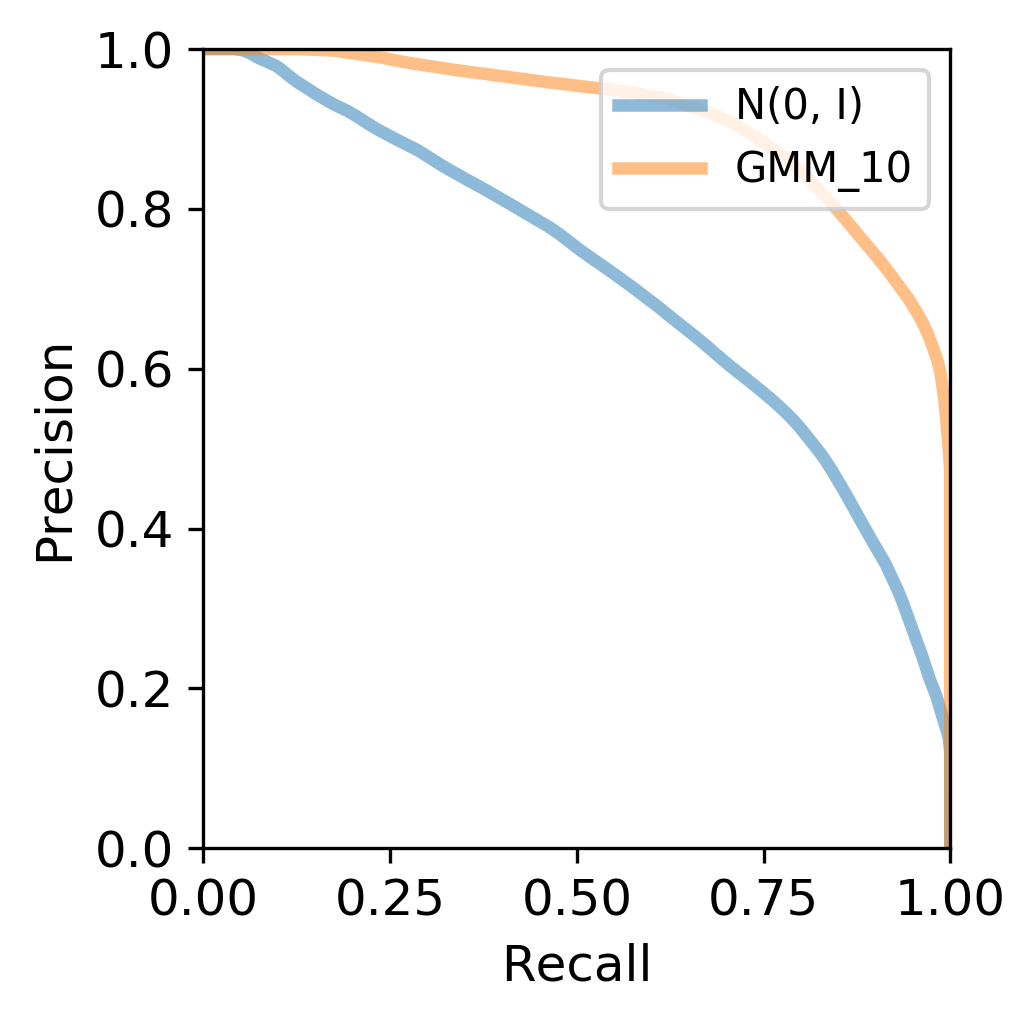}\\
       \midrule
     RAE-GP  & RAE-L2 & RAE-SN \\
     \midrule
     \includegraphics[width=0.3\linewidth]{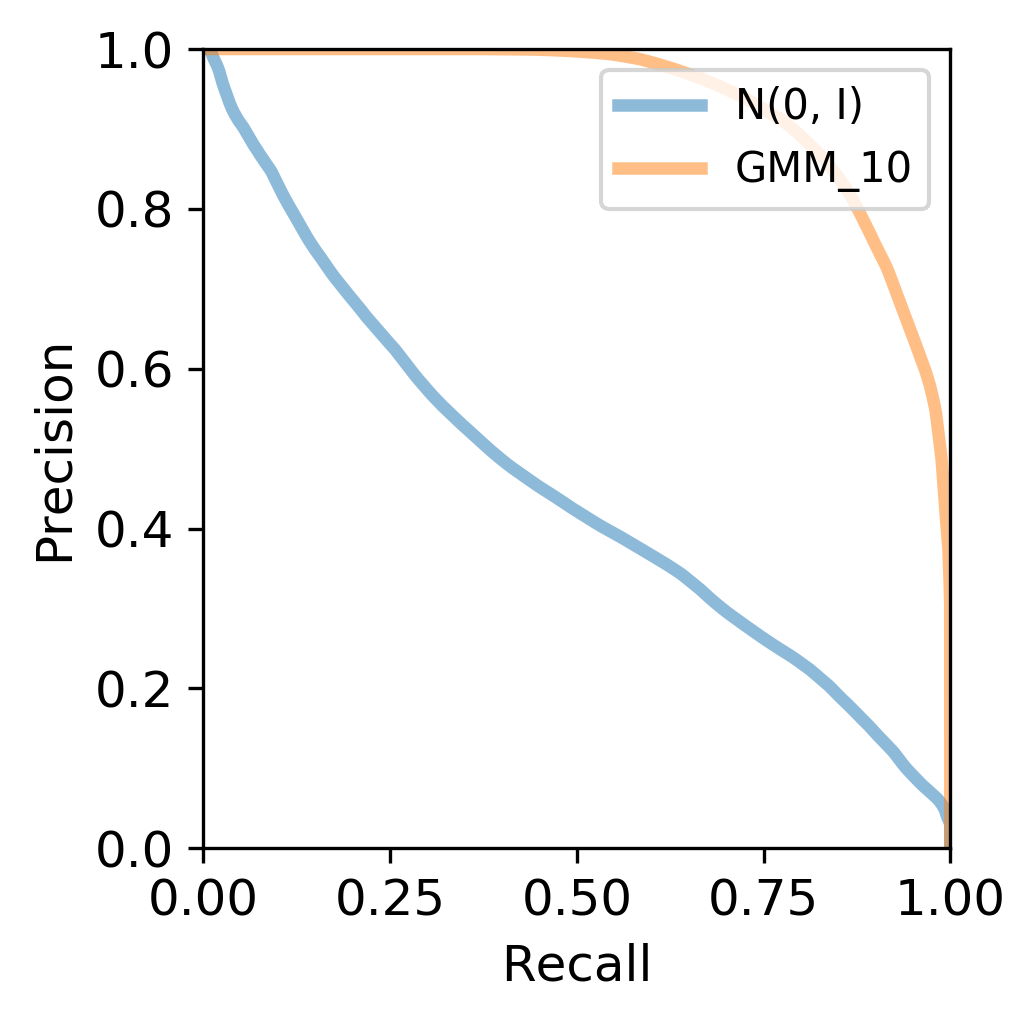}&
     \includegraphics[width=0.3\linewidth]{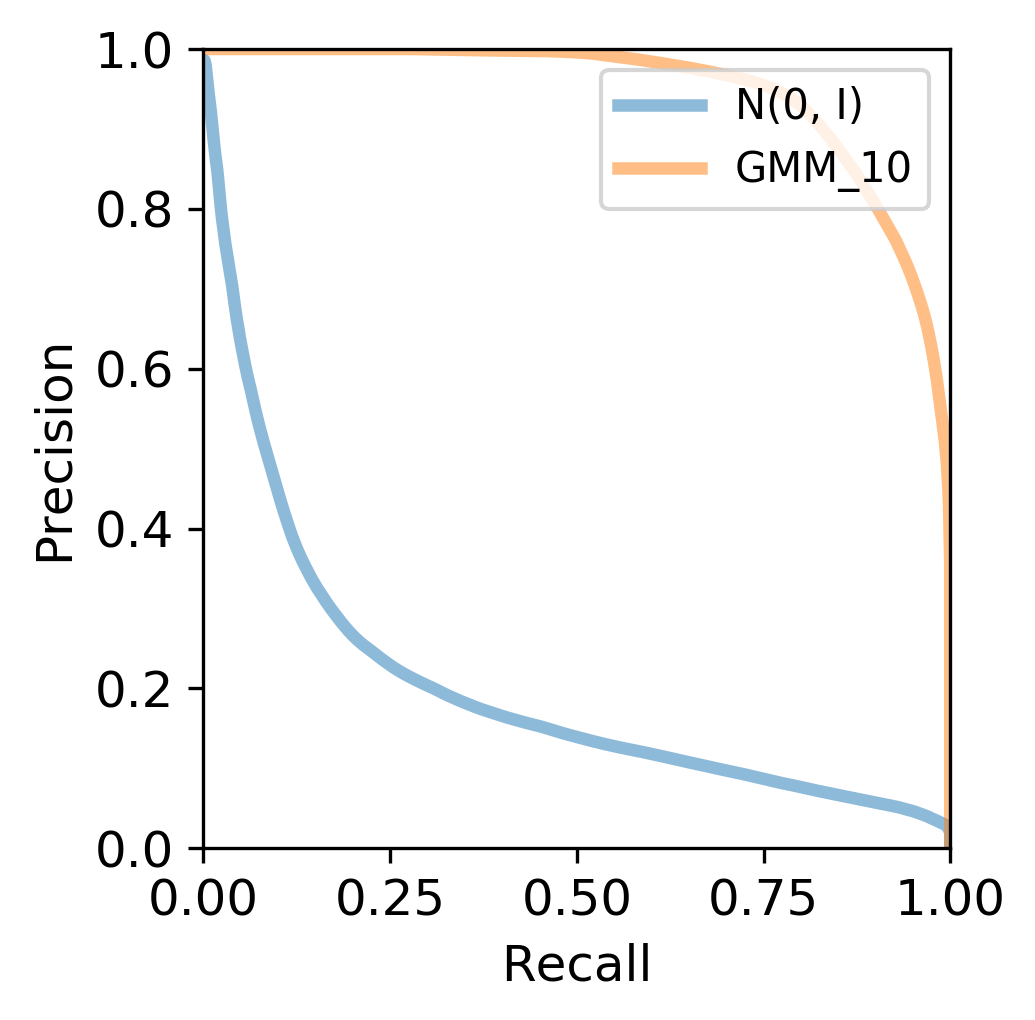}&
     \includegraphics[width=0.3\linewidth]{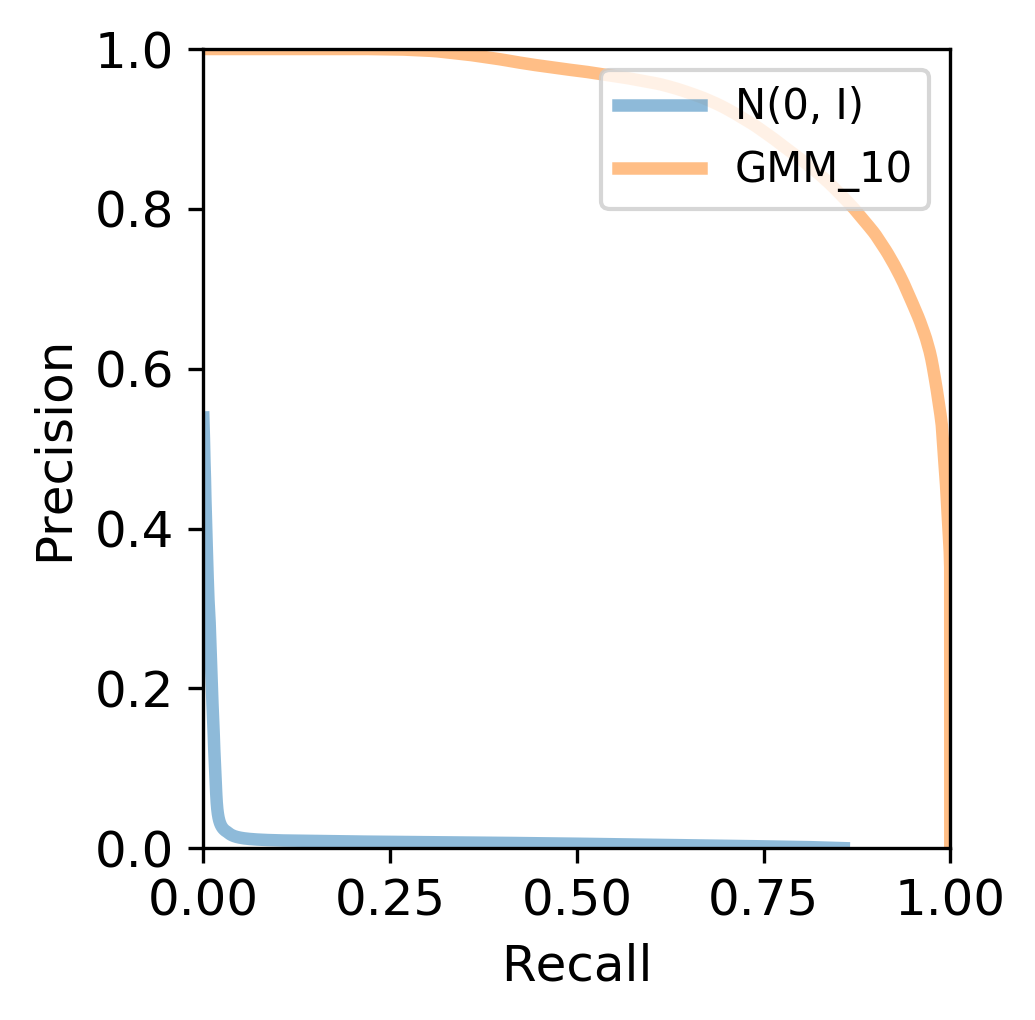}\\
       \midrule
     RAE  & AE &\\
     \midrule
     \includegraphics[width=0.3\linewidth]{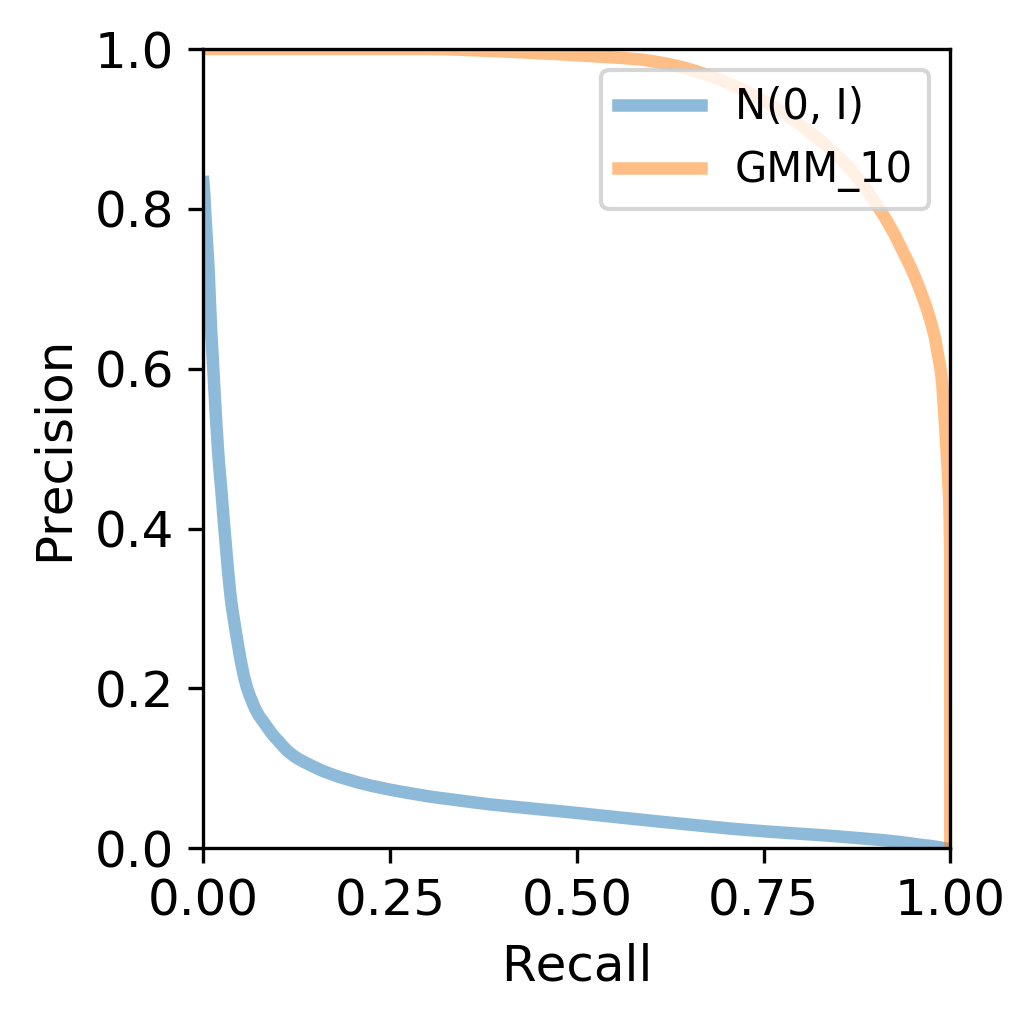}&
     \includegraphics[width=0.3\linewidth]{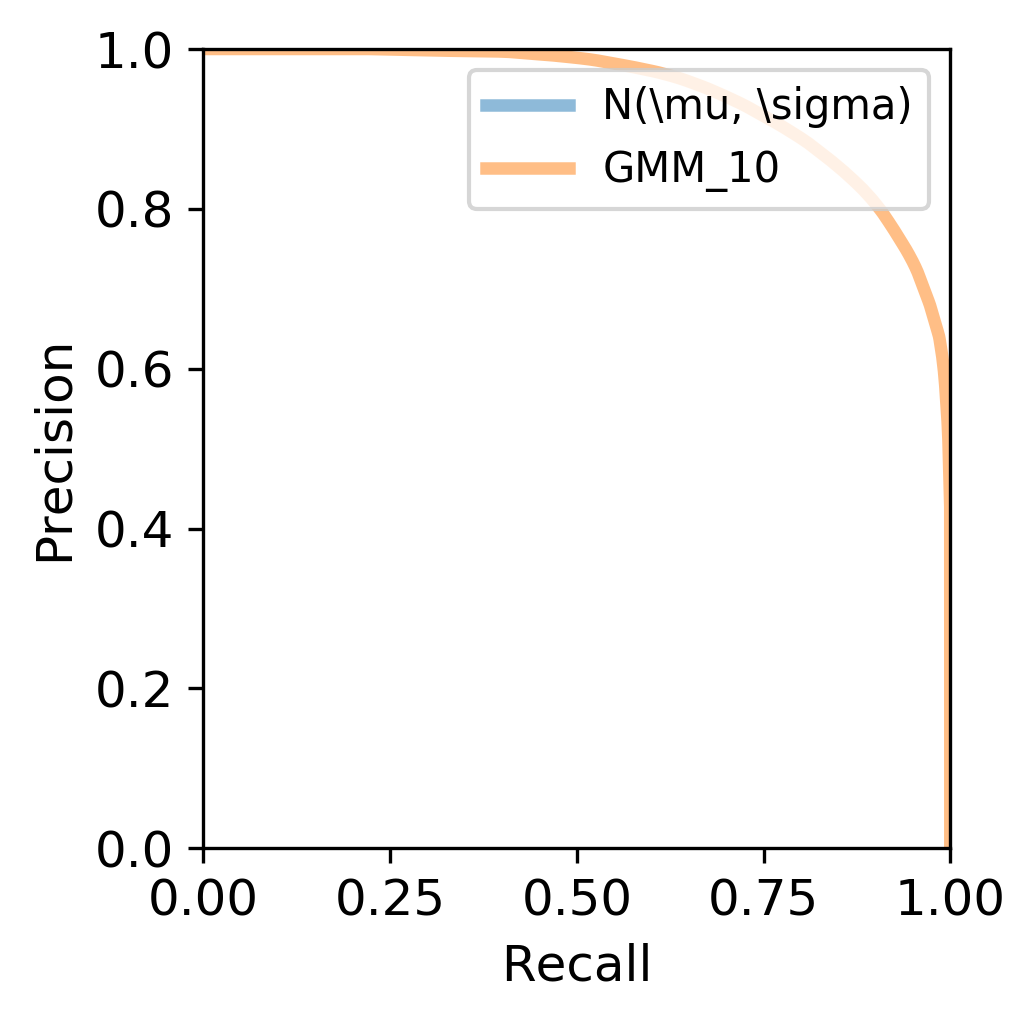}&
     \\ 
      
     \bottomrule

   \end{tabular}
\end{sc}}
 \vspace{2.5mm}
 \caption{PRD curves of all methods on image data experiments on MNIST. For
   each plot, we show the PRD curve when applying the fixed or the
   fitted one by ex-post density estimation (XPDE).
  XPDE greatly boosts both precision and recall for all models.}
 \label{fig:all_prd_mnist}
\end{figure*}

\begin{figure*}[h]
 \centering
 \scalebox{.9}
{\setlength\tabcolsep{3pt}
\begin{sc}
 \small
   \begin{tabular}{c c c}
     \toprule
     & Cifar 10 &\\
     \midrule
VAE  & CV-VAE & WAE \\
     \midrule
     \includegraphics[width=0.3\linewidth]{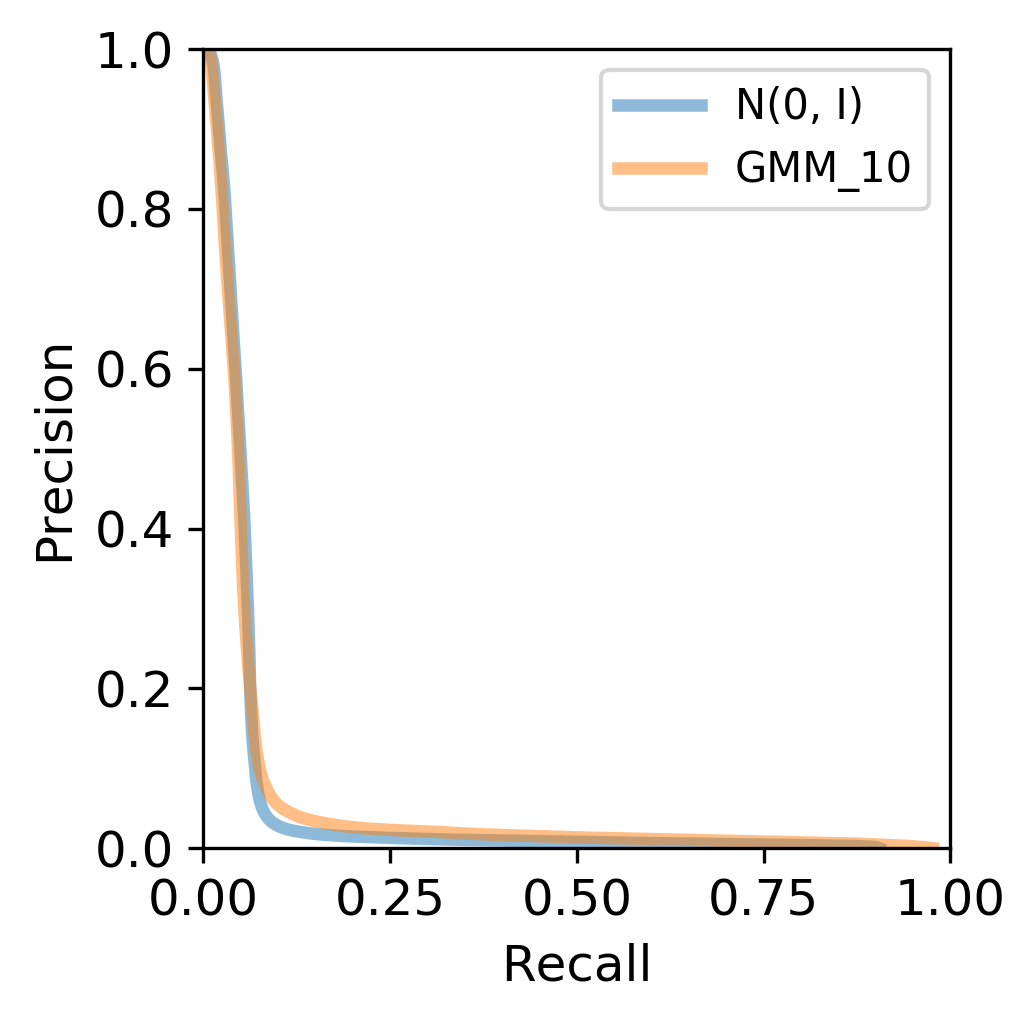}&
     \includegraphics[width=0.3\linewidth]{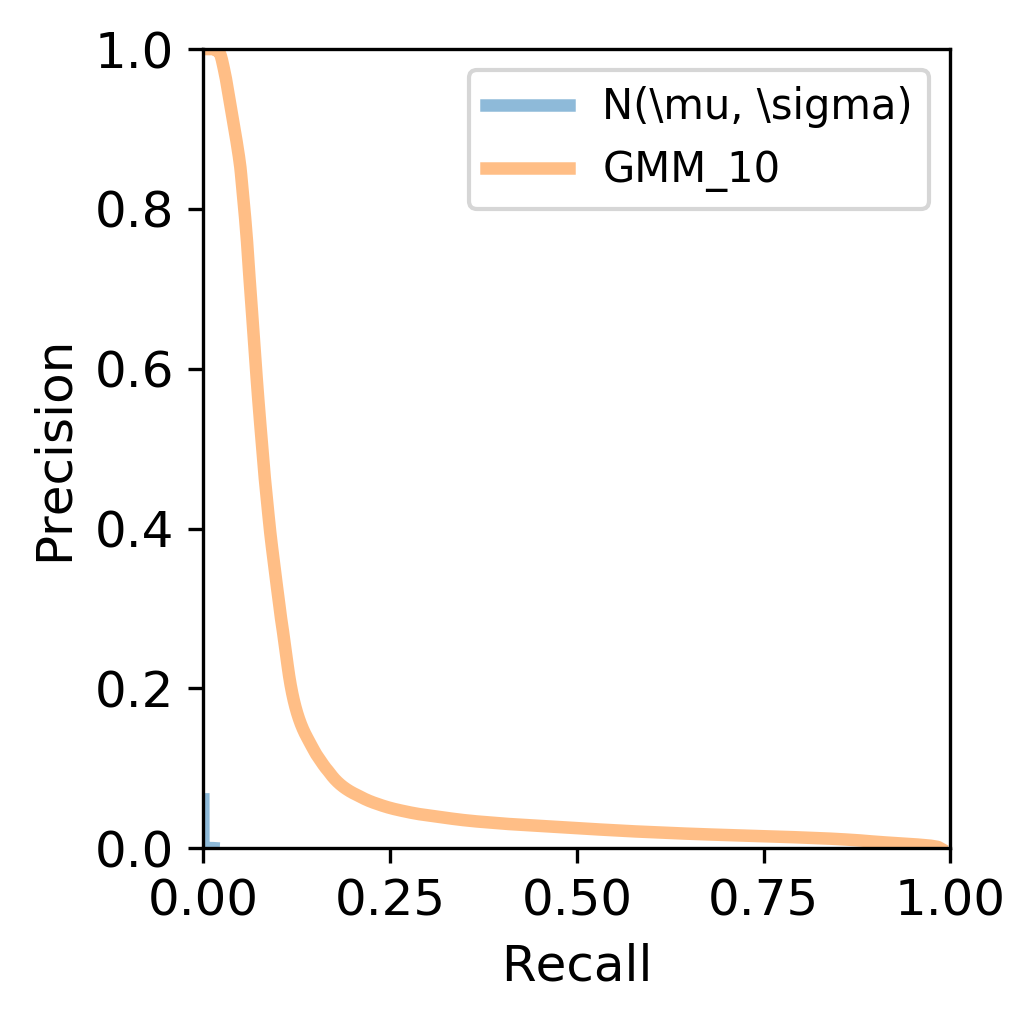}&
     \includegraphics[width=0.3\linewidth]{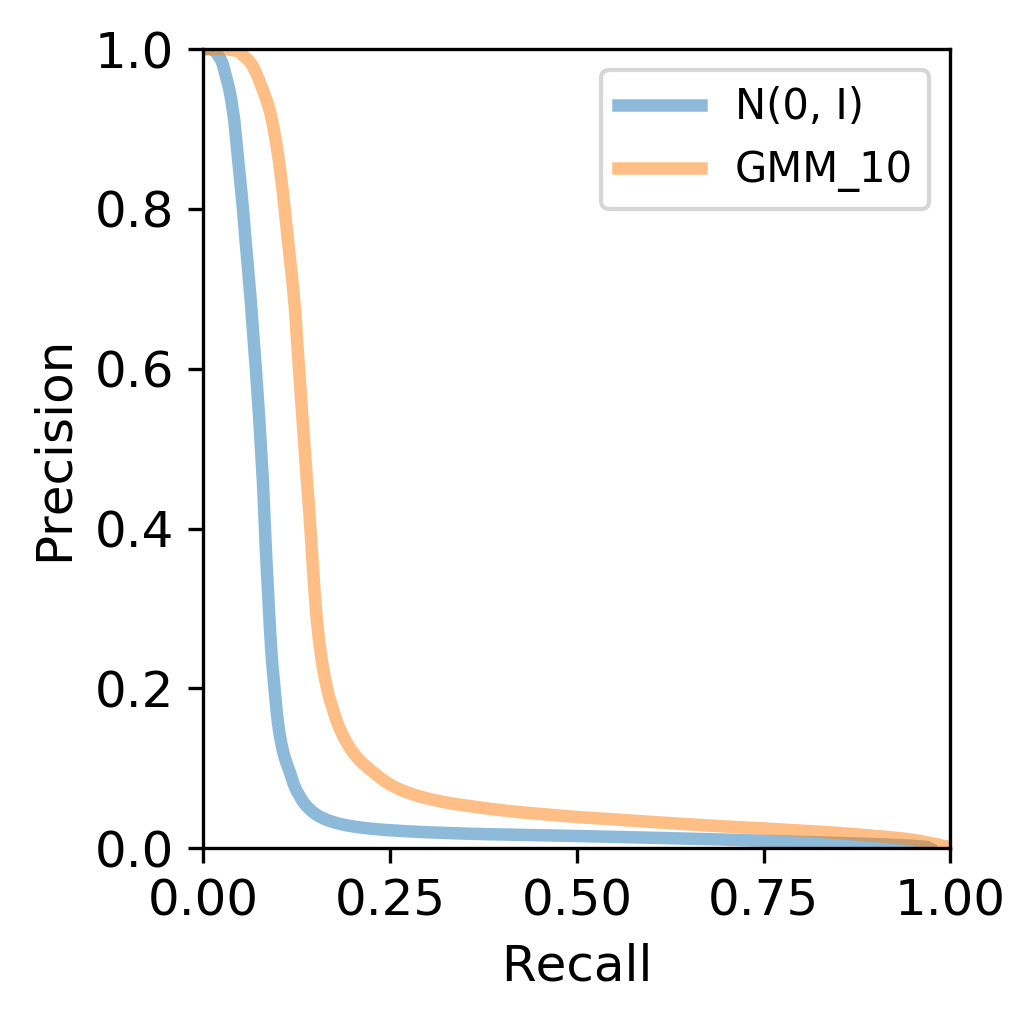}\\
       \midrule
     RAE-GP  & RAE-L2 & RAE-SN \\
     \midrule
     \includegraphics[width=0.3\linewidth]{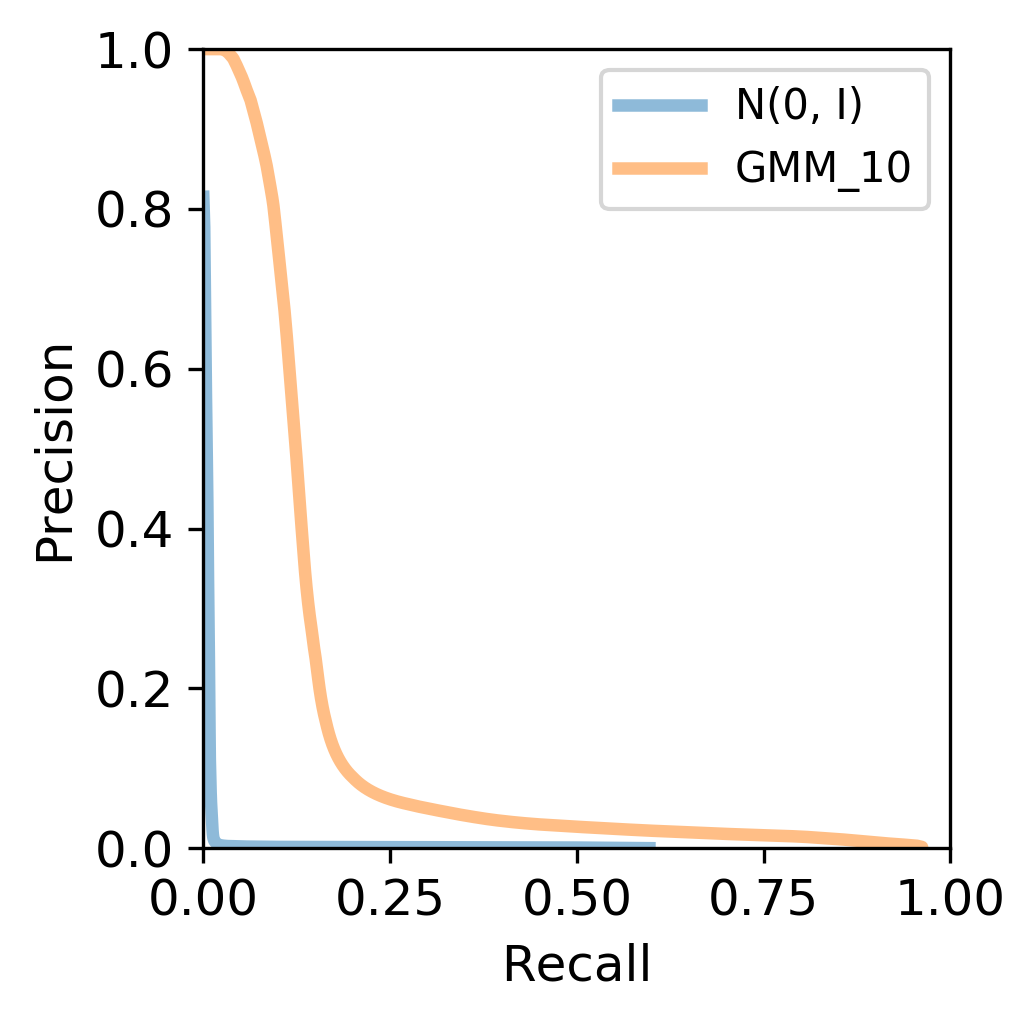}&
     \includegraphics[width=0.3\linewidth]{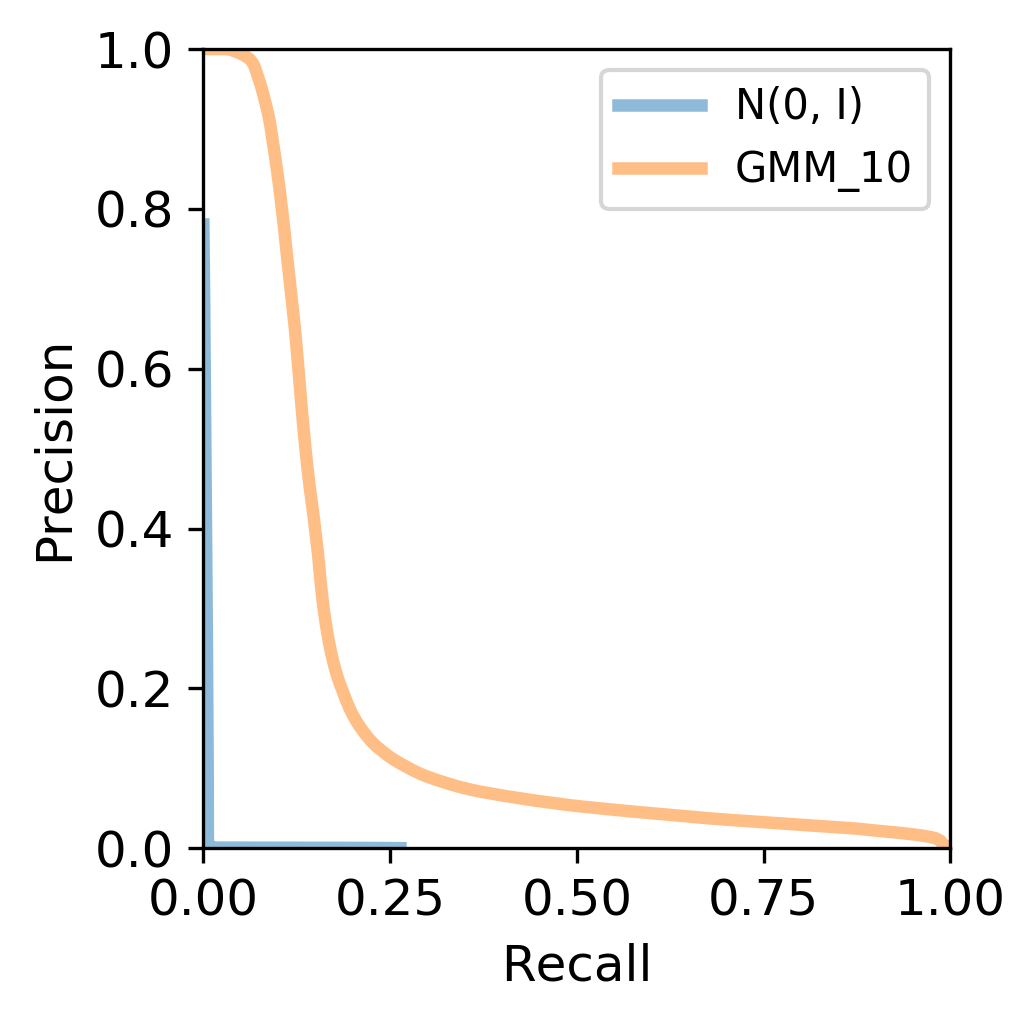}&
     \includegraphics[width=0.3\linewidth]{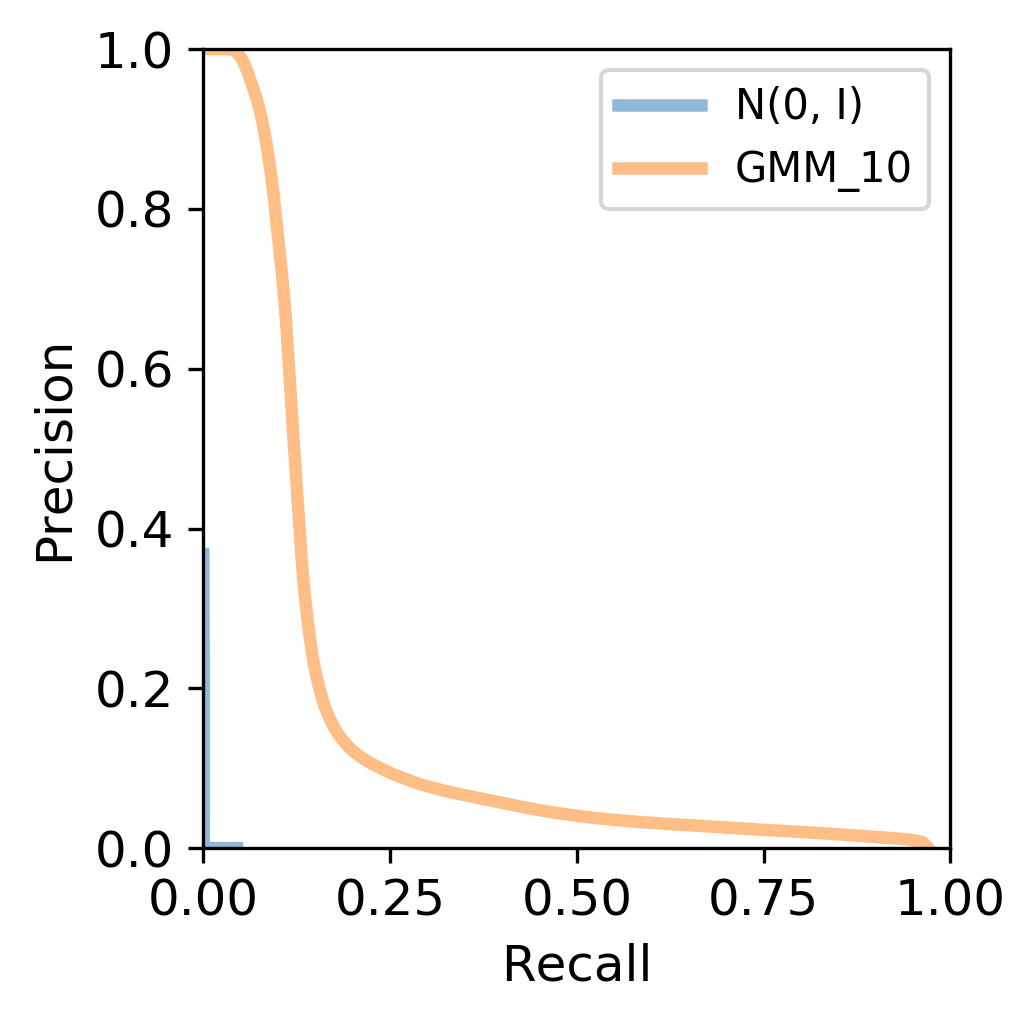}\\
       \midrule
     RAE  & AE &\\
     \midrule
     \includegraphics[width=0.3\linewidth]{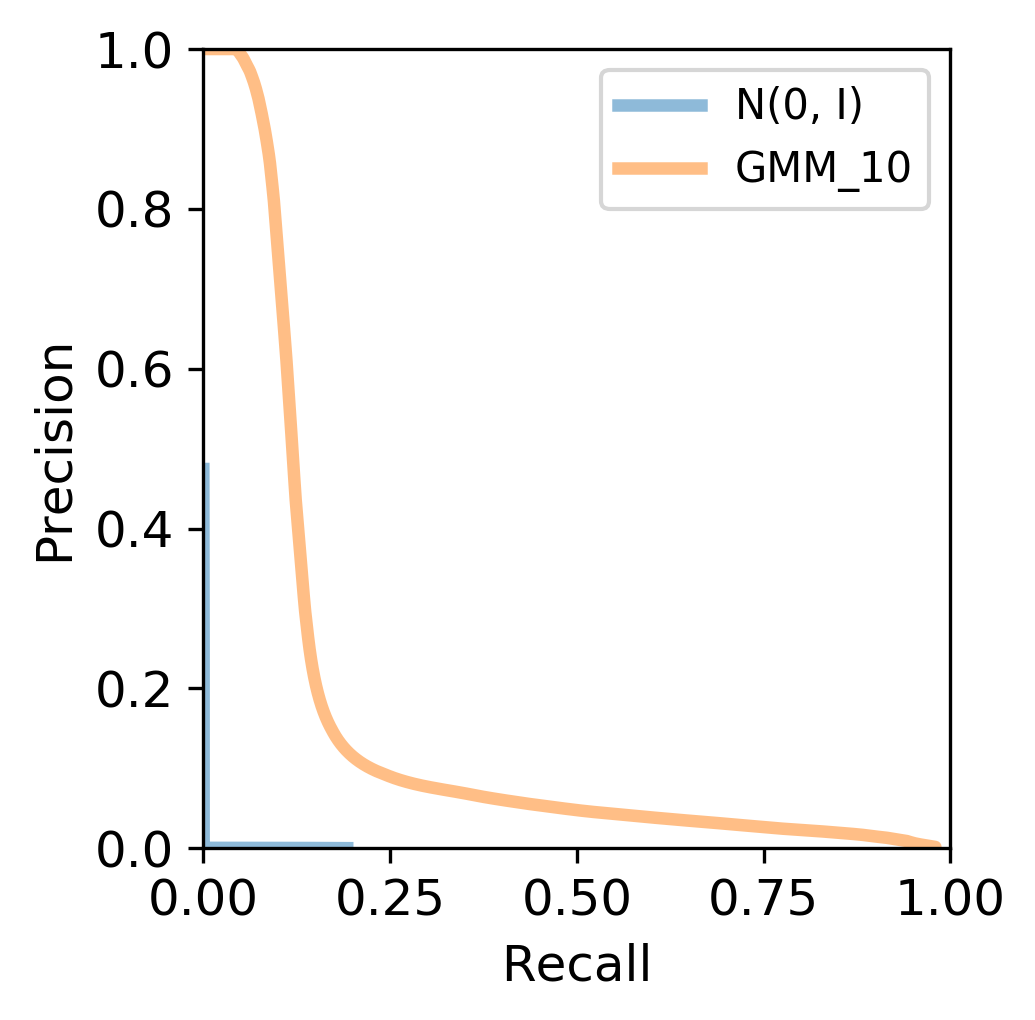}&
     \includegraphics[width=0.3\linewidth]{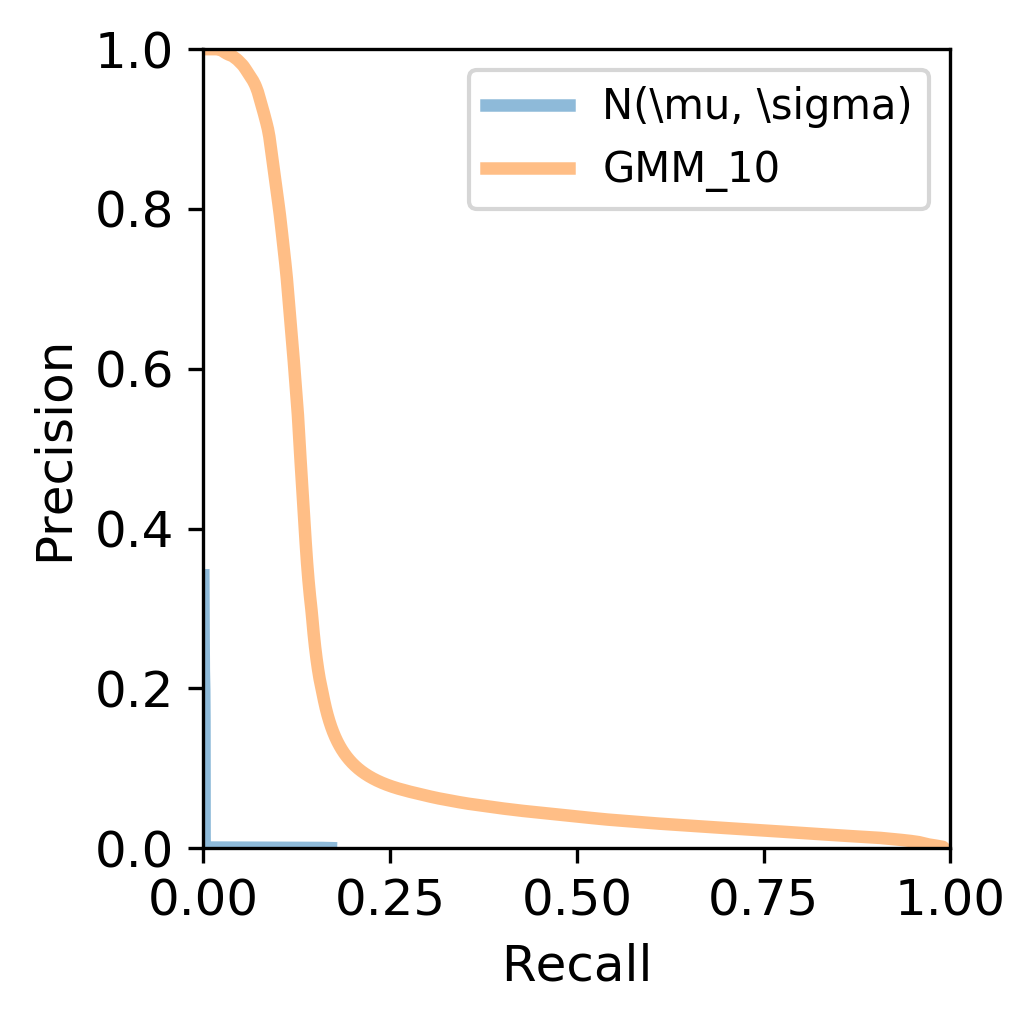}&
     \\       
     \bottomrule

   \end{tabular}
\end{sc}}
 \vspace{2.5mm}
 \caption{PRD curves of all methods on image data experiments on CIFAR10. For
   each plot, we show the PRD curve when applying the fixed or the
   fitted one by ex-post density estimation (XPDE).
  XPDE greatly boosts both precision and recall for all models.}
 \label{fig:all_prd_cifar10}
\end{figure*}

\begin{figure*}[h]
 \centering
 \scalebox{.9}
{\setlength\tabcolsep{3pt}
\begin{sc}
 \small
   \begin{tabular}{c c c}
     \toprule
     & CelebA &\\
     \midrule
VAE  & CV-VAE & WAE \\
     \midrule
     \includegraphics[width=0.3\linewidth]{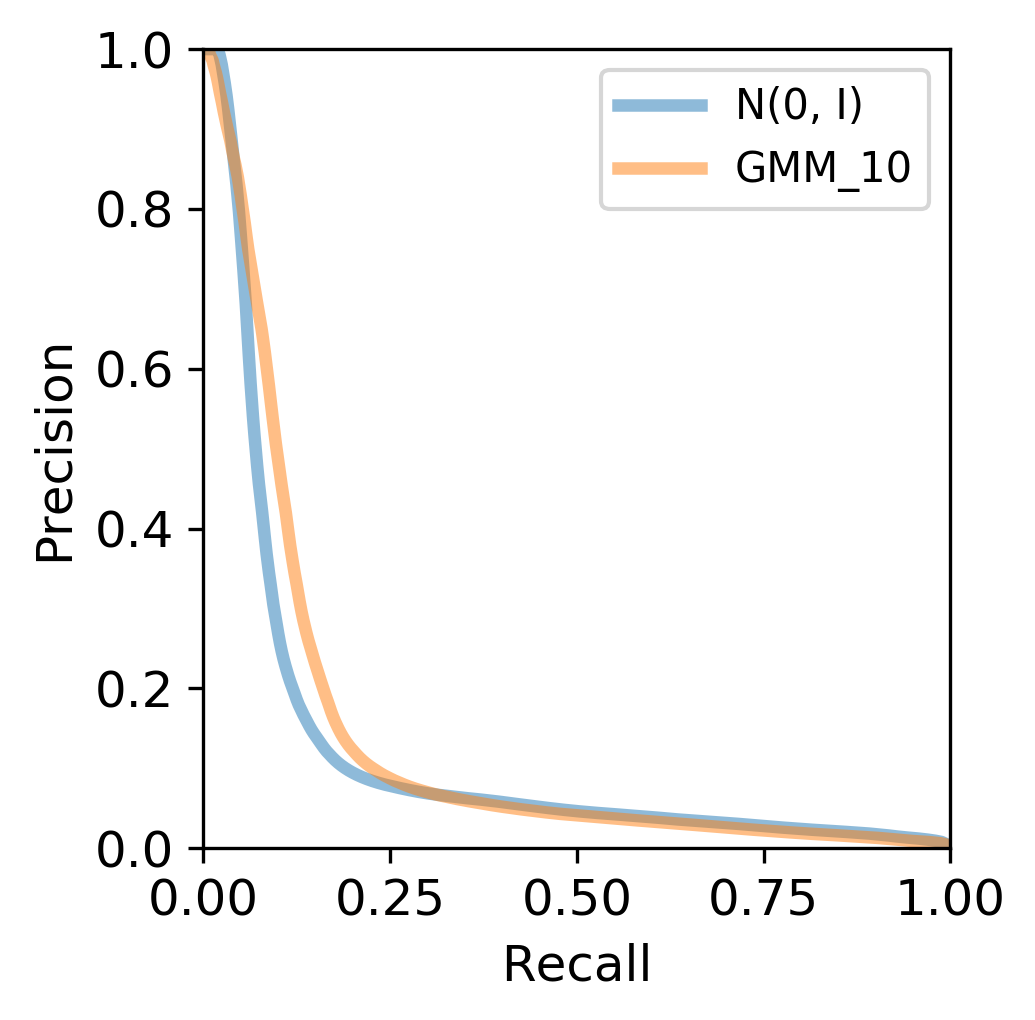}&
     \includegraphics[width=0.3\linewidth]{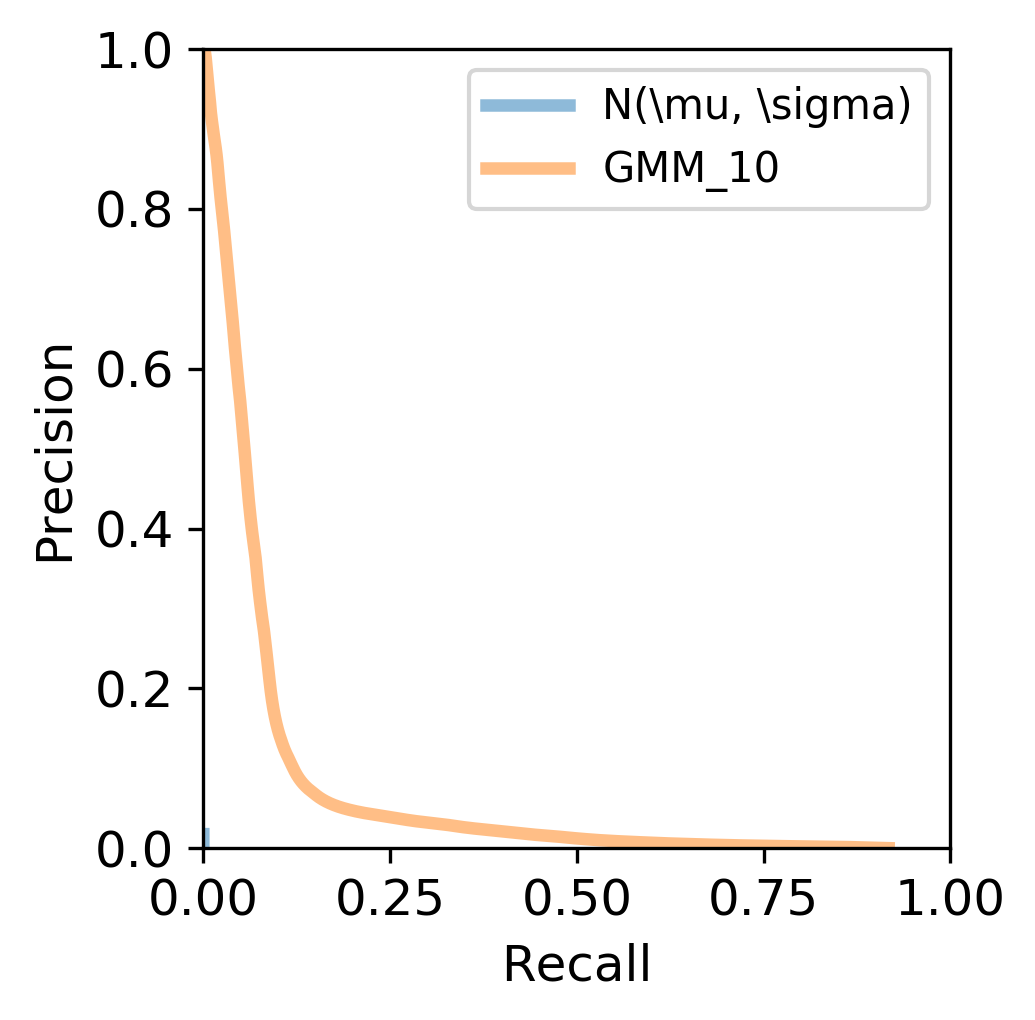}&
     \includegraphics[width=0.3\linewidth]{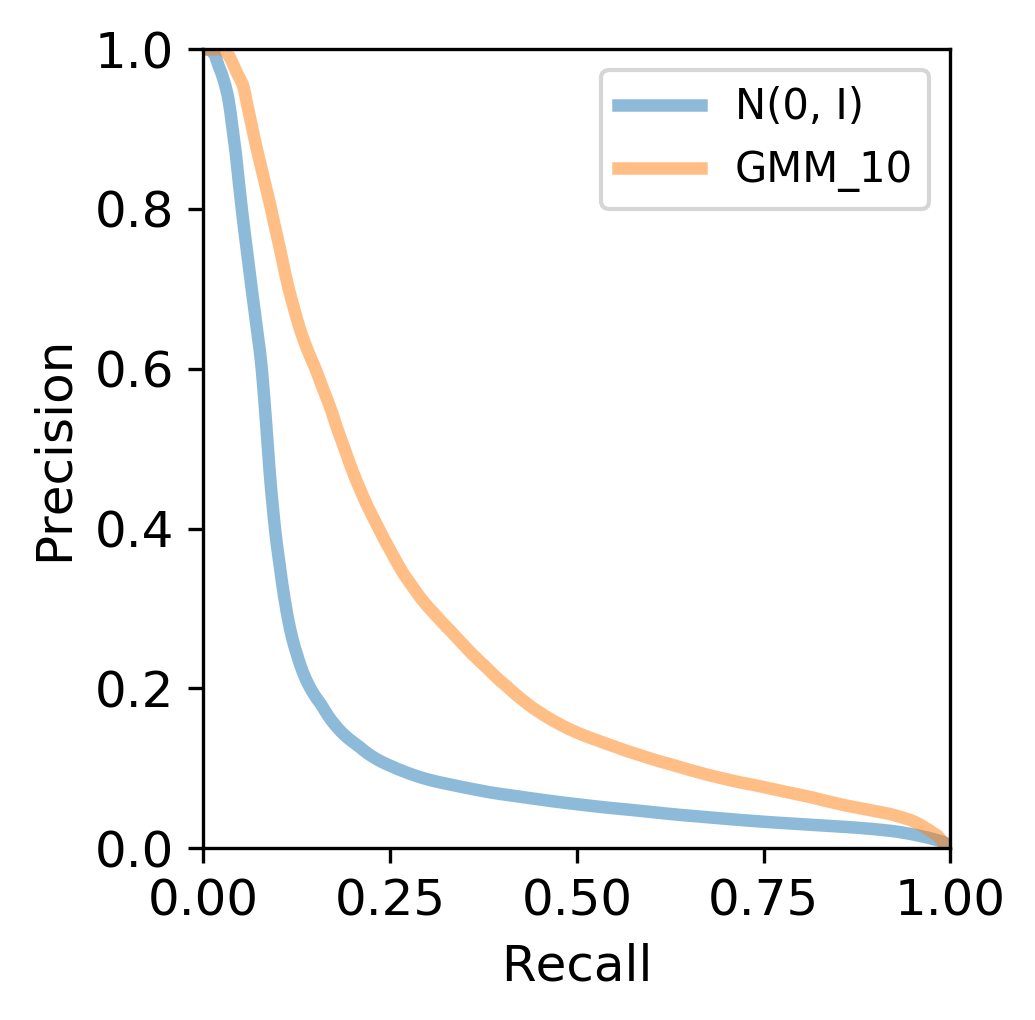}\\
       \midrule
     RAE-GP  & RAE-L2 & RAE-SN \\
     \midrule
     \includegraphics[width=0.3\linewidth]{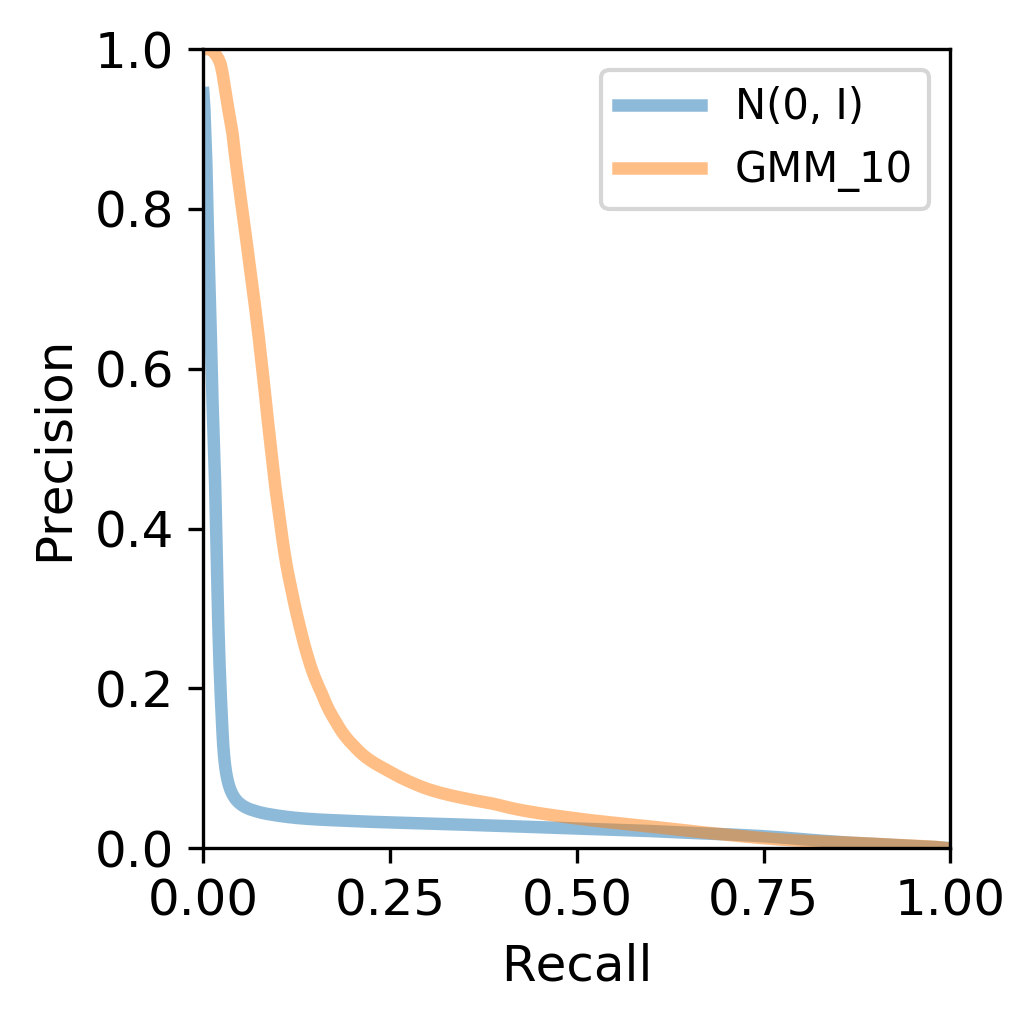}&
     \includegraphics[width=0.3\linewidth]{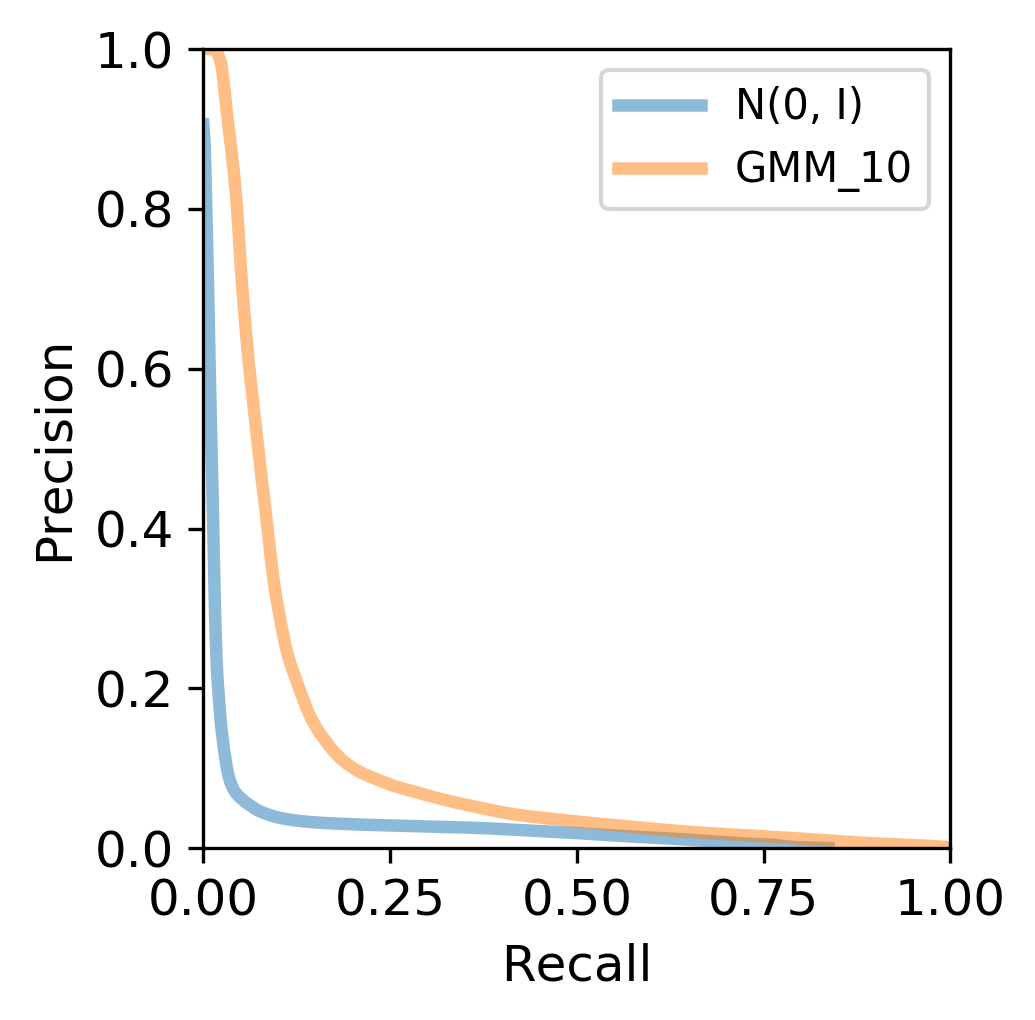}&
     \includegraphics[width=0.3\linewidth]{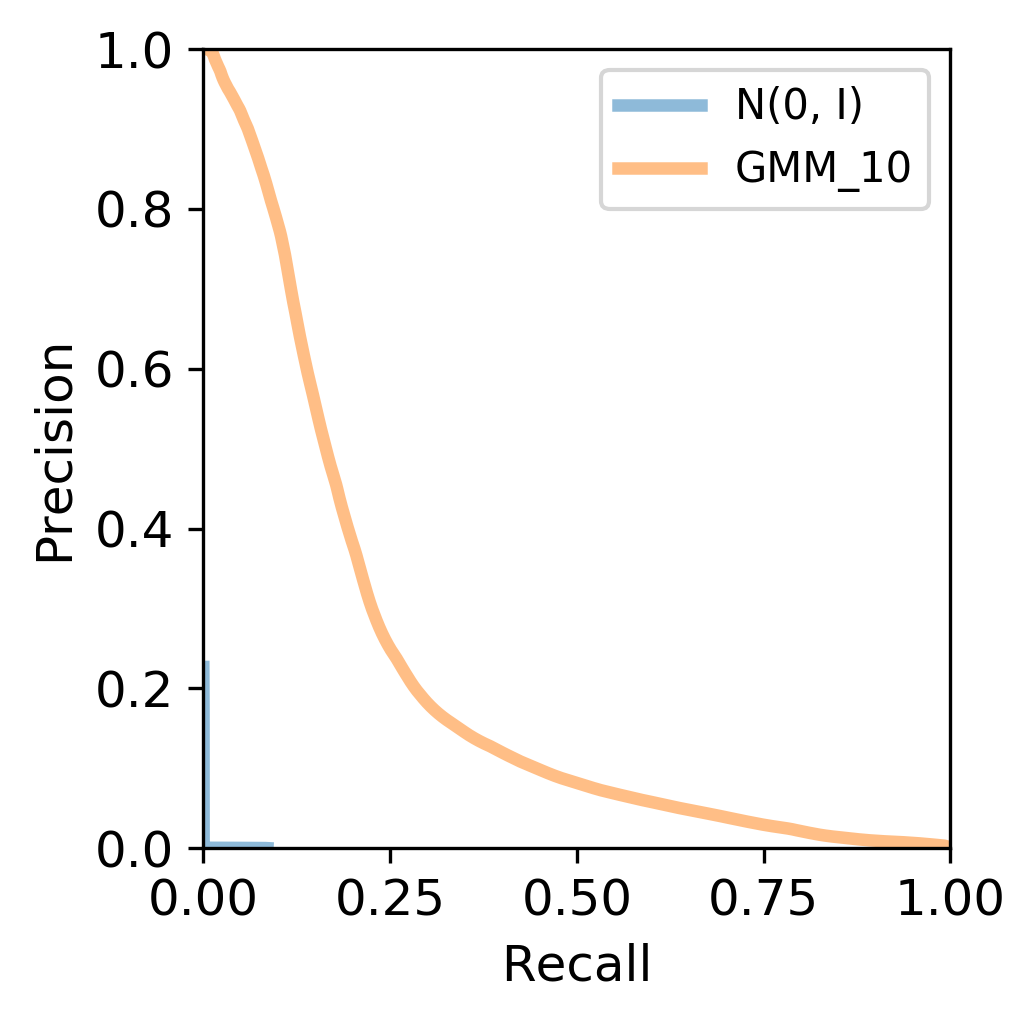}\\
       \midrule
     RAE  & AE &\\
     \midrule
     \includegraphics[width=0.3\linewidth]{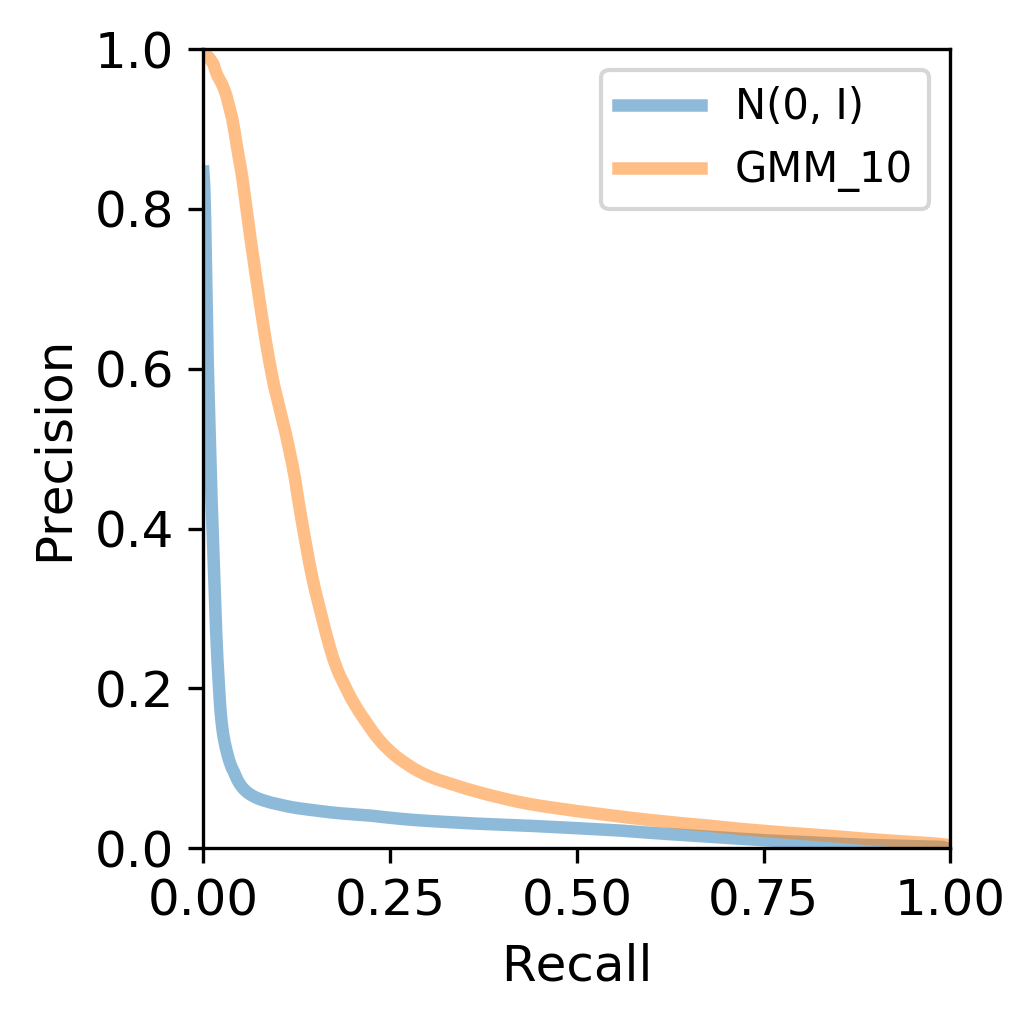}&
     \includegraphics[width=0.3\linewidth]{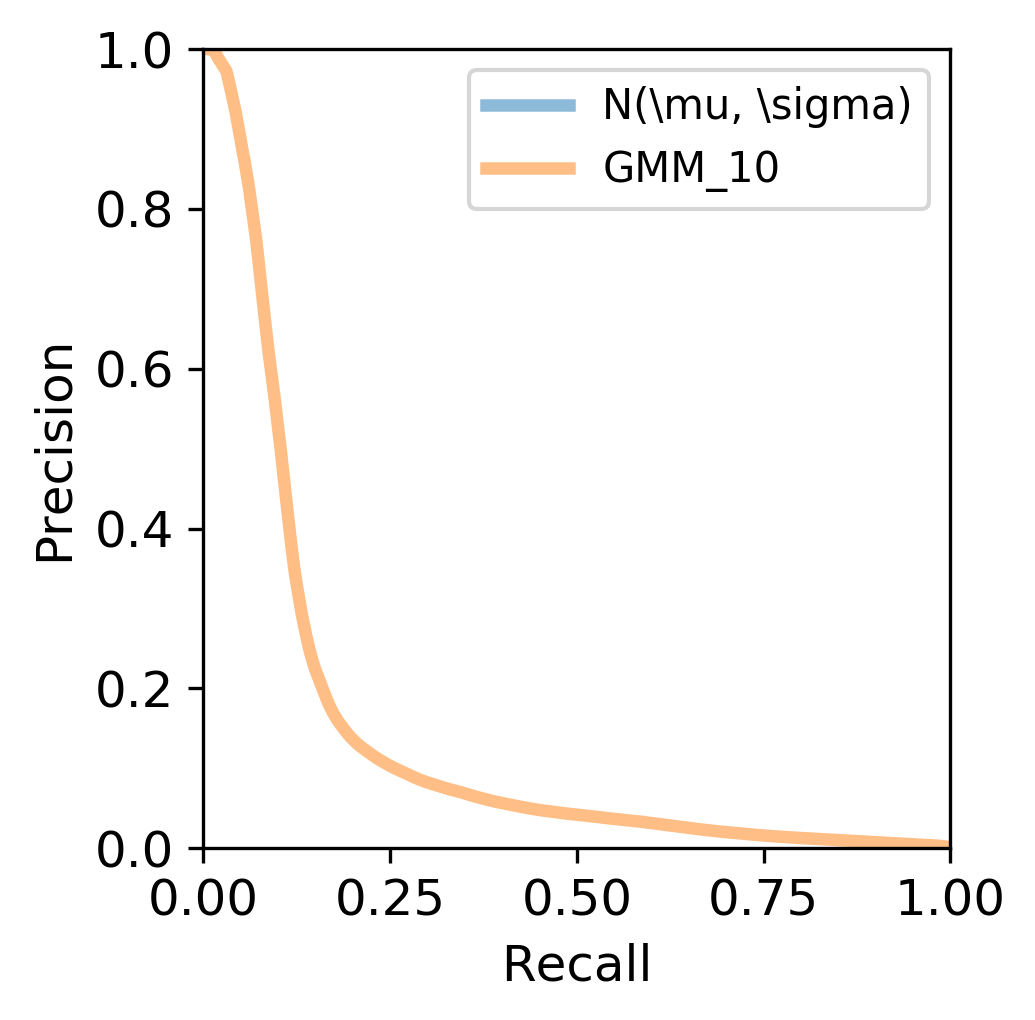}&
     \\       
     \bottomrule

   \end{tabular}
\end{sc}}
 \vspace{2.5mm}
 \caption{PRD curves of all methods on image data experiments on CELEBA. For
   each plot, we show the PRD curve when applying the fixed or the
   fitted one by ex-post density estimation (XPDE).
  XPDE greatly boosts both precision and recall for all models.}
 \label{fig:all_prd_celeba}
\end{figure*}

\clearpage
\section{More Qualitative Results}
\label{sec:more-pics}

\begin{figure*}[h]
 \centering
 \scalebox{.9}
{\setlength\tabcolsep{3pt}
\begin{sc}
 \small
   \begin{tabular}{ r c c c}
   \toprule
       & Reconstructions & Random Samples & Interpolations \\
       \midrule
     \raisebox{10pt}{GT}
     & \includegraphics[trim={ 0 288 0
       0},clip,width=0.32\linewidth]{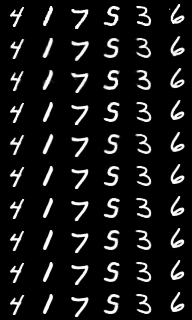}
     \\[-3.5pt]

     \raisebox{10pt}{VAE}
     & \includegraphics[trim={0 256 0 32},clip,width=0.32\linewidth]{figs/pics_new/MNIST_recons.png}&\includegraphics[trim={0 256 0 32},clip,width=0.32\linewidth]{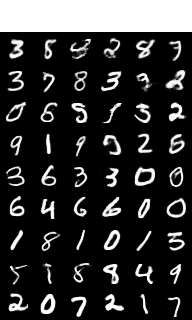} & \includegraphics[trim={0 256 0 32},clip,width=0.32\linewidth]{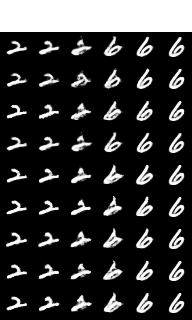} \\[-3.5pt]

     \raisebox{10pt}{CV-VAE}
     & \includegraphics[trim={0 224 0 64},clip,width=0.32\linewidth]{figs/pics_new/MNIST_recons.png}&\includegraphics[trim={0 224 0 64},clip,width=0.32\linewidth]{figs/pics_new/MNIST_random.png} & \includegraphics[trim={0 224 0 64},clip,width=0.32\linewidth]{figs/pics_new/MNIST_interp.png} \\[-3.5pt]

     \raisebox{10pt}{WAE}
     & \includegraphics[trim={ 0 192 0 96},clip,width=0.32\linewidth]{figs/pics_new/MNIST_recons.png}&\includegraphics[trim={ 0 192 0 96},clip,width=0.32\linewidth]{figs/pics_new/MNIST_random.png} & \includegraphics[trim={ 0 192 0 96},clip,width=0.32\linewidth]{figs/pics_new/MNIST_interp.png} \\[-3.5pt]

          \raisebox{10pt}{2sVAE}
     & \includegraphics[trim={ 0 160 0 128},clip,width=0.32\linewidth]{figs/pics_new/MNIST_recons.png}&\includegraphics[trim={0 160 0 128},clip,width=0.32\linewidth]{figs/pics_new/MNIST_random.png} & \includegraphics[trim={0 160 0 128},clip,width=0.32\linewidth]{figs/pics_new/MNIST_interp.png} \\[-3.5pt]

     \raisebox{10pt}{RAE-GP}
     & \includegraphics[trim={0 128 0 160},clip,width=0.32\linewidth]{figs/pics_new/MNIST_recons.png}&\includegraphics[trim={0 128 0 160},clip,width=0.32\linewidth]{figs/pics_new/MNIST_random.png} & \includegraphics[trim={0 128 0 160},clip,width=0.32\linewidth]{figs/pics_new/MNIST_interp.png} \\[-3.5pt]

     \raisebox{10pt}{RAE-L2}
     & \includegraphics[trim={0 96 0 192},clip,width=0.32\linewidth]{figs/pics_new/MNIST_recons.png}&\includegraphics[trim={0 96 0 192},clip,width=0.32\linewidth]{figs/pics_new/MNIST_random.png} & \includegraphics[trim={0 96 0 192},clip,width=0.32\linewidth]{figs/pics_new/MNIST_interp.png} \\[-3.5pt]

     \raisebox{10pt}{RAE-SN}
     & \includegraphics[trim={0 64 0 224},clip,width=0.32\linewidth]{figs/pics_new/MNIST_recons.png}&\includegraphics[trim={0 64 0 224},clip,width=0.32\linewidth]{figs/pics_new/MNIST_random.png} & \includegraphics[trim={0 64 0 224},clip,width=0.32\linewidth]{figs/pics_new/MNIST_interp.png} \\[-3.5pt]

     \raisebox{10pt}{RAE}
     & \includegraphics[trim={0 32 0 256},clip,width=0.32\linewidth]{figs/pics_new/MNIST_recons.png}&\includegraphics[trim={ 0 32 0 256},clip,width=0.32\linewidth]{figs/pics_new/MNIST_random.png} & \includegraphics[trim={ 0 32 0 256},clip,width=0.32\linewidth]{figs/pics_new/MNIST_interp.png} \\[-3.5pt]

     \raisebox{10pt}{AE}
     & \includegraphics[trim={ 0 0 0 288},clip,width=0.32\linewidth]{figs/pics_new/MNIST_recons.png}&\includegraphics[trim={ 0 0 0 288},clip,width=0.32\linewidth]{figs/pics_new/MNIST_random.png} & \includegraphics[trim={ 0 0 0 288},clip,width=0.32\linewidth]{figs/pics_new/MNIST_interp.png} \\

\raisebox{10pt}{GT}
     & \includegraphics[trim={ 0 288 0
       0},clip,width=0.32\linewidth]{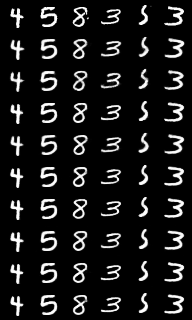}
     \\[-3.5pt]

     \raisebox{10pt}{VAE}
     & \includegraphics[trim={0 256 0 32},clip,width=0.32\linewidth]{figs/second_pics_new/MNIST_recons.png}&\includegraphics[trim={0 256 0 32},clip,width=0.32\linewidth]{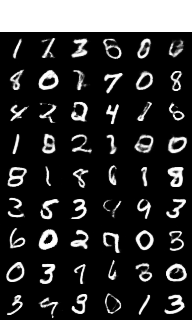} & \includegraphics[trim={0 256 0 32},clip,width=0.32\linewidth]{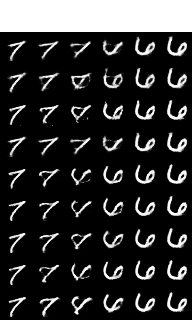} \\[-3.5pt]

     \raisebox{10pt}{CV-VAE}
     & \includegraphics[trim={0 224 0 64},clip,width=0.32\linewidth]{figs/second_pics_new/MNIST_recons.png}&\includegraphics[trim={0 224 0 64},clip,width=0.32\linewidth]{figs/second_pics_new/MNIST_random.png} & \includegraphics[trim={0 224 0 64},clip,width=0.32\linewidth]{figs/second_pics_new/MNIST_interp.png} \\[-3.5pt]

     \raisebox{10pt}{WAE}
     & \includegraphics[trim={ 0 192 0 96},clip,width=0.32\linewidth]{figs/second_pics_new/MNIST_recons.png}&\includegraphics[trim={ 0 192 0 96},clip,width=0.32\linewidth]{figs/second_pics_new/MNIST_random.png} & \includegraphics[trim={ 0 192 0 96},clip,width=0.32\linewidth]{figs/second_pics_new/MNIST_interp.png} \\[-3.5pt]

          \raisebox{10pt}{2sVAE}
     & \includegraphics[trim={ 0 160 0 128},clip,width=0.32\linewidth]{figs/second_pics_new/MNIST_recons.png}&\includegraphics[trim={0 160 0 128},clip,width=0.32\linewidth]{figs/second_pics_new/MNIST_random.png} & \includegraphics[trim={0 160 0 128},clip,width=0.32\linewidth]{figs/second_pics_new/MNIST_interp.png} \\[-3.5pt]

     \raisebox{10pt}{RAE-GP}
     & \includegraphics[trim={0 128 0 160},clip,width=0.32\linewidth]{figs/second_pics_new/MNIST_recons.png}&\includegraphics[trim={0 128 0 160},clip,width=0.32\linewidth]{figs/second_pics_new/MNIST_random.png} & \includegraphics[trim={0 128 0 160},clip,width=0.32\linewidth]{figs/second_pics_new/MNIST_interp.png} \\[-3.5pt]

     \raisebox{10pt}{RAE-L2}
     & \includegraphics[trim={0 96 0 192},clip,width=0.32\linewidth]{figs/second_pics_new/MNIST_recons.png}&\includegraphics[trim={0 96 0 192},clip,width=0.32\linewidth]{figs/second_pics_new/MNIST_random.png} & \includegraphics[trim={0 96 0 192},clip,width=0.32\linewidth]{figs/second_pics_new/MNIST_interp.png} \\[-3.5pt]

     \raisebox{10pt}{RAE-SN}
     & \includegraphics[trim={0 64 0 224},clip,width=0.32\linewidth]{figs/second_pics_new/MNIST_recons.png}&\includegraphics[trim={0 64 0 224},clip,width=0.32\linewidth]{figs/second_pics_new/MNIST_random.png} & \includegraphics[trim={0 64 0 224},clip,width=0.32\linewidth]{figs/second_pics_new/MNIST_interp.png} \\[-3.5pt]

     \raisebox{10pt}{RAE}
     & \includegraphics[trim={0 32 0 256},clip,width=0.32\linewidth]{figs/second_pics_new/MNIST_recons.png}&\includegraphics[trim={ 0 32 0 256},clip,width=0.32\linewidth]{figs/second_pics_new/MNIST_random.png} & \includegraphics[trim={ 0 32 0 256},clip,width=0.32\linewidth]{figs/second_pics_new/MNIST_interp.png} \\[-3.5pt]

     \raisebox{10pt}{AE}
     & \includegraphics[trim={ 0 0 0 288},clip,width=0.32\linewidth]{figs/second_pics_new/MNIST_recons.png}&\includegraphics[trim={ 0 0 0 288},clip,width=0.32\linewidth]{figs/second_pics_new/MNIST_random.png} & \includegraphics[trim={ 0 0 0 288},clip,width=0.32\linewidth]{figs/pics_new/MNIST_interp.png} \\

     \bottomrule

   \end{tabular}
\end{sc}}
 \vspace{2.5mm}
 \caption{Qualitative evaluation for sample quality for VAEs, WAEs and RAEs on
 MNIST.
 Left: reconstructed samples (top row is ground truth).
 Middle: randomly generated samples.
 Right: spherical interpolations between two images (first and last column).
 }
 \label{fig:mnist-pics}
\end{figure*}

\begin{figure*}[h]
 \centering
 \scalebox{.9}
{\setlength\tabcolsep{3pt}
\begin{sc}
 \small
   \begin{tabular}{ r c c c}
   \toprule
       & Reconstructions & Random Samples & Interpolations \\
       \midrule
     \raisebox{10pt}{GT}
     & \includegraphics[trim={ 0 288 0
       0},clip,width=0.32\linewidth]{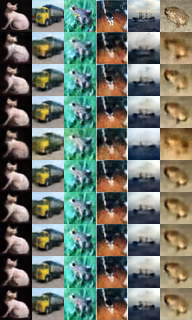}
     \\[-3.5pt]

     \raisebox{10pt}{VAE}
     & \includegraphics[trim={0 256 0 32},clip,width=0.32\linewidth]{figs/pics_new/CIFAR_10_recons.png}&\includegraphics[trim={ 0 256 0 32},clip,width=0.32\linewidth]{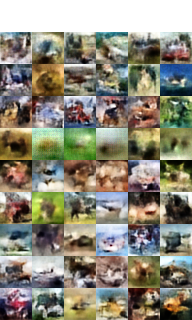} & \includegraphics[trim={ 0 256 0 32},clip,width=0.32\linewidth]{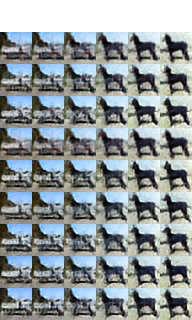} \\[-3.5pt]

     \raisebox{10pt}{CV-VAE}
     & \includegraphics[trim={0 224 0 64},clip,width=0.32\linewidth]{figs/pics_new/CIFAR_10_recons.png}&\includegraphics[trim={0 224 0 64},clip,width=0.32\linewidth]{figs/pics_new/CIFAR_10_random.png} & \includegraphics[trim={0 224 0 64},clip,width=0.32\linewidth]{figs/pics_new/CIFAR_10_interp.png} \\[-3.5pt]

     \raisebox{10pt}{WAE}
     & \includegraphics[trim={ 0 192 0 96},clip,width=0.32\linewidth]{figs/pics_new/CIFAR_10_recons.png}&\includegraphics[trim={0 192 0 96},clip,width=0.32\linewidth]{figs/pics_new/CIFAR_10_random.png} & \includegraphics[trim={0 192 0 96},clip,width=0.32\linewidth]{figs/pics_new/CIFAR_10_interp.png} \\[-3.5pt]

     \raisebox{10pt}{2sVAE}
     & \includegraphics[trim={ 0 160 0 128},clip,width=0.32\linewidth]{figs/pics_new/CIFAR_10_recons.png}&\includegraphics[trim={0 160 0 128},clip,width=0.32\linewidth]{figs/pics_new/CIFAR_10_random.png} & \includegraphics[trim={0 160 0 128},clip,width=0.32\linewidth]{figs/pics_new/CIFAR_10_interp.png} \\[-3.5pt]

     \raisebox{10pt}{RAE-GP}
     & \includegraphics[trim={0 128 0 160},clip,width=0.32\linewidth]{figs/pics_new/CIFAR_10_recons.png}&\includegraphics[trim={0 128 0 160},clip,width=0.32\linewidth]{figs/pics_new/CIFAR_10_random.png} & \includegraphics[trim={0 128 0 160},clip,width=0.32\linewidth]{figs/pics_new/CIFAR_10_interp.png} \\[-3.5pt]

     \raisebox{10pt}{RAE-L2}
     & \includegraphics[trim={0 96 0 192},clip,width=0.32\linewidth]{figs/pics_new/CIFAR_10_recons.png}&\includegraphics[trim={0 96 0 192},clip,width=0.32\linewidth]{figs/pics_new/CIFAR_10_random.png} & \includegraphics[trim={0 96 0 192},clip,width=0.32\linewidth]{figs/pics_new/CIFAR_10_interp.png} \\[-3.5pt]

     \raisebox{10pt}{RAE-SN}
     & \includegraphics[trim={0 64 0 224},clip,width=0.32\linewidth]{figs/pics_new/CIFAR_10_recons.png}&\includegraphics[trim={0 64 0 224},clip,width=0.32\linewidth]{figs/pics_new/CIFAR_10_random.png} & \includegraphics[trim={0 64 0 224},clip,width=0.32\linewidth]{figs/pics_new/CIFAR_10_interp.png} \\[-3.5pt]

     \raisebox{10pt}{RAE}
     & \includegraphics[trim={0 32 0 256},clip,width=0.32\linewidth]{figs/pics_new/CIFAR_10_recons.png}&\includegraphics[trim={ 0 32 0 256},clip,width=0.32\linewidth]{figs/pics_new/CIFAR_10_random.png} & \includegraphics[trim={ 0 32 0 256},clip,width=0.32\linewidth]{figs/pics_new/CIFAR_10_interp.png} \\[-3.5pt]

     \raisebox{10pt}{AE}
     & \includegraphics[trim={ 0 0 0 288},clip,width=0.32\linewidth]{figs/pics_new/CIFAR_10_recons.png}&\includegraphics[trim={ 0 0 0 288},clip,width=0.32\linewidth]{figs/pics_new/CIFAR_10_random.png} & \includegraphics[trim={ 0 0 0 288},clip,width=0.32\linewidth]{figs/pics_new/CIFAR_10_interp.png} \\

     \raisebox{10pt}{GT}
     & \includegraphics[trim={ 0 288 0
       0},clip,width=0.32\linewidth]{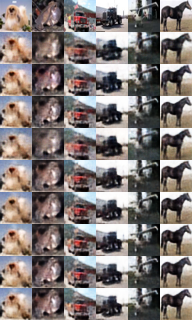}
     \\[-3.5pt]

     \raisebox{10pt}{VAE}
     & \includegraphics[trim={0 256 0 32},clip,width=0.32\linewidth]{figs/second_pics_new/CIFAR_10_recons.png}&\includegraphics[trim={ 0 256 0 32},clip,width=0.32\linewidth]{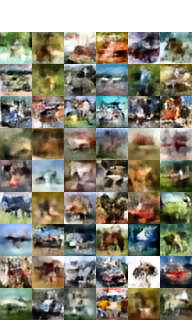} & \includegraphics[trim={ 0 256 0 32},clip,width=0.32\linewidth]{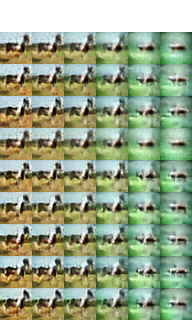} \\[-3.5pt]

     \raisebox{10pt}{CV-VAE}
     & \includegraphics[trim={0 224 0 64},clip,width=0.32\linewidth]{figs/second_pics_new/CIFAR_10_recons.png}&\includegraphics[trim={0 224 0 64},clip,width=0.32\linewidth]{figs/second_pics_new/CIFAR_10_random.png} & \includegraphics[trim={0 224 0 64},clip,width=0.32\linewidth]{figs/second_pics_new/CIFAR_10_interp.png} \\[-3.5pt]

     \raisebox{10pt}{WAE}
     & \includegraphics[trim={ 0 192 0 96},clip,width=0.32\linewidth]{figs/second_pics_new/CIFAR_10_recons.png}&\includegraphics[trim={0 192 0 96},clip,width=0.32\linewidth]{figs/second_pics_new/CIFAR_10_random.png} & \includegraphics[trim={0 192 0 96},clip,width=0.32\linewidth]{figs/second_pics_new/CIFAR_10_interp.png} \\[-3.5pt]

     \raisebox{10pt}{2sVAE}
     & \includegraphics[trim={ 0 160 0 128},clip,width=0.32\linewidth]{figs/second_pics_new/CIFAR_10_recons.png}&\includegraphics[trim={0 160 0 128},clip,width=0.32\linewidth]{figs/second_pics_new/CIFAR_10_random.png} & \includegraphics[trim={0 160 0 128},clip,width=0.32\linewidth]{figs/second_pics_new/CIFAR_10_interp.png} \\[-3.5pt]

     \raisebox{10pt}{RAE-GP}
     & \includegraphics[trim={0 128 0 160},clip,width=0.32\linewidth]{figs/second_pics_new/CIFAR_10_recons.png}&\includegraphics[trim={0 128 0 160},clip,width=0.32\linewidth]{figs/second_pics_new/CIFAR_10_random.png} & \includegraphics[trim={0 128 0 160},clip,width=0.32\linewidth]{figs/second_pics_new/CIFAR_10_interp.png} \\[-3.5pt]

     \raisebox{10pt}{RAE-L2}
     & \includegraphics[trim={0 96 0 192},clip,width=0.32\linewidth]{figs/second_pics_new/CIFAR_10_recons.png}&\includegraphics[trim={0 96 0 192},clip,width=0.32\linewidth]{figs/second_pics_new/CIFAR_10_random.png} & \includegraphics[trim={0 96 0 192},clip,width=0.32\linewidth]{figs/second_pics_new/CIFAR_10_interp.png} \\[-3.5pt]

     \raisebox{10pt}{RAE-SN}
     & \includegraphics[trim={0 64 0 224},clip,width=0.32\linewidth]{figs/second_pics_new/CIFAR_10_recons.png}&\includegraphics[trim={0 64 0 224},clip,width=0.32\linewidth]{figs/second_pics_new/CIFAR_10_random.png} & \includegraphics[trim={0 64 0 224},clip,width=0.32\linewidth]{figs/second_pics_new/CIFAR_10_interp.png} \\[-3.5pt]

     \raisebox{10pt}{RAE}
     & \includegraphics[trim={0 32 0 256},clip,width=0.32\linewidth]{figs/second_pics_new/CIFAR_10_recons.png}&\includegraphics[trim={ 0 32 0 256},clip,width=0.32\linewidth]{figs/second_pics_new/CIFAR_10_random.png} & \includegraphics[trim={ 0 32 0 256},clip,width=0.32\linewidth]{figs/second_pics_new/CIFAR_10_interp.png} \\[-3.5pt]

     \raisebox{10pt}{AE}
     & \includegraphics[trim={ 0 0 0 288},clip,width=0.32\linewidth]{figs/second_pics_new/CIFAR_10_recons.png}&\includegraphics[trim={ 0 0 0 288},clip,width=0.32\linewidth]{figs/second_pics_new/CIFAR_10_random.png} & \includegraphics[trim={ 0 0 0 288},clip,width=0.32\linewidth]{figs/second_pics_new/CIFAR_10_interp.png} \\
     \bottomrule

   \end{tabular}
\end{sc}}
 \vspace{2.5mm}
 \caption{Qualitative evaluation for sample quality for VAEs, WAEs and RAEs on
 CIFAR-10.
 Left: reconstructed samples (top row is ground truth).
 Middle: randomly generated samples.
 Right: spherical interpolations between two images (first and last column).
 }
 \label{fig:cifar-pics}
\end{figure*}

\clearpage
\section{Investigating Overfitting}
\label{sec:knn}

\begin{figure}[h]
 \centering
 \scalebox{.9}
{\setlength\tabcolsep{3pt}
\begin{sc}
 \small
   \begin{tabular}{ l c c c c c c}
   \toprule
       & \multicolumn{2}{c}{MNIST} & \multicolumn{2}{c}{CIFAR-10} & \multicolumn{2}{c}{CELEBA} \\
       \midrule
       \raisebox{5pt}{\rotatebox[origin=c]{0}{VAE}} &

       \begin{tikzpicture}\node[anchor=south west,inner sep=0] at (0,0) {\includegraphics[width=0.16\linewidth]{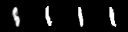}};\draw[red, ultra thick] (0, 0) rectangle (.55, .55);\end{tikzpicture}

       &\begin{tikzpicture}\node[anchor=south west,inner sep=0] at (0,0) {\includegraphics[width=0.16\linewidth]{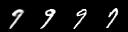}};\draw[red, ultra thick] (0, 0) rectangle (.55, .55);\end{tikzpicture}

       &\begin{tikzpicture}\node[anchor=south west,inner sep=0] at (0,0) {\includegraphics[width=0.16\linewidth]{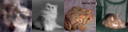}};\draw[red, ultra thick] (0, 0) rectangle (.55, .55);\end{tikzpicture}

       &\begin{tikzpicture}\node[anchor=south west,inner sep=0] at (0,0) {\includegraphics[width=0.16\linewidth]{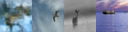}};\draw[red, ultra thick] (0, 0) rectangle (.55, .55);\end{tikzpicture}

       & \begin{tikzpicture}\node[anchor=south west,inner sep=0] at (0,0) {\includegraphics[width=0.16\linewidth]{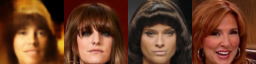}};\draw[red, ultra thick] (0, 0) rectangle (.55, .55);\end{tikzpicture}

       & \begin{tikzpicture}\node[anchor=south west,inner sep=0] at (0,0) {\includegraphics[width=0.16\linewidth]{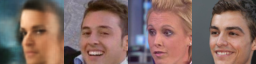}};\draw[red, ultra thick] (0, 0) rectangle (.55, .55);\end{tikzpicture}\\

       \raisebox{5pt}{\rotatebox[origin=c]{0}{CV-VAE}}

       & \begin{tikzpicture}\node[anchor=south west,inner sep=0] at (0,0) {\includegraphics[width=0.16\linewidth]{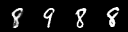}};\draw[red, ultra thick] (0, 0) rectangle (.55, .55);\end{tikzpicture}

       &\begin{tikzpicture}\node[anchor=south west,inner sep=0] at (0,0) {\includegraphics[width=0.16\linewidth]{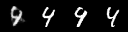}};\draw[red, ultra thick] (0, 0) rectangle (.55, .55);\end{tikzpicture}

       &\begin{tikzpicture}\node[anchor=south west,inner sep=0] at (0,0) {\includegraphics[width=0.16\linewidth]{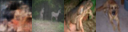}};\draw[red, ultra thick] (0, 0) rectangle (.55, .55);\end{tikzpicture}

       &\begin{tikzpicture}\node[anchor=south west,inner sep=0] at (0,0) {\includegraphics[width=0.16\linewidth]{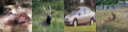}};\draw[red, ultra thick] (0, 0) rectangle (.55, .55);\end{tikzpicture}

       & \begin{tikzpicture}\node[anchor=south west,inner sep=0] at (0,0) {\includegraphics[width=0.16\linewidth]{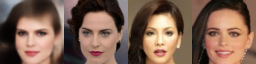}};\draw[red, ultra thick] (0, 0) rectangle (.55, .55);\end{tikzpicture}

       & \begin{tikzpicture}\node[anchor=south west,inner sep=0] at (0,0) {\includegraphics[width=0.16\linewidth]{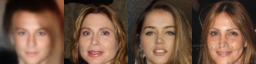}};\draw[red, ultra thick] (0, 0) rectangle (.55, .55);\end{tikzpicture}\\

       \raisebox{5pt}{\rotatebox[origin=c]{0}{WAE}} & \begin{tikzpicture}\node[anchor=south west,inner sep=0] at (0,0) {\includegraphics[width=0.16\linewidth]{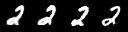}};\draw[red, ultra thick] (0, 0) rectangle (.55, .55);\end{tikzpicture}

       &\begin{tikzpicture}\node[anchor=south west,inner sep=0] at (0,0) {\includegraphics[width=0.16\linewidth]{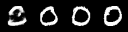}};\draw[red, ultra thick] (0, 0) rectangle (.55, .55);\end{tikzpicture}

       &\begin{tikzpicture}\node[anchor=south west,inner sep=0] at (0,0) {\includegraphics[width=0.16\linewidth]{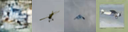}};\draw[red, ultra thick] (0, 0) rectangle (.55, .55);\end{tikzpicture}

       &\begin{tikzpicture}\node[anchor=south west,inner sep=0] at (0,0) {\includegraphics[width=0.16\linewidth]{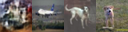}};\draw[red, ultra thick] (0, 0) rectangle (.55, .55);\end{tikzpicture}

       & \begin{tikzpicture}\node[anchor=south west,inner sep=0] at (0,0) {\includegraphics[width=0.16\linewidth]{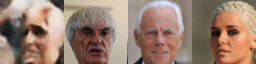}};\draw[red, ultra thick] (0, 0) rectangle (.55, .55);\end{tikzpicture}

       & \begin{tikzpicture}\node[anchor=south west,inner sep=0] at (0,0) {\includegraphics[width=0.16\linewidth]{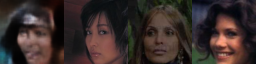}};\draw[red, ultra thick] (0, 0) rectangle (.55, .55);\end{tikzpicture}\\

       \raisebox{5pt}{\rotatebox[origin=c]{0}{RAE-GP}}

       & \begin{tikzpicture}\node[anchor=south west,inner sep=0] at (0,0) {\includegraphics[width=0.16\linewidth]{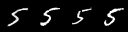}};\draw[red, ultra thick] (0, 0) rectangle (.55, .55);\end{tikzpicture}

       &\begin{tikzpicture}\node[anchor=south west,inner sep=0] at (0,0) {\includegraphics[width=0.16\linewidth]{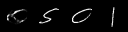}};\draw[red, ultra thick] (0, 0) rectangle (.55, .55);\end{tikzpicture}

       &\begin{tikzpicture}\node[anchor=south west,inner sep=0] at (0,0) {\includegraphics[width=0.16\linewidth]{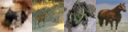}};\draw[red, ultra thick] (0, 0) rectangle (.55, .55);\end{tikzpicture}

       &\begin{tikzpicture}\node[anchor=south west,inner sep=0] at (0,0) {\includegraphics[width=0.16\linewidth]{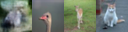}};\draw[red, ultra thick] (0, 0) rectangle (.55, .55);\end{tikzpicture}

       & \begin{tikzpicture}\node[anchor=south west,inner sep=0] at (0,0) {\includegraphics[width=0.16\linewidth]{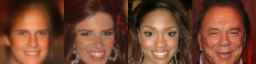}};\draw[red, ultra thick] (0, 0) rectangle (.55, .55);\end{tikzpicture}

       & \begin{tikzpicture}\node[anchor=south west,inner sep=0] at (0,0) {\includegraphics[width=0.16\linewidth]{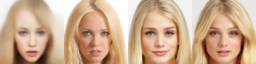}};\draw[red, ultra thick] (0, 0) rectangle (.55, .55);\end{tikzpicture}\\

       \raisebox{5pt}{\rotatebox[origin=c]{0}{RAE-L2}}

       & \begin{tikzpicture}\node[anchor=south west,inner sep=0] at (0,0) {\includegraphics[width=0.16\linewidth]{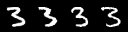}};\draw[red, ultra thick] (0, 0) rectangle (.55, .55);\end{tikzpicture}

       &\begin{tikzpicture}\node[anchor=south west,inner sep=0] at (0,0) {\includegraphics[width=0.16\linewidth]{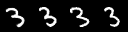}};\draw[red, ultra thick] (0, 0) rectangle (.55, .55);\end{tikzpicture}

       &\begin{tikzpicture}\node[anchor=south west,inner sep=0] at (0,0) {\includegraphics[width=0.16\linewidth]{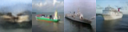}};\draw[red, ultra thick] (0, 0) rectangle (.55, .55);\end{tikzpicture}

       &\begin{tikzpicture}\node[anchor=south west,inner sep=0] at (0,0) {\includegraphics[width=0.16\linewidth]{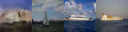}};\draw[red, ultra thick] (0, 0) rectangle (.55, .55);\end{tikzpicture}

       & \begin{tikzpicture}\node[anchor=south west,inner sep=0] at (0,0) {\includegraphics[width=0.16\linewidth]{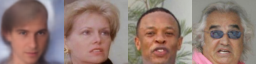}};\draw[red, ultra thick] (0, 0) rectangle (.55, .55);\end{tikzpicture}

       & \begin{tikzpicture}\node[anchor=south west,inner sep=0] at (0,0) {\includegraphics[width=0.16\linewidth]{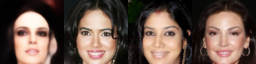}};\draw[red, ultra thick] (0, 0) rectangle (.55, .55);\end{tikzpicture}\\

       \raisebox{5pt}{\rotatebox[origin=c]{0}{RAE-SN}}

       & \begin{tikzpicture}\node[anchor=south west,inner sep=0] at (0,0) {\includegraphics[width=0.16\linewidth]{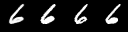}};\draw[red, ultra thick] (0, 0) rectangle (.55, .55);\end{tikzpicture}

       &\begin{tikzpicture}\node[anchor=south west,inner sep=0] at (0,0) {\includegraphics[width=0.16\linewidth]{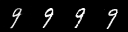}};\draw[red, ultra thick] (0, 0) rectangle (.55, .55);\end{tikzpicture}

       &\begin{tikzpicture}\node[anchor=south west,inner sep=0] at (0,0) {\includegraphics[width=0.16\linewidth]{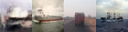}};\draw[red, ultra thick] (0, 0) rectangle (.55, .55);\end{tikzpicture}

       &\begin{tikzpicture}\node[anchor=south west,inner sep=0] at (0,0) {\includegraphics[width=0.16\linewidth]{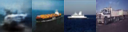}};\draw[red, ultra thick] (0, 0) rectangle (.55, .55);\end{tikzpicture}

       & \begin{tikzpicture}\node[anchor=south west,inner sep=0] at (0,0) {\includegraphics[width=0.16\linewidth]{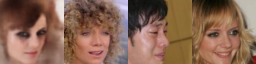}};\draw[red, ultra thick] (0, 0) rectangle (.55, .55);\end{tikzpicture}

       & \begin{tikzpicture}\node[anchor=south west,inner sep=0] at (0,0) {\includegraphics[width=0.16\linewidth]{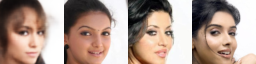}};\draw[red, ultra thick] (0, 0) rectangle (.55, .55);\end{tikzpicture}\\

       \raisebox{5pt}{\rotatebox[origin=c]{0}{RAE}}

       & \begin{tikzpicture}\node[anchor=south west,inner sep=0] at (0,0) {\includegraphics[width=0.16\linewidth]{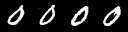}};\draw[red, ultra thick] (0, 0) rectangle (.55, .55);\end{tikzpicture}

       &\begin{tikzpicture}\node[anchor=south west,inner sep=0] at (0,0) {\includegraphics[width=0.16\linewidth]{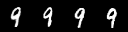}};\draw[red, ultra thick] (0, 0) rectangle (.55, .55);\end{tikzpicture}

       &\begin{tikzpicture}\node[anchor=south west,inner sep=0] at (0,0) {\includegraphics[width=0.16\linewidth]{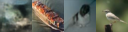}};\draw[red, ultra thick] (0, 0) rectangle (.55, .55);\end{tikzpicture}

       &\begin{tikzpicture}\node[anchor=south west,inner sep=0] at (0,0) {\includegraphics[width=0.16\linewidth]{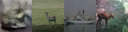}};\draw[red, ultra thick] (0, 0) rectangle (.55, .55);\end{tikzpicture}

       & \begin{tikzpicture}\node[anchor=south west,inner sep=0] at (0,0) {\includegraphics[width=0.16\linewidth]{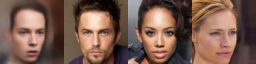}};\draw[red, ultra thick] (0, 0) rectangle (.55, .55);\end{tikzpicture}

       & \begin{tikzpicture}\node[anchor=south west,inner sep=0] at (0,0) {\includegraphics[width=0.16\linewidth]{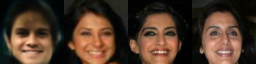}};\draw[red, ultra thick] (0, 0) rectangle (.55, .55);\end{tikzpicture}\\

       \raisebox{5pt}{\rotatebox[origin=c]{0}{AE}}

       & \begin{tikzpicture}\node[anchor=south west,inner sep=0] at (0,0) {\includegraphics[width=0.16\linewidth]{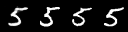}};\draw[red, ultra thick] (0, 0) rectangle (.55, .55);\end{tikzpicture}

       &\begin{tikzpicture}\node[anchor=south west,inner sep=0] at (0,0) {\includegraphics[width=0.16\linewidth]{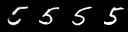}};\draw[red, ultra thick] (0, 0) rectangle (.55, .55);\end{tikzpicture}

       &\begin{tikzpicture}\node[anchor=south west,inner sep=0] at (0,0) {\includegraphics[width=0.16\linewidth]{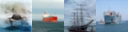}};\draw[red, ultra thick] (0, 0) rectangle (.55, .55);\end{tikzpicture}

       &\begin{tikzpicture}\node[anchor=south west,inner sep=0] at (0,0) {\includegraphics[width=0.16\linewidth]{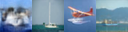}};\draw[red, ultra thick] (0, 0) rectangle (.55, .55);\end{tikzpicture}

       & \begin{tikzpicture}\node[anchor=south west,inner sep=0] at (0,0) {\includegraphics[width=0.16\linewidth]{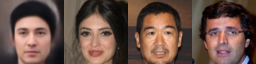}};\draw[red, ultra thick] (0, 0) rectangle (.55, .55);\end{tikzpicture}

       & \begin{tikzpicture}\node[anchor=south west,inner sep=0] at (0,0) {\includegraphics[width=0.16\linewidth]{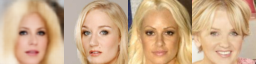}};\draw[red, ultra thick] (0, 0) rectangle (.55, .55);\end{tikzpicture}\\

     \bottomrule

   \end{tabular}
\end{sc}}
 \vspace{2.5mm}
 \caption{Nearest neighbors to generated samples (leftmost image, red box) from
 training set. It seems that the models have generalized well and fitting only
 10 Gaussians to the latent space prevents overfitting.}
 \label{fig:nn-pics}
\end{figure}

\clearpage
\section{Visualizing Ex-Post Density Estimation}
\label{sec:viz_ex_post_density}

\begin{figure*}[h]
 \centering
 \scalebox{.9}
{\setlength\tabcolsep{3pt}
\begin{sc}
 \small
   \begin{tabular}{c c c}
   \toprule
       $\mathcal{N}(0,I)$ & $\mathcal{N}(\mu,\Sigma)$ &                                                          $\mathsf{GMM} (k=10)$ \\
       \midrule

       \includegraphics[width=0.3\linewidth]{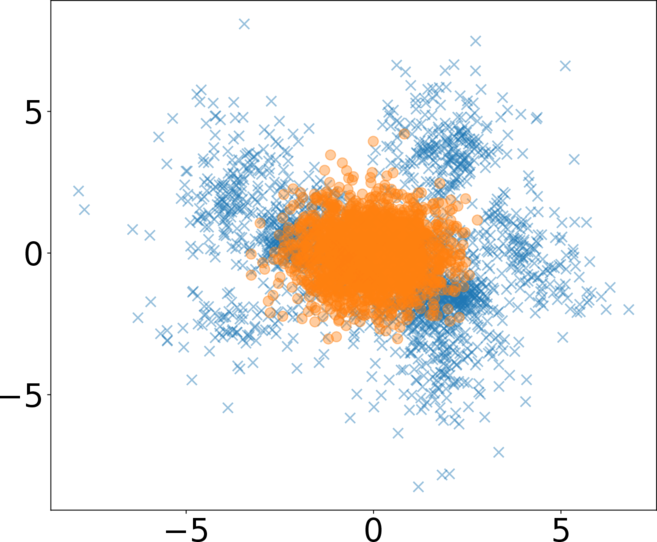}
       &\includegraphics[width=0.3\linewidth]{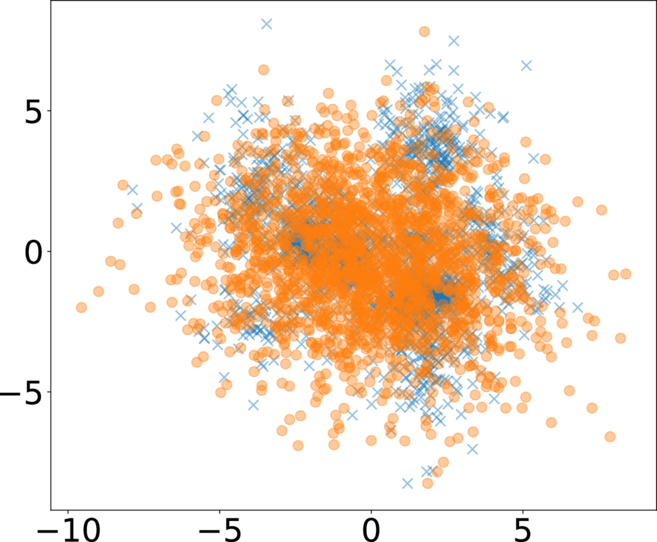}& \includegraphics[width=0.3\linewidth]{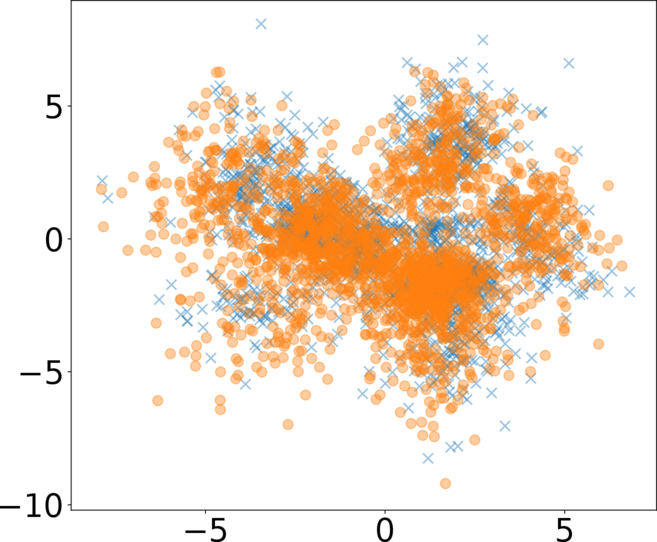} \\[-3.5pt]

     \bottomrule

   \end{tabular}
\end{sc}}
 \vspace{2.5mm}
 \caption{Different density estimations of the 2-dimensional latent space of a VAE
   learned on MNIST. The blue points are 2000 test set samples while
   the orange ones are drawn from the estimator indicated in each
   column: isotropic Gaussian (left), multivariate Gaussian with mean
   and covariance estimated on the training set (center) and a
   10-component GMM (right).
   This clearly shows the aggregated posterior mismatch w.r.t. to the
   isotropic Gaussian prior imposed by VAEs and how ex-post
   density estimation can help fix the estimate.
 }
 \label{fig:2dmnist-lv}
\end{figure*}

To visualize that ex-post density estimation does indeed help reduce
the mismatch between the aggregated posterior and the prior we train a
VAE on the MNIST dataset whose latent space is $2$ dimensional.
The unique advantage of this setting is that one can simply visualize
the density of test sample in the latent space by plotting them as a
scatterplot.
As it can be seen from figure \ref{fig:2dmnist-lv}, an expressive density estimator effectively fixes the miss-match and this as reported earlier results in better sample quality. 

\begin{figure*}[h]
 \centering
 \scalebox{.9}
{\setlength\tabcolsep{3pt}
\begin{sc}
 \small
   \begin{tabular}{r c c c}
   \toprule
       &$\mathcal{N}(0,I)$ & $\mathcal{N}(\mu,\Sigma)$ &                                                          $\mathsf{GMM} (k=10)$ \\
       \midrule

      \raisebox{50pt}{VAE}     &
       \includegraphics[width=0.22\linewidth]{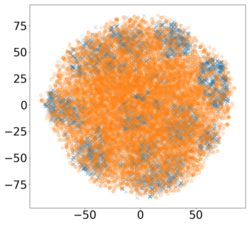}
       &\includegraphics[width=0.22\linewidth]{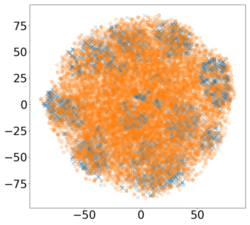}& \includegraphics[width=0.22\linewidth]{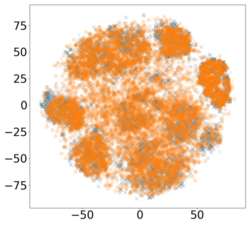} \\[-3.5pt]

     \raisebox{50pt}{CV-VAE}     &
       \includegraphics[width=0.22\linewidth]{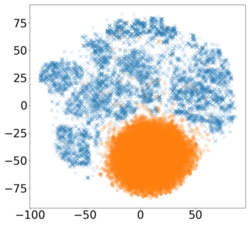}
       &\includegraphics[width=0.22\linewidth]{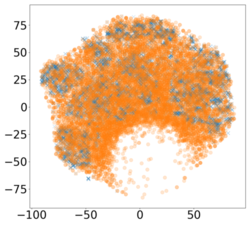}& \includegraphics[width=0.22\linewidth]{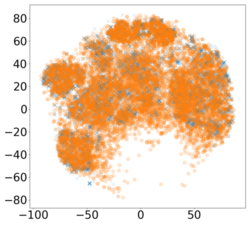} \\[-3.5pt]
      \raisebox{50pt}{WAE}     &
       \includegraphics[width=0.22\linewidth]{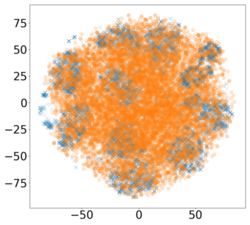}
       &\includegraphics[width=0.22\linewidth]{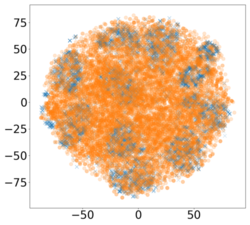}& \includegraphics[width=0.22\linewidth]{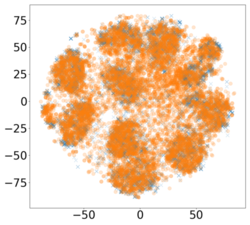} \\[-3.5pt]
      \raisebox{50pt}{RAE-GP}     &
       \includegraphics[width=0.22\linewidth]{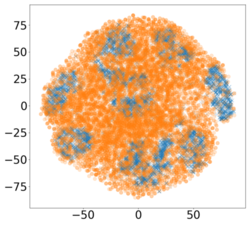}
       &\includegraphics[width=0.22\linewidth]{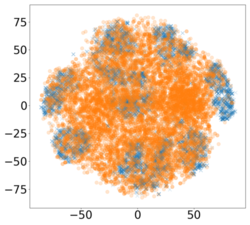}& \includegraphics[width=0.22\linewidth]{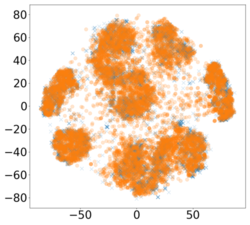} \\[-3.5pt]
      \raisebox{50pt}{RAE-L2}     &
       \includegraphics[width=0.22\linewidth]{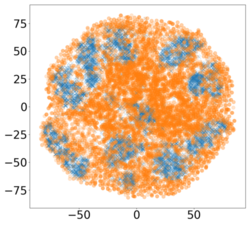}
       &\includegraphics[width=0.22\linewidth]{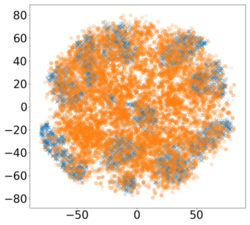}& \includegraphics[width=0.22\linewidth]{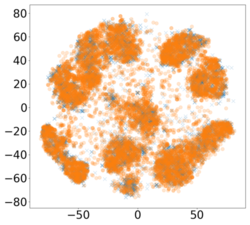} \\[-3.5pt]
      \raisebox{50pt}{RAE-SN}     &
       \includegraphics[width=0.22\linewidth]{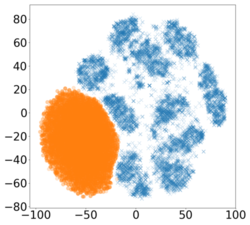}
       &\includegraphics[width=0.22\linewidth]{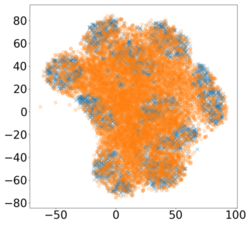}& \includegraphics[width=0.22\linewidth]{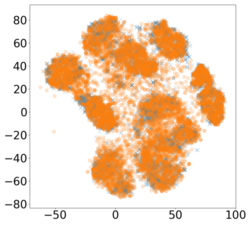} \\[-3.5pt]
      \raisebox{50pt}{RAE}     &
       \includegraphics[width=0.22\linewidth]{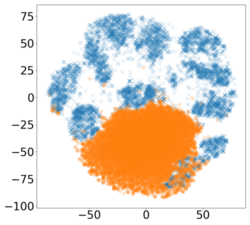}
       &\includegraphics[width=0.22\linewidth]{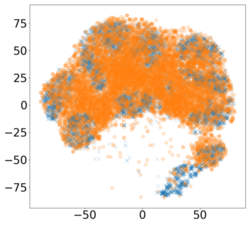}& \includegraphics[width=0.22\linewidth]{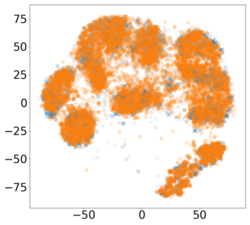} \\[-3.5pt]
     \raisebox{50pt}{AE}     &
       \includegraphics[width=0.22\linewidth]{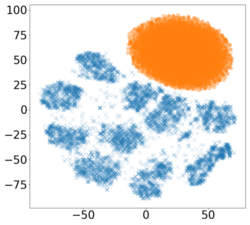}
       &\includegraphics[width=0.22\linewidth]{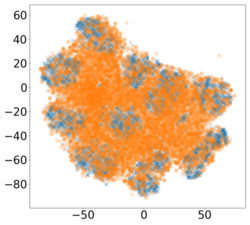}& \includegraphics[width=0.22\linewidth]{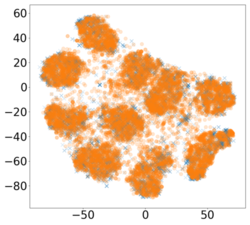} \\[-3.5pt]
     \bottomrule

   \end{tabular}
\end{sc}}
 \vspace{2.5mm}
 \caption{Different density estimations of the 16-dimensional latent
   spaces learned by all models on MNIST (see Table 1) here projected
   in 2d via T-SNE.
   The blue points are 2000 test set samples while
   the orange ones are drawn from the estimator indicated in each
   column: isotropic Gaussian (left), multivariate Gaussian with mean
   and covariance estimated on the training set (center) and a
   10-component GMM (right).
   Ex-post density estimation greatly improves sampling the latent space.
 }

 \label{fig:mnist-lv}
\end{figure*}

Here in figure \ref{fig:mnist-lv} we perform the same visualization on
with all the models trained on the MNIST dataset as employed on our
large evaluation in Table 1. Clearly every model depicts rather large mis-match between aggregate posterior and prior. Once again the advantage of ex-post density estimate is clearly visible.

\clearpage
\section{Combining multiple regularization terms}
\label{sec:more-regs}
The rather intriguing facts that AE without explicit decoder regularization performs reasonably well as seen from table \ref{tab:fids-main}, indicates that convolutional neural networks when combined with gradient based optimizers inherit some implicit regularization. This motivates us to investigate a few different combinations of regularizations e.g. we regularize the decoder of an auto-encoder while drop the regularization in the $z$ space. The results of this experiment is reported in the row marked AE-L2 in table \ref{tab:fids-aux}.

Further more a recent GAN literature \cite{lucic2018gans} report that often a combination of regularizations boost performance of neural networks. Following this, we combine multiple regularization techniques in out framework. However note that this rather drastically increases the hyper parameters and the models become harder to train and goes against the core theme of this work, which strives for simplicity. Hence we perform simplistic effort to tune all the hyper parameters to see if this can provide boost in the performance, which seem not to be the case. These experiments are summarized in the second half of the table \ref{tab:fids-aux}

\begin{table}[!t]
  \setlength{\tabcolsep}{5pt}
  \centering
  \begin{sc}
    \scriptsize
    \begin{tabular}{l@{\hskip 0.05in} l@{\hskip 0.1in} r@{\hskip 0.1in} r@{\hskip 0.1in} r@{\hskip 0.1in} r@{\hskip 0.1in} r@{\hskip 0.1in} r@{\hskip 0.1in} r@{\hskip 0.1in} r@{\hskip 0.1in} r@{\hskip 0.1in} r@{\hskip 0.1in} r@{\hskip 0.1in} r@{\hskip 0.1in} r@{\hskip 0.1in} r@{\hskip 0.1in}}
      \toprule
       & \multicolumn{4}{c}{MNIST} & \multicolumn{4}{c}{CIFAR} & \multicolumn{4}{c}{CelebA} \\
    \cmidrule(r){2-5}\cmidrule(r){6-9}\cmidrule(r){10-13}
      & \multirow{2}[3]{*}{Rec.} & \multicolumn{3}{c}{Samples} & \multirow{2}[3]{*}{Rec.} & \multicolumn{3}{c}{Samples} & \multirow{2}[3]{*}{Rec.} & \multicolumn{3}{c}{Samples}\\
      \cmidrule(lr){3-5} \cmidrule(lr){7-9}\cmidrule(lr){11-13}
       & &  $\mathcal{N}$ & $\mathsf{GMM}$ & $\mathsf{Interp.}$& &  $\mathcal{N}$ & $\mathsf{GMM}$&$\mathsf{Interp.}$& &  $\mathcal{N}$ & $\mathsf{GMM}$&$\mathsf{Interp.}$\\
      \cmidrule(r){2-5}\cmidrule(r){6-9}\cmidrule(r){10-13}
      
      \ourmodel{}-GP        & 14.04 & \textbf{22.21} & 11.54 &  15.32 & 32.17 & {83.05} & 76.33 & 64.08 & \textbf{39.71} & 116.30 & 45.63 & 47.00 \\
      \ourmodel{}-L2        & {10.53} & 22.22 & \textbf{8.69} & \textbf{14.54} & 32.24 & \textbf{80.80} & {74.16} & {62.54} & 43.52 & 51.13 & 47.97 & 45.98 \\
      \ourmodel{}-SN        & 15.65 & {19.67} & 11.74 & 15.15 &
                                                                {27.61}
                                                               & 84.25
                                                                                          & {75.30} & 63.62 & {36.01} & \textbf{44.74} & \textbf{40.95} & \textbf{39.53}\\
            \ourmodel{}           & 11.67 & 23.92 & 9.81 & 14.67 & {29.05} & 83.87 & 76.28 & 63.27 & 40.18 & 48.20 & 44.68 & 43.67\\
      AE                    & 12.95 & 58.73 & 10.66 & 17.12 & {30.52} & 84.74 & 76.47 & \textbf{61.57} & 40.79 & 127.85 & 45.10 & 50.94\\
      AE-L2 &  11.19 & 315.15 & 9.36 &  17.15 & 34.35 & 247.48 & 75.40 & 61.09 & 44.72 & 346.29 & 48.42 & 56.16 & \\

      \midrule
      \ourmodel{}-GP-L2        & \textbf{9.70} & 72.64 & 9.07 & 16.07 & 33.25 & 187.07 & 79.03 & 62.48 & 47.06 & 72.09 & 51.55 & 50.28 \\
      \ourmodel{}-L2-SN        & 10.67 &  50.63 & 9.42 & 15.73 & \textbf{24.17} & 240.27 & \textbf{74.10} & 61.71 & 39.90 & 180.39 & 44.39 & 42.97 \\
      \ourmodel{}-SN-GP        & 17.00 & 139.61 & 13.12 & 16.62 & 33.04 & 284.36 & 75.23 & 62.86 & 63.75 & 299.69 & 71.05 & 68.87 \\
      \ourmodel{}-L2-SN-GP        & 16.75 & 144.51 & 13.93 & 16.75 & 29.96 & 290.34 & 74.22 & 61.93 & 68.86 & 318.67 & 75.04 & 74.29 \\
      \bottomrule
  \end{tabular}
  \end{sc}
  \vspace{2.5mm}
  \caption{
  Comparing multiple regularization schemes for RAE models. The
  improvement in reconstruction, random sample quality and
  interpolated test samples is generally comparable, but hardly much better.
  This can be explained with the fact that the additional
  regularization losses make tuning their hyperparameters more
  difficult, in practice.
  }
  \label{tab:fids-aux}
\end{table}


\end{document}